\documentclass{article} 
\usepackage{iclr2025_conference,times}

\usepackage{multirow}

\usepackage{amssymb, amsmath, amsthm}
\usepackage{multirow}
\usepackage{hyperref}
\usepackage{cleveref}
\usepackage{url}
\usepackage{microtype}
\usepackage{graphicx}
\usepackage{booktabs} 
\usepackage{caption}
\usepackage{wrapfig}
\usepackage{subcaption}
\usepackage{microtype}
\usepackage{listings}
\usepackage{xcolor}
\usepackage{xspace}
\usepackage{makecell}
\usepackage{rotating}

\usepackage{algorithm}
\usepackage{algorithmic}
\usepackage{footmisc}

\usepackage{longtable}

\usepackage{tocloft}
\newcommand{\listappendixname}{List of Appendices}
\newlistof[chapter]{appendices}{app}{\listappendixname}
\newcommand{\appendixtoc}[1]{
    \addcontentsline{app}{appendices}{\protect\numberline{\thesection}#1}}

\newcommand{\llamaii}{LLaMA-2\xspace}
\newcommand{\llamaiii}[1]{LLaMA-3-{#1}B-Base\xspace}
\newcommand{\mistral}[1]{Mistral-{#1}B-Base-v0.2\xspace}
\newcommand{\gemma}[1]{Gemma-2-{#1}B-Base\xspace}

\newcommand{\gpt}{GPT-4\xspace}
\newcommand{\gptMarch}{GPT-4 (2023-03-14)\xspace}

\newcommand{\gptTurbo}{GPT-4-Turbo\xspace}

\newcommand{\instructSM}[1][\textsc{Instruct-SkillMix}]{\textsc{{#1}}\xspace}

\newcommand{\instructSMSDALong}[1][\textsc{Instruct-SkillMix-Seed-Dataset-Agnostic}]{\textsc{{#1}}\xspace}
\newcommand{\instructSMSDA}[1][\textsc{Instruct-SkillMix}]{\textsc{{#1}}\xspace}
\newcommand{\instructSMSDAShort}[1][\textsc{Instr-SkillMix}]

\newcommand{\instructSMSDDLong}[1][\textsc{Instruct-SkillMix-Seed-Dataset-Dependent}]{\textsc{{#1}}\xspace}
\newcommand{\instructSMSDD}[1][\textsc{Instruct-SkillMix-D}]{\textsc{{#1}}\xspace}

\usepackage{listings}

\usepackage{minitoc}
\usepackage[toc,page,header]{appendix}
\lstdefinestyle{mypython}{
  language=python,
  breaklines=true,
  basicstyle=\fontsize{8.5}{13}\selectfont\ttfamily,
  keywordstyle=\bfseries\color{green!40!black},
}

\usepackage[strict]{changepage}

\usepackage{framed}
\definecolor{questionshade}{rgb}{0.95,0.95,1}
\definecolor{darkblue}{rgb}{0, 0, 0.55}
\definecolor{cb_red}{RGB}{213,94,0}
\definecolor{cb_blue}{RGB}{0,114,178}
\definecolor{cb_yellow}{RGB}{240,228,66}
\definecolor{cb_gray}{RGB}{204,204,204}
\definecolor{cb_orange}{RGB}{230,159,0}
\definecolor{cb_skyblue}{RGB}{86,180,233}
\definecolor{cb_green}{RGB}{0,158,115}
\definecolor{cb_purple}{RGB}{204,121,167}

\newenvironment{question}{%
  \MakeFramed{\advance\hsize-\width\FrameRestore}%
  \noindent\hspace{-4.55pt}
  \begin{adjustwidth}{}{7pt}%
  \tt \scriptsize

}
{%
\vspace{2pt}\end{adjustwidth}
\endMakeFramed%
  
}

\newcommand{\ca}[1]{{\color{cb_purple}#1}}

\newcommand{\cc}[1]{{\color{cb_blue}#1}}
\newcommand{\res}[2]{\ca{#1}/\cc{#2}}

\usepackage{hyperref}
\usepackage{url}

\title{{\sc instruct-skillmix}: A Powerful Pipeline for LLM Instruction Tuning}

\author{%
  Simran Kaur$^{1*}$, Simon Park$^{1*}$, Anirudh Goyal$^{2}$, Sanjeev Arora$^{1}$ \\
  $^{1}$ Princeton Language and Intelligence (PLI), Princeton University \\
  $^{2}$ Meta
}

\iclrfinalcopy 
\begin{document}

\maketitle
\def\thefootnote{*}\footnotetext{Equal contribution.} 
\def\thefootnote{\arabic{footnote}} 

\begin{abstract}

We introduce \instructSM{}\footnote{Source code can be found at \url{https://github.com/princeton-pli/Instruct-SkillMix}.}, an automated approach for creating diverse, high quality SFT data for instruction-following. The pipeline involves two stages, each leveraging an existing powerful LLM: (1) {\em Skill extraction}: uses the LLM to extract core ``skills'' for instruction-following by directly prompting the model. This is inspired by ``LLM metacognition'' of \cite{didolkar2024metacognitive}; (2) {\em Data generation:} uses the powerful LLM to generate (instruction, response) data that exhibit a randomly chosen pair of these skills. Here, the use of random skill combinations promotes diversity and difficulty. The estimated cost of creating the dataset is under \$600.

Vanilla SFT (i.e., no PPO, DPO, or RL methods) on data generated from \instructSM{} leads to strong gains on instruction following benchmarks such as AlpacaEval 2.0, MT-Bench, and WildBench. With just $4$K examples, \llamaiii{8} achieves 42.76\% length-controlled win rate on AlpacaEval 2.0, a level similar to frontier models like Claude 3 Opus and LLaMA-3.1-405B-Instruct. 

Ablation studies also suggest plausible reasons for why creating open instruction-tuning datasets via naive crowd-sourcing has proved difficult. In our dataset, adding $20\%$ low quality answers (``shirkers'') causes a noticeable degradation in performance.

The \instructSM{} pipeline seems flexible and adaptable to other settings. 

\end{abstract}
\section{Introduction}
{\em Instruction tuning} (sometimes also called {\em imitation learning}) is the first step in converting a base LLM trained on next-word prediction into a helpful and interactive agent. Whereas early versions of instruction tuning involved supervised fine-tuning (SFT) on traditional NLP question-answer datasets~\citep{wei2021FLAN}, nowadays, the SFT data is collected at high cost from  skilled human annotators. We will use the term ``instruction tuning'' to refer solely to supervised fine-tuning (SFT) on such Q\&A pairs --- and not to reinforcement-learning methods such as PPO/DPO/RLHF~\citep{schulmanWDRK17, rafailovSMMEF23} etc., which usually follow SFT in the pipeline.

Human-generated data is expensive (e.g., even the tiny model Instruct-GPT was estimated to require $20K$ human hours~\cite{oaialignapproach2022}), which has motivated the creation of open-domain alternatives. ShareGPT~\citep{chiang2023vicuna} contains conversations collected from a model-hosting website, whereas OpenAssistant~\citep{köpf2023openassistant} and Dolly~\citep{DatabricksBlog2023DollyV2} contain crowd-sourced human data. Another intriguing method, popularized by Self-Instruct~\citep{wang2023selfinstruct} (and its variants, e.g., Alpaca~\citep{taori2023alpaca}) leverages synthetic datasets. Here, a strong LLM is prompted using a small set of human-created examples to generate a large number of (query, answer) examples on a variety of topics. 

Open evaluations of instruction-following ability have also sprung up. The popular AlpacaEval 2.0~\citep{dubois2024alpacafarm, dubois2024AlpacaEval2.0} is based upon curated queries from various sources. In such evaluations, a model's response to a provided query is compared against a strong reference model's response, and the model is ranked based upon its {\em win rate} --- the percentage of queries for which the model produces a better answer than the reference model, as judged by a powerful LLM. Rankings on AlpacaEval and related benchmarks like WildBench broadly align with the human rankings of a model's performance \citep{dubois2024AlpacaEval2.0, lin2024wildbench}.

\subsection{Surprising difficulty of instruction tuning} 
A persistent puzzle in this field is that SFT on the above public datasets does {\em not} yield good performance on the evaluations. It was initially suspected this is due to a lack of diversity in the training data. But, efforts to produce more diverse synthetic data --- e.g., UltraChat~\citep{ding2023enhancing}, a synthetic dataset of $1.5$M multi-turn conversations created via meticulously tracking lexical and topical diversity as well as coherence --- did not significantly improve performance. 

Another hypothesis places the blame on the uneven quality of open datasets --- which are usually a hodge-podge of collected queries (e.g., Dolly \citep{DatabricksBlog2023DollyV2}) --- whereas proprietary datasets are produced to careful specifications using strict quality-control. One finding that supports this hypothesis is that SFT on the 1K Q\&A pairs in Alpaca-52K with the longest responses, outperforms SFT on all $52$K pairs~\citep{zhao2024long}. In other words, the $51$K other data-points are redundant, or even interfere with the ``signal'' present in the best $1$K examples. This finding has inspired ``less is more'' approaches --- including an extreme one based upon just a judicious set of in-context examples~\citep{lin2023URIAL} to provide a surprisingly reasonable level of instruction tuning and alignment --- but they did not significantly improve the performance either.

Some have cautioned against hopes for a miracle out of instruction tuning. \citet{gudibande2023false} suggest, based upon careful experiments, that basic capabilities of the LLM arise from pre-training and its massive training corpus. Most deficiencies left after pre-training will not be fixable by, say, a million SFT examples. While this perspective feels broadly correct, it does not quite explain why open efforts to instruction tune \mistral{7} fail to match the performance of its proprietary \emph{Instruct} counterpart, which has only undergone SFT.

The above difficulties have lately lowered interest level in instruction tuning, with many researchers now turning to RL-based methods (e.g., PPO, DPO), which have been  used in recent open-source projects to greatly improve proprietary chat models~\citep{meng2024simposimplepreferenceoptimization}, which had  already trained on  expensive human data.

\begin{figure}[!h]
    \centering
    \includegraphics[width=\textwidth]{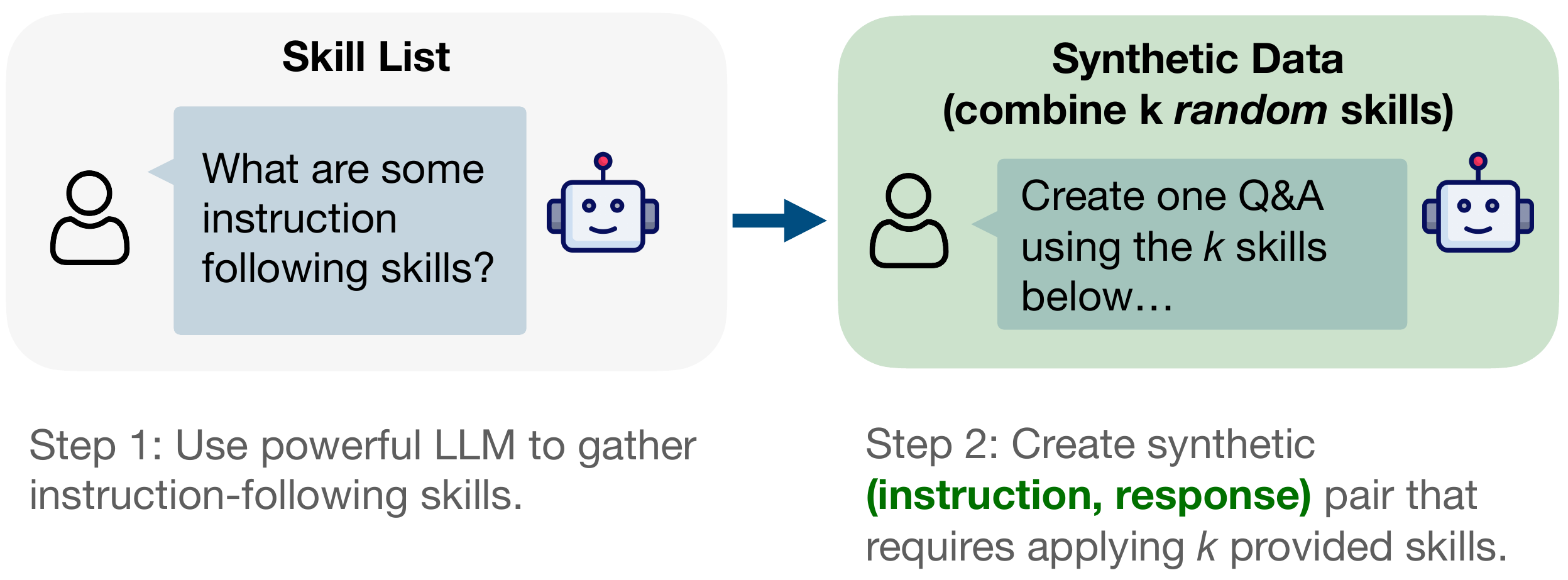}
    \caption{\textbf{Sketch of \instructSM{} pipeline.} See Figures \ref{fig:ism-sdd-pipeline} and \ref{fig:ism-sda-pipeline} for more details on two different implementations of \instructSM{}. }
    \label{fig:ism-main-pipeline}
\end{figure}

\vspace{-8pt}
\subsection{Our contributions}
We describe a more  efficient and effective approach for creating synthetic instruction tuning datasets. Past open efforts invested significant human effort in ensuring {\em high coverage} of topics and scenarios to sufficiently equip the  LLM for scenarios it might encounter at deployment time. We take a subtly different tack. Accepting that pre-training is the dominant source of the LLM's ``inner knowledge,'' we focus on merely teaching the LLM to draw upon that inner knowledge and present it nicely during conversations.

The key idea is to use a strong LLM as a teacher. The recent discovery of {\em LLM Metacognition}~\citep{didolkar2024metacognitive} suggests that frontier models have significant capability to ``think about thinking,'' which in humans is referred to as {\em metacognition}~\citep{Flavell1979MetacognitionAC}. Specifically, it was shown that given a task dataset, frontier LLMs can help assemble a list of named ``skills'' needed to solve that task. This requires no human involvement apart from an automated interaction with an LLM~\footnote{Skill lists generated by different frontier models are related but not isomorphic. Skills generated by one model are comprehensible to other models. See~\citet{didolkar2024metacognitive} for such experiments.
}.

The first phase (``{\em Skill Extraction}'') of our pipeline \instructSM{} uses this idea  and a frontier LLM  to identify a list of ``basic skills'' needed for instruction-following. Unlike~\citet{didolkar2024metacognitive}, which extracts skills from existing SFT datasets, we instead identify skills by directly prompting a strong LLM. (We also tried extracting skills using examples from Alpaca and Ultrachat, and it works quite well, but noticeably worse than our main method.)  See Section~\ref{subsec:skill-extraction}. 

The second phase of our pipeline, {\em Data Generation}, uses the list of extracted skills to produce synthetic query-response examples. Here, we repeatedly draw a random pair of skills from the list and prompt the powerful LLM to produce a suitable query that tests those two skills, and to also produce a good response to the query. This generation is inspired by the {\sc skillmix} evaluation~\citep{yu2023skillmix} for LLMs' compositional generalization, which also uses a predetermined list of skills. Hence we call our method {\sc instruct-skillmix}. See Section~\ref{subsec:data-generation}

Using merely $2$K to $4$K such Q\&A examples, vanilla SFT allows popular small base models (\mistral{7}, \llamaiii{8}, and \gemma{9}) to match or surpass some apex models on AlpacaEval 2.0, such as the original \gpt, LLaMA-3.1-405B-Instruct and Claude 3 Opus (Table~\ref{tab:main-results}). The estimated cost of creating this $4$K dataset using the \gpt API is under $600$ US dollars.

We stress that although reminiscent of prior efforts using synthetic data such as UltraChat, our pipeline is fully automated with no human design elements (e.g., choice of topics, lexicon etc.). The only human involvement involves the short prompts used for skill extraction and question generation, which we adapted from the math setting of~\cite{didolkar2024metacognitive}. While our pipeline currently focuses on simple instruction-following, the method seems extensible in future to safety/alignment, as well as domain-specific Q\&A.

\begin{table}[!h]
    \caption{\textbf{Evaluation results of \emph{base} models supervised-finetuned on \instructSM{} versus the proprietary \emph{instruct} versions and other proprietary models.} For our models, we report the results for best checkpoint selected using held-out queries. For other models(*), we report the published numbers available on publicly available leaderboards. ``\textbf{\# Data}'' refers to the number of (instruction, response) pairs in the training data. See Table \ref{tab:evaluation-results-IF-benchmarks} for a more detailed view, including comparisons to past open datasets.}
    \vspace{5pt}
    \label{tab:main-results}
    \centering
    \small 
    \begin{tabular}{l | c | c | c }
        \Xhline{2\arrayrulewidth}
        & & \multicolumn{1}{ c}{\rule{0pt}{1.2em} AlpacaEval2.0} & \multicolumn{1}{| c}{WildBench}  \\

        \textbf{Model} & \textbf{\# Data} & \textbf{LC WR(\%)} & $\textbf{WB-Reward}_\infty^{\text{gpt4t}}$ \\
        \Xhline{2\arrayrulewidth}
        
        \multicolumn{4}{l}{\rule{0pt}{1.2em}\textbf{LLaMA-3-8B}} \\
        \Xhline{2\arrayrulewidth}
        Ours & 4K & \textbf{42.76} & \textbf{-36.91} \\
        *LLaMA-3-8B-Instruct & - & 22.90 & -46.30  \\ 
        \Xhline{2\arrayrulewidth}
        
        \multicolumn{4}{l}{\rule{0pt}{1.2em}\textbf{Mistral-7B-v0.2 }} \\
        \Xhline{2\arrayrulewidth}
        Ours & 4K & \textbf{36.70} & \textbf{-29.25} \\
        SFT on Alpaca-52K & 52K & 8.64 & -80.47 \\
        *Mistral-7B-Instruct-v0.2 & - & 17.10 & -54.70  \\ 
        \Xhline{2\arrayrulewidth}
        
        \multicolumn{4}{l}{\rule{0pt}{1.2em}\textbf{Gemma-2-9B}} \\
        \Xhline{2\arrayrulewidth}
        Ours & 2K & 36.18 & -37.83 \\
        Gemma-2-9B-Instruct & - & \textbf{37.21} & \textbf{-28.78}  \\ 
        \Xhline{2\arrayrulewidth}
        
        \multicolumn{4}{l}{\rule{0pt}{1.2em}\textbf{*Other Proprietary Models }} \\
        \Xhline{2\arrayrulewidth}
        LLaMA-3.1-405B-Instruct & - & 39.30 & -  \\
        Mistral Large & - & 32.70 & -46.40  \\ 
        Claude 3 Opus & - & 40.50 & -21.20  \\
        Claude 3 Sonnet & - & 34.90 & -30.30  \\
        GPT-4-Omni (2024-05-13) & - & \textbf{57.50} & \textbf{+1.70} \\
        GPT-4 (2023-03-14) & - & 35.30 & - \\
        \Xhline{2\arrayrulewidth}
    \end{tabular}
    
\end{table}
\section{\instructSM{}}\label{sec:instruct-skillmix}

This section describes our methodology for extracting skills from powerful LLMs
\footnote{We use \gptTurbo for our main experiments (2024-04-09 checkpoint unless specified otherwise). See Appendix~\ref{appendix:diff-teacher} for results when using Claude 3.5 Sonnet (2024-06-20).} and how to use these extracted skills to create a diverse, high quality dataset. A simplified version of our pipeline and prompts are depicted in Figures~\ref{fig:ism-main-pipeline} and ~\ref{fig:ism-pipeline}. Section~\ref{sec:experiments} reports the evaluation results when finetuning on this dataset.

\subsection{Skill Extraction Procedure}\label{subsec:skill-extraction}
The method involves an automated interaction with a frontier LLM (\gptTurbo).  We ask the frontier LLM to first generate a list of topics that arise in instruction-following. For each topic returned by the LLM, we further prompt it to generate a list of skills that are needed to answer typical queries on that topic. Additionally, we ask the LLM to create a list of query types (e.g., ``Information Seeking'') that might arise in that topic. See Appendix~\ref{appendix:skill-extraction-SDA} for details about the prompts used, and Appendix~\ref{appendix:list-of-skills-SDA} for the list of all extracted skills. Since this method  relies solely upon the LLM's inner meta-knowledge, this method should extend easily to other types of instruction-following.

\paragraph{An Earlier Attempt:}
Our initial attempt to extract skills leveraged existing instruction tuning datasets, which is a more direct analog of the method in~\cite{didolkar2024metacognitive}. However, we suspected this to be sub-optimal due to known limitations of past instruction tuning datasets. We therefore designed the method described above, and found it superior. It also has scientific benefit of being independent of existing datasets like Alpaca and Ultrachat. However, the dataset from the initial method, called \instructSMSDDLong{} (\instructSMSDD; see Appendix \ref{appendix:pipeline-more-detailed}) is still very useful for  ablations that pinpoint ways in which our skill-based pipeline improves on past  synthetic datasets for instruction-following  (see Tables \ref{tab:main-results-alpaca}, \ref{tab:main-results-changing-k}, \ref{tab:main-results-gpt-version}, and \ref{tab:main-results-claude-grader}).  

\subsection{Data Generation}\label{subsec:data-generation}
Inspired by the recent {\sc skillmix} evaluation~\citep{yu2023skillmix}, we generate instruction-following examples by randomly picking $k$ skills as well as a random query type. The frontier LLM is prompted to create Q\&A pairs that illustrate these $k$ skills and the query type. We refer to the resulting dataset as \instructSM{}. For example, \instructSM{(k=2)}-1K refers to 1,000 examples of data created from random combinations of $k=2$ skills. See Appendix~\ref{appendix:synthetic-data-creation-SDD} and \ref{appendix:synthetic-data-creation-SDA} for the details about the prompts used for data generation.

See Appendix \ref{appendix:pipeline-more-detailed} for more details, and an estimate of the low cost of \instructSM{} pipeline.

\paragraph{Where does diversity come from?} 
The first source of diversity is the skill labels. A skill label represents some part of the frontier LLM's meta-knowledge of human behavior and needs, which it observed in its vast training set or during instruction tuning. Replacing a concrete Q\&A example with a skill label converts it into a pointer to a region in the frontier LLM's meta-knowledge, which the model can then freely draw upon to create new examples. The second source of diversity is the use of random $k$-tuples of skills when generating new examples. The motivation here is that, in most cases, distinct tuples will lead to very distinct flavor of examples. 

For instance, the skill pair (\texttt{critical thinking and communication}, \texttt{literature and language skills}) leads to the following instruction 
\begin{question}
    I'm a high school English teacher aiming to develop a curriculum unit for my 11th-grade class, focusing on American literature. I want this unit to go beyond just reading and understanding the texts. Specifically, I'm looking to enhance my students' critical thinking and communication skills through engaging activities related to the literature. Can you suggest detailed ways to incorporate these skills, ideally with concrete examples and expected learning outcomes?
\end{question}
whereas the skill pair (\texttt{critical thinking and communication}, \texttt{skill in virtual and system design}) leads to
\begin{question}
    As an IT manager, I am overseeing the development of a virtual workspace to enhance communication and efficiency among remote teams. This workspace must support multimedia content, including video conferencing and live document editing. What are the critical steps I should take in its design and implementation, balancing technical robustness with ease of use? Could you provide specific technologies to consider and any potential obstacles?
\end{question}
Even though the two skill pairs share a common skill, they lead to rather distinct Q\&A pairs, involving creative and nuanced situations with subtle moving parts. Since the number of $k$-tuples scales as ${N \choose k}$, where $N$ is the number of skills, using pairs of skills foster a lot of diversity --- e.g., $125,000$ possibilities with $N=500, k=2$. The pipeline in our experiments mainly uses $k=2$, but generating answers to these queries will certainly end up using many other unnamed skills as well, and thus serve as a rich source for learning how to follow instructions.
\section{Experiments} \label{sec:experiments}

\subsection{Experimental Setup}

\paragraph{SFT on \instructSM{(k)}.}
We finetune \llamaiii{8}, \mistral{7}, \gemma{9}, \llamaii-7B-Base, and \llamaii-13B-Base on a varying number of examples from \instructSMSDD{(k)} and \instructSMSDA{(k)}. We train for multiple epochs and select the best checkpoint by performance on $100$ held-out questions. Similar to \citet{ouyang2022training, zhou2023lima}, we observe that using cross-entropy loss on a validation set does not lead to the best checkpoint. See Appendix~\ref{appendix:checkpoint-selection} for a more detailed discussion of the checkpoint selection procedure. As a baseline, we also finetune on different subsets of Alpaca-52K, including the 1K or 5K examples with the longest completions. For further training details (e.g., hyperparameters), see Appendix~\ref{appendix:hyperparams}.

\paragraph{Evaluation.}
We evaluate our models on popular instruction following benchmarks: AlpacaEval 2.0~\citep{dubois2024AlpacaEval2.0}, MT-Bench~\citep{zheng2023judging}, and WildBench~\citep{lin2024wildbench}. For AlpacaEval, we report the length-controlled win rate (LC WR) of the responses of our model against a reference response, which corrects for the length bias of the judge model. For MT-Bench, we report the average score of the responses of our model graded by a judge model. For WildBench, we report the WB-Reward (weighted win-rate) of the response of our model against one reference response when graded by a judge model. For further evaluation details, see Appendix~\ref{appendix:evaluation-details}. See Table~\ref{tab:evaluation-results-LLM-benchmarks} in Appendix~\ref{appendix:evaluation-results} for evaluations on additional benchmarks.
\subsection{Main Results}
For the main results of the paper, we report the evaluation results when models are finetuned on \instructSMSDA{} in Table \ref{tab:main-results}, and summarize our findings below. For a more detailed version of Table \ref{tab:main-results}, see Table~\ref{tab:evaluation-results-IF-benchmarks}. For additional ablations, see Appendix~\ref{appendix:ablations}. For evaluations on other LLM benchmarks, see Table~\ref{tab:evaluation-results-LLM-benchmarks}. 

\paragraph{\instructSMSDA{} achieves SOTA performance amongst SFT models.}
\llamaiii{8} finetuned on 4K examples from \instructSMSDA{(k=2)} yields LC win rate of 42.76\% on AlpacaEval 2.0. This score is higher than Claude 3 Opus, LLaMA-3.1-405B-Instruct, and \gptMarch. \mistral{7} finetuned on the same data achieves -29.25 on WildBench, which outperforms Claude 3 Sonnet and Mistral Large. \gemma{9} finetuned on 2K examples from \instructSMSDA{(k=2)} gets a score of 8.12 on MT-Bench, which is better than GPT-3.5-Turbo (2023-03-01). To best of our knowledge, these scores are higher than any base model that has \emph{only} undergone supervised instruction finetuning (i.e., no RLHF, DPO, PPO, or variants).

\paragraph{Early saturation.}
Performance from our method rises rapidly, reaching unprecedented levels with 1K examples. Unfortunately, improvements stop already with 4K examples. This turns out to be a consequence of its high efficiency at inducing good instruction-following. Specifically, with 4K examples, the win-rate against \gpt approaches $50\%$ on {\em heldout} queries from our pipeline, and thus overfitting sets in.

\paragraph{Observed limitations.}
The open benchmarks used in this study have known limitations, related to the insufficient number of under-specified or ambiguous queries, and no testing of long-form generations such as multi-page essays. Our current pipeline shares some of these limitations. Fixing this seems very doable via suitable modification to our \instructSM{} pipeline, but this is left for future work. This aligns with the observation in \citet{bai2024longwriterunleashing10000word} that a model's effective generation length seems to be limited by the typical length of examples seen during SFT, and is exacerbated by the relative scarcity of long-form samples in the SFT data. This underscores the critical influence of training data composition on a model's post-fine-tuning capabilities, and would be interesting to investigate in future work.

\section{Ablation study} \label{sec:ablations}
Whereas pretraining is the source of an LLM's basic capabilities \citep{gudibande2023false}, the sole goal of instruction tuning is to impart skills,
such as answer-structuring, empathy, helpfulness, etc. 

Vanilla SFT on Q\&A data generated by a teacher LLM is akin to  {\em imitation learning.}  Our ablation studies below  help understand the contribution of different elements to the effectiveness of  imitation learning  method using \instructSM{} Q\&A. The main finding is that the source of largest improvement is the skill extraction step. 

\subsection{Benefits of Skill Extraction (with mixing turned off)}
To highlight the benefits of our skill-based method versus current synthetic approaches, we use the pioneering Alpaca dataset, whose responses are rewritten by GPT-4 (2023-03-14) \citep{peng2023AlpacaGPT4}. The fairest comparison here would be with our \instructSMSDD{(k=1)} data, where the underlying skills were derived from a random sample of \textit{Alpaca-52K}, and each of our datapoints uses one of those extracted skills. All results below involve finetuning \mistral{7} on different subsets of the Alpaca-52K dataset: 
(1) \textit{Alpaca-1K Longest}: 1,000 examples with the longest responses \citep{zhao2024long}; 
(2) \textit{Alpaca-5K Longest}: 5,000 examples with the longest responses;
(3) \textit{Alpaca-5K Random}: 5,200 randomly sampled examples from which we extracted our skills; and 
(4) \textit{Alpaca-52K}: the full 52,002 examples.

\begin{table}[!h]
    \caption{\textbf{Evaluation results of \mistral{7} finetuned on \instructSMSDD{} vs. on Alpaca-52K. } Note that skills extracted from Alpaca-5K Random were used to create the \instructSMSDD{} datasets.}
    \vspace{5pt}
    \label{tab:main-results-alpaca}
    \centering
    \small 
    \begin{tabular}{l | c | c | c | c }
        \Xhline{2\arrayrulewidth}
        & & \multicolumn{1}{ c}{\rule{0pt}{1.2em} AlpacaEval 2.0} & \multicolumn{1}{| c}{\rule{0pt}{1.2em} }  & \multicolumn{1}{| c}{WildBench}  \\
        \textbf{SFT Dataset} & \textbf{\# Data} & \textbf{LC WR(\%)} & \textbf{MT-Bench} & $\textbf{WB-Reward}_\infty^{\text{gpt4t}}$ \\
        \Xhline{2\arrayrulewidth}
        \instructSMSDD{(k=2)} & 4K & \textbf{29.77} & 7.17 & \textbf{-39.06}\\
        \instructSMSDD{(k=1)} & 1K & 27.04 & \textbf{7.22} & -46.83 \\
        Alpaca-1K Longest & 1K & 10.09 & 6.88 & -63.38 \\
        Alpaca-5K Longest & 5K & 8.92 & 6.90 & -62.55 \\
        Alpaca-5K Random  & 5K & 11.10 & 6.86 & -74.41 \\
        Alpaca-52K Full & 52K & 8.64 & 6.45 & -80.47 \\
        \Xhline{2\arrayrulewidth}
    \end{tabular}
\end{table}

As shown in Table~\ref{tab:main-results-alpaca}, finetuning on 1,000 examples with the longest completions from Alpaca-52K yields 10.09\% LC win rate on AlpacaEval 2.0. On the other hand, finetuning on only 1K examples of \instructSMSDD{(k=1)} yields 27.04\% LC win rate. Note that since the skills in \instructSMSDD{} are mostly derived from Alpaca-52K, the observed improvements in the win rate are indicative of the improved quality of \instructSMSDD{} queries.

\subsection{Mixing skills helps, but not as much as skill extraction}
In Table~\ref{tab:main-results-changing-k}, models finetuned on \instructSMSDD{(k=2)} data marginally outperform models SFT on \instructSMSDD{(k=1)} on AlpacaEval and WildBench, whereas performance on MT-bench is about the same. The marginal improvements from increasing $k$ are less noticeable for \instructSMSDA{}. 

\begin{table}[!h]
    \caption{\textbf{Evaluation results of \mistral{7} SFT on \instructSM{} where k=1 vs. k=2.} In each entry, we report \ca{\instructSMSDD{}}/\cc{\instructSMSDA{}}}
    \vspace{5pt}
    \label{tab:main-results-changing-k}
    \centering
    \small
    \begin{tabular}{l | c | c  c | c | c }
        \Xhline{2\arrayrulewidth}
        & & \multicolumn{2}{ c}{\rule{0pt}{1.2em} AlpacaEval 2.0} & \multicolumn{1}{| c}{\rule{0pt}{1.2em} }  & \multicolumn{1}{| c}{WildBench}  \\

        \textbf{Model} & \textbf{\# Data} &\textbf{WR(\%)} & \textbf{LC WR(\%)} & \textbf{MT-Bench} & $\textbf{WB-Reward}_\infty^{\text{gpt4t}}$ \\
        \Xhline{2\arrayrulewidth}
        
        \multicolumn{6}{l}{\rule{0pt}{1.2em}\textbf{SFT on \instructSM{(k=2)}}} \\
        \Xhline{2\arrayrulewidth}
        \multirow{3}{*}{\mistral{7}} & 1K & \res{33.87}{42.48} & \res{27.48}{38.34} & \res{6.92}{7.33} & \res{-41.46}{-30.65} \\
        & 2K & \res{37.05}{40.83} & \res{31.57}{36.18} & \res{7.04}{7.20} & \res{-43.46}{-31.92} \\
        & 4K & \res{35.08}{40.74} & \res{29.77}{36.70} & \res{7.17}{7.16} & \res{-39.06}{\textbf{-29.25}} \\
        \Xhline{2\arrayrulewidth}
        
        \multicolumn{6}{l}{\rule{0pt}{1.2em}\textbf{SFT on \instructSM{(k=1)}}} \\
        \Xhline{2\arrayrulewidth}
        \multirow{3}{*}{\mistral{7}} & 1K & \res{30.06}{41.75} & \res{27.04}{38.34} & \res{7.22}{7.49} & \res{-46.83}{-30.95} \\
         & 2K & \res{35.07}{-} & \res{31.66}{-} & \res{7.39}{-} & \res{-46.97}{-} \\
         & 4K & \res{33.57}{-} & \res{28.85}{-} & \res{7.13}{-} & \res{-44.43}{-} \\
        \Xhline{2\arrayrulewidth}
    \end{tabular}
\end{table}

\subsection{Quality of Queries (and Skills) Matters}\label{subsec:imitation-learning}

The effectiveness of this approach depends on the quality of the queries used in the fine-tuning process, where high-quality queries enable the frontier LLM teacher to provide richer instruction to the student model undergoing instruction tuning. This relationship between the quality of queries and the skills being imparted is supported by two key observations. First, the frontier LLM proves to be a more effective teacher when the skill list being used was also entirely generated using its help (as opposed to giving it skills derived from existing datasets).\footnote{We also observed improved performance when the teacher model generated data based on its own set of skills, rather than using skills extracted by a different teacher model, further highlighting the advantages of leveraging the teacher model's metacognitive capabilities during dataset creation (see Appendix~\ref{appendix:mix-teacher}).} Across all model types, dataset size, and the evaluation benchmark, we generally see an improvement when finetuning on \instructSMSDA{} compared to \instructSMSDD{} (see Table~\ref{tab:evaluation-results-IF-benchmarks} for more details). Second, incorporating these sub-optimal skills from existing datasets as a part of ``teaching'' (e.g., with \instructSMSDD{}) is still more effective than using an equal number of random (or even the longest) examples from Alpaca-52K when responses are also by the same frontier LLM. These findings suggest that the quality of the queries (and the skills used to create those queries) drives how well data generated by the frontier LLM is able to impart its skills on the model undergoing instruction tuning.
\subsection{Effect of Teacher and Grader }\label{subsec:confounding-factors}

SFT performance derives from the  model used to generate Q\&A data, which plays the {\em teacher} role in imitation learning. The student's performance is evaluated by the grader model. The main results  reported in this paper used \gptTurbo as the teacher, and some checkpoint of \gpt or \gptTurbo as the grader. 

\paragraph{Effect of the teacher}
Many SFT efforts in 2023 used  earlier versions of \gpt or GPT-3.5, which were weaker than \gptTurbo. To pin-point the effect of this change, we try doing a head-to-head comparison once we fix the teacher. The responses in Alpaca-1K Longest are written by GPT-4 (2023-03-14), whereas \instructSM{} data is generated by \gptTurbo. Thus, we use \gptTurbo to regenerate answers to Alpaca-1K Longest ~\citep{zhao2024long}, and we also use GPT-4 (2023-03-14) to regenerate \instructSMSDD{}.

\begin{table}[!h]
    \caption{\small \textbf{Evaluation results of \mistral{7} finetuned on \instructSMSDD vs. Alpaca-1K Longest generated from two different versions of \gpt.} For a fixed data generator model, SFT \mistral{7} on \instructSMSDD{} outperforms SFT on Alpaca-1K Longest.}
    \vspace{5pt}
    \label{tab:main-results-gpt-version}
    \centering
    \small 
    \begin{tabular}{l | l | c  c | c}
        \Xhline{2\arrayrulewidth}
        & & \multicolumn{2}{c}{\rule{0pt}{1.2em} AlpacaEval 2.0} & \multicolumn{1}{| c}{\rule{0pt}{1.2em}} \\
        \textbf{Model for Data Generation} & \textbf{Dataset} & \textbf{WR(\%)} & \textbf{LC WR(\%)} & \textbf{MT-Bench} \\
        \Xhline{2\arrayrulewidth}
        \multirow{2}{*}{GPT-4 (2023-03-14)} & Alpaca-1K Longest & 12.75 & 10.09 & 6.83 \\
        & \instructSMSDD{}-1K & 13.29 & 15.01 & 7.10 \\
        \Xhline{2\arrayrulewidth}
        \multirow{2}{*}{GPT-4-Turbo (2024-04-09)} & Alpaca-1K Longest & 35.23 & 19.62 & 6.99 \\
        & \instructSMSDD{}-1K & 33.87 & 27.48 & 6.92 \\
        \Xhline{2\arrayrulewidth}
    \end{tabular}
    
\end{table}

Table~\ref{tab:main-results-gpt-version} compares the performance of \mistral{7} when finetuned on the two datasets using the two  versions of \gpt. For each fixed data generator model, the \instructSM{} dataset leads to a better performance. Furthermore, replacing \gpt with the stronger \gptTurbo in data generation makes \instructSM{} pull even further ahead of Alpaca-1K Longest, which highlights that our pipeline is better positioned than Alpaca dataset to elicit better supervision from a more powerful LLM teacher.

\paragraph{Effect of choice of grader}
We use \gptTurbo to generate data and AlpacaEval 2.0 uses \gpt for grading, creating a scenario where both the teacher model and grader model are from the same family. This raises the question of whether model family overlap leads to a potential grading bias and inflated scores. To quantify this effect, we used Claude 3 Opus as the grader for AlpacaEval 2.0. Table~\ref{tab:main-results-claude-grader} shows that although Claude is a more generous grader across the board, it generally preserves the relative rankings among the models. Importantly, it exhibits even stronger preference for our student models' generations than does \gpt. 

\begin{table}[!h]
    \caption{\small \textbf{Evaluation results when using two different graders for AlpacaEval 2.0.} Relative ranking of evaluated models are generally preserved when using different graders. Here, ISM-D refers to \instructSMSDD{}.}
    \vspace{5pt}
    \label{tab:main-results-claude-grader}
    \centering
    \small 
    \begin{tabular}{l | c  c | c  c  }
        \Xhline{2\arrayrulewidth}
        & \multicolumn{2}{c}{\rule{0pt}{1.2em} Grader: GPT-4 (2023-11-06)} & \multicolumn{2}{| c}{\rule{0pt}{1.2em} Grader: Claude 3 Opus} \\
        \textbf{Model} & \textbf{WR(\%)} & \textbf{LC WR(\%)} & \textbf{WR(\%)} & \textbf{LC WR(\%)} \\
        \Xhline{2\arrayrulewidth}
        \mistral{7} SFT on ISM-D-1K & 33.87 & 27.48 & 50.56 & 38.50 \\
        \mistral{7} SFT on ISM-D-2K & 37.05 & 31.57 & 48.94 & 38.29 \\
        \mistral{7} SFT on ISM-D-4K & 35.08 & 29.77 & 52.55 & 44.16 \\
        \Xhline{2\arrayrulewidth}
        (Reference Model) LLaMA-3-70B-Instruct & 33.20 & 34.40 & 39.68 & 42.33 \\
        (Reference Model) Mistral-7B-Instruct-v0.2 & 14.70 & 17.10 & 15.16 & 18.89 \\
        (Reference Model) LLaMA-2-70B-Chat & 13.90 & 14.70 & 16.67 & 17.85 \\
        \Xhline{2\arrayrulewidth}
    \end{tabular}
\end{table}
\section{Effect of Low Quality Data}\label{subsec:neg-pareto-princ}
Our fully synthetic pipeline produces a large number of high-quality questions and answers that look impressive but also (for want of a better word) ``robotic.'' Data sourced from human workers shows greater variation, and one begins to wonder if that additional  diversity could be beneficial. We tried interventions such as generating $20\%$ using a different prompt --- e.g., require a shorter answer, or a poor quality answer. In a human pipeline, this variation would be expected.  We can think of this as ``data from shirkers,'' and one would expect a fair bit of it in naive crowdsourcing. (In corporate settings it would be mitigated via quality control measures.) See Appendix \ref{appendix:neg-pareto-princ} for an example of a poor quality response.

We replace 20\% of the responses in \instructSM{(k=2)}-2K with short responses (``respond in one paragraph'') to create {\sc \small BREV-}\instructSM{(k=2)}-2K. Finetuning \mistral{7} on {\sc \small BREV-}\instructSMSDD{}  was surprising: brevity constraint on just $20\%$ of data almost halved the average response length on AlpacaEval, from 2817 to 1746 characters. LC win rate dropped from 31.57\% to 23.93\%.

We alternatively replace 20\% of the responses in the same datasets with responses that are still long but have poor quality (i.e., deliberately sloppy and unhelpful) to create \textsc{Junk-}\instructSM{(k=2)}-2K. \mistral{7} finetuned on the \textsc{Junk-}\instructSMSDD{} yields less than 1\% win rate on AlpacaEval and 5.01 on MT-Bench.

\begin{table}[!h]
    \caption{\textbf{Evaluation results of models finetuned on low quality \instructSM{}.} Replacing just 20\% of the dataset with low quality data has a  super-proportionate harm on the model performance. Amount of harm greatly differs between the two versions of the pipeline.}
    \vspace{5pt}
    \label{tab:main-results-brevity}
    \centering
    \small 
    \begin{tabular}{l | c | c c | c | c }
        \Xhline{2\arrayrulewidth}
        & & \multicolumn{2}{c}{\rule{0pt}{1.2em} AlpacaEval 2.0} & \multicolumn{1}{| c}{\rule{0pt}{1.2em} }  & \multicolumn{1}{| c}{WildBench}  \\
        \textbf{Model} & \textbf{\# Data} & \textbf{LC WR(\%)} & \textbf{Avg Len} & \textbf{MT-Bench} & $\textbf{WB-Reward}_\infty^{\text{gpt4t}}$ \\
        \Xhline{2\arrayrulewidth}
        \multicolumn{5}{l}{\rule{0pt}{1.2em}\textbf{SFT on \instructSMSDD{(k=2)}}} \\
        \Xhline{2\arrayrulewidth}
        \multirow{3}{*}{\mistral{7}} & 2K & 31.57 & 2817 & 7.04 & -43.46 \\
        & 2K (Brevity 20\%) & 23.93 & 1746 & 6.69 & -49.85 \\
        & 2K (Junk 20\%) & 0.77 & 1104 & 5.01 & -47.50 \\
        \Xhline{2\arrayrulewidth}
        \multicolumn{5}{l}{\rule{0pt}{1.2em}\textbf{SFT on \instructSMSDA{(k=2)}}} \\
        \Xhline{2\arrayrulewidth}
        \multirow{3}{*}{\mistral{7}} & 2K & 36.18 & 2936 & 7.20 & -31.92 \\
        & 2K (Brevity 20\%) & 31.61 & 2336 & 7.32 & -32.27 \\
        & 2K (Junk 20\%) & 24.60 & 2435 & 6.90 & -47.50 \\
        \Xhline{2\arrayrulewidth}
    \end{tabular}
\end{table}

\paragraph{Lower-quality data harms performance.}
As shown in Table~\ref{tab:main-results-brevity}, replacing just 20\% of the data with poor quality responses harms performance. For \instructSMSDD{}, the harm is super-proportionate. These observation may help explain why creating open-domain instruction tuning data has proved so difficult via naive crowd-sourcing. 

\paragraph{High-quality data's protective effect.}
While adding some low-quality data to \instructSMSDA{} already causes a noticeable performance drop, doing the same to \instructSMSDD{} is catastrophic. This suggests that \instructSMSDA{} is more robust to ``shirkers,'' corroborating our previous observations in Table~\ref{tab:evaluation-results-IF-benchmarks} of the superior performance of \instructSMSDA{} over \instructSMSDD{}. This finding  suggests that higher quality data can somewhat protect against negative effects of ``shirkers,'' which needs further study.
\section{Related Work}
Prior works observe improvements from instruction finetuning on \emph{fewer}, but \emph{higher quality} data generated by humans \citep{zhou2023lima, touvron2023llama}. However, efforts to curate high quality data from humans are quite expensive, and licensing can become complicated. This has led to an increase in the popularity of semi-automated and less expensive approaches.

\paragraph{Selecting high quality data.}
Synthetic data creation has become a predominant approach for curating instruction tuning datasets, especially in the academic realm \citep{wang2023selfinstruct, dubois2024alpacafarm, xu2023wizardlm, gunasekar2023textbooks}. These synthetic datasets are generally created by providing in-context examples to a powerful LLM to produce the synthetic data, followed by some post-filtering \citep{wang2023selfinstruct}. Recent efforts have also focused on data selection strategies for high quality subsets of the original dataset, which lead to performance gains \citep{tunstall2023zephyr, chen2024alpagasus, liu2024makes, zhao2024long}. Notably, \citet{zhao2024long} show that finetuning on just the 1K longest completions from Alpaca-52K outperforms finetuning on the entire Alpaca-52K dataset. Whereas the data selection methods just described focus on \emph{general-purpose} instruction tuning, \citet{xia2024less} explore an optimizer-aware data selection strategy
for \emph{targeted} instruction tuning.

\paragraph{Encouraging data diversity.}
Common approaches to elicit diversity in datasets include mixing multiple datasets~\citep{wang2022supernaturalinstructions, longpre2023flan, wang2023far}, as well as rewriting the data in multiple ways and changing formatting~\citep{allenzhu2024physics, honovich2022unnatural}. The Self-Instruct framework \citep{wang2023selfinstruct} and variants such as Alpaca-52K~\citep{dubois2024alpacafarm} encourage diversity by  identifying similar pairs using ROUGE-L similarity.
Other approaches to ensure diversity
impose constraints on the topic in order to enhance wide coverage~\citep{ding2023enhancing, xu2023wizardlm}, or require synthetic data to use a random subset of words or concepts chosen from some vocabulary~\citep{eldan2023tinystories, gunasekar2023textbooks, li2024syntheticdataalmostscratch}. The latter approach is also suggested by recent work that provides a mathematical model for emergence via LLM scaling \citep{arora2023theory} and used in the evaluation setting in \citet{yu2023skillmix}.

\paragraph{AlpacaEval.}
AlpacaEval~\citep{li2023AlpacaEval, dubois2024AlpacaEval2.0} is a popular evaluation for assessing  instruction-following capabilities of LLMs. The tested model provides answers  805 carefully curated instructions, and its answers are compared against  reference outputs of a designated baseline model. For each instruction, another evaluator LLM outputs a preference between the two responses (output of the model being evaluated vs. reference output by the baseline mode). The primary evaluation metric is the \emph{win rate}, which represents the expected probability that the grader model favors the response generated by the evaluated model over the response produced by the baseline model. Given that a raw win rate shows bias towards longer responses, AlpacaEval 2.0~\citep{dubois2024AlpacaEval2.0} introduces the \emph{length-corrected (LC) win rate} as a proxy for what the raw win rate would be if the evaluated model's response lengths and baseline model's response lengths matched. 

\paragraph{WildBench.}
WildBench~\citep{lin2024wildbench} is another benchmark for assessing the instruction following capabilities of LLMs. Unlike the AlpacaEval instructions, 50\% of which are only ``information seeking'' type questions, the instructions for WildBench cover a more diverse distribution of task categories, including coding and creative writing. 
Whereas the grading in AlpacaEval is more liberal (since there is no penalty for poor responses), the grading in WildBench is more finegrained: a model answer is compared against a reference answer, but is graded on a scale of (1) win by a big margin, (2) win by a small margin, (3) tie, (4) lose by a small margin, and (5) lose by a big margin. This ensures that models that output bad answers to certain types of questions are penalized.

\paragraph{RL-inspired approaches.}
Since we do not use RL, we defer discussion of these approaches to Appendix \ref{appendix:comparison-to-rl}.
\section{Conclusion}
While one would have certainly expected the cost factor as well as scaling ability to ultimately favor synthetic data, the surprising finding in this paper is that, when done well, synthetic data can be much more {\em effective} than human data for instruction tuning. Our \instructSM pipeline,  uses the recent discovery of LLM Metacognition~\citep{didolkar2024metacognitive} to extract skills using a powerful LLM and then leverages an LLM to create quality instruction data using random pairs of those skills. 

Vanilla SFT of base models on just 1K to 4K examples from our pipeline outperforms the proprietary {\em instruct} versions of the same model, as well as older and larger instruction tuning efforts like Vicuna and Ultrachat that used orders of magnitude more datapoints. The performance also approaches those of frontier models, which trained on expensive human data as well as with RL techniques. Unfortunately, our method saturates at 4K examples, when win-rate  on heldout queries approaches $50\%$.   

Ablation studies in Section~\ref{subsec:confounding-factors} rule out potential confounding factors, such as the use of a strong teacher, or bias due to teacher and grader belonging to the same family. These ablations reinforce that the improvement is primarily due to the uniformly high quality of examples produced by our skill-based pipeline. Each example contains a query with nontrivial scenarios and lots of moving parts, which improve imitation learning. 
 
Section~\ref{subsec:neg-pareto-princ}  offers a preliminary exploration of pitfalls of  naive collection of instruction tuning data. In particular, the presence of some lower quality data noticeably harms the model's performance. This insight should be more rigorously investigated, including via new theory. The experiment also suggests that our less preferred  \instructSMSDD{} method (which involves  extracting skills from an existing dataset) is more susceptible to such bad data than our preferred \instructSMSDA{}.  

One potential benefit of \instructSMSDD{} may be that it gives some insights into an efficient method for dataset distillation~\citep{wang2020datasetdistillation} for text datasets, which has not yet proved possible. 

Finally, it should be  noted that our results look stronger on paper than they actually are. Open evaluations such as AlpacaEval 2.0 have blind spots, especially the fact that win rate of even $50\%$ against a frontier model still allows  unacceptably high frequency of unsuitable responses in a deployment setting. The new WildBench evaluation does test for more corner cases. We hope that \instructSM ideas can also leverage LLM metacognition to create a better evaluation.

Although our SFT data does not address safety and alignment, our skill-based ideas may be useful there.  A related next step would be to leverage our ideas of skill extraction to improve RL-based methods (whether for instruction-following or alignment). We hope to address these in future work.

\clearpage
\section{Reproducibility Statement}
We provide the full lists of extracted skills, topics, and query types in Appendix~\ref{appendix:list-of-skills}. We provide the set of prompts used to generate the data from these lists in Appendix~\ref{appendix:synthetic-data-creation-SDD} and \ref{appendix:synthetic-data-creation-SDD}. We provide the set of training hyperparameters in Appendix~\ref{appendix:hyperparams}. We discuss the details of the checkpoint selection method in Appendix~\ref{appendix:checkpoint-selection}. We provide the details of evaluation settings in Appendix~\ref{appendix:evaluation-details}.
\section{Acknowledgements}
SK, SP, and SA acknowledge funding from NSF, DARPA, ONR, and OpenAI.
\clearpage

\bibliography{references}
\bibliographystyle{plainnat}
\clearpage

\appendix
\clearpage
\appendix

\listofappendices
\clearpage

\section{\instructSM{} Pipeline (More Details)}\label{appendix:pipeline-more-detailed}
\appendixtoc{\instructSM{} Pipeline (More Details)}

\subsection{\instructSMSDD{} and \instructSMSDA{} Pipelines}
\paragraph{Method 1: Leveraging existing instruction datasets.}
Even though existing instruction-following datasets may not induce good chat capability via vanilla SFT, these datasets still exhibit (possibly in an uneven fashion) some ``skills'' needed by the model. Thus, we adapt the methodology presented in \citet{didolkar2024metacognitive} and use \gptTurbo to extract instruction-following skills from random samples of existing instruction and alignment datasets (5,200 samples from Alpaca-52K and 1,000 samples from UltraChat). We then use \gptTurbo to cluster similar skills into broader categories, forming our final list of instruction-following skills. See Appendix~\ref{appendix:list-of-skills-SDD} for the list of all extracted skills and Appendix~\ref{appendix:skill-extraction-SDD} and \ref{appendix:semantic-clustering-SDD} for details about the prompts used for skill extraction.

\paragraph{Method 2: Directly prompting a powerful LLM.}
While Method 1 works surprisingly well, it generated unease about possibly relying on existing seed datasets of uneven quality, and thus potentially inheriting their limitations and biases.  Therefore we also tried an alternative pipeline that solely relies on the powerful LLM's ideas about list of skills it leverages for instruction-tuning.

We will refer to the datasets generated from the seed-dataset dependent and the seed-dataset agnostic versions as \instructSMSDDLong{} and \instructSMSDALong{}, respectively. Unless stated otherwise, \instructSM{} refers to the \instructSMSDALong{} data.

\begin{figure}[h]
    \centering
    \begin{subfigure}[b]{0.9\textwidth}
        \centering
        \includegraphics[width=\textwidth]{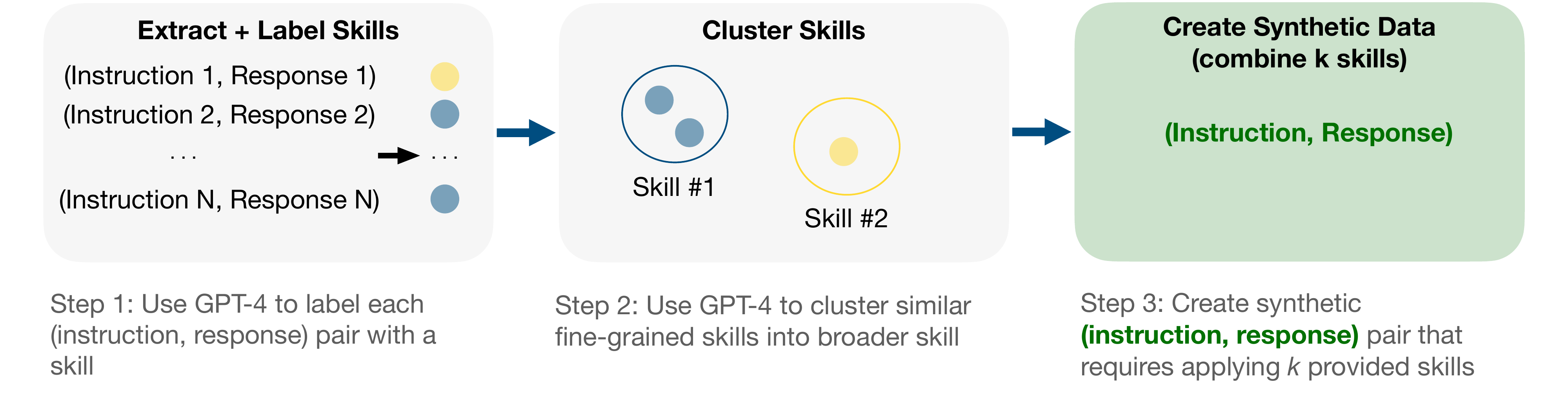}
        \caption{Top row: \instructSMSDD{} pipeline (short for \instructSMSDDLong{}).}
        \label{fig:ism-sdd-pipeline}
    \end{subfigure}
    \begin{subfigure}[b]{0.9\textwidth}
        \centering
        \includegraphics[width=\textwidth]{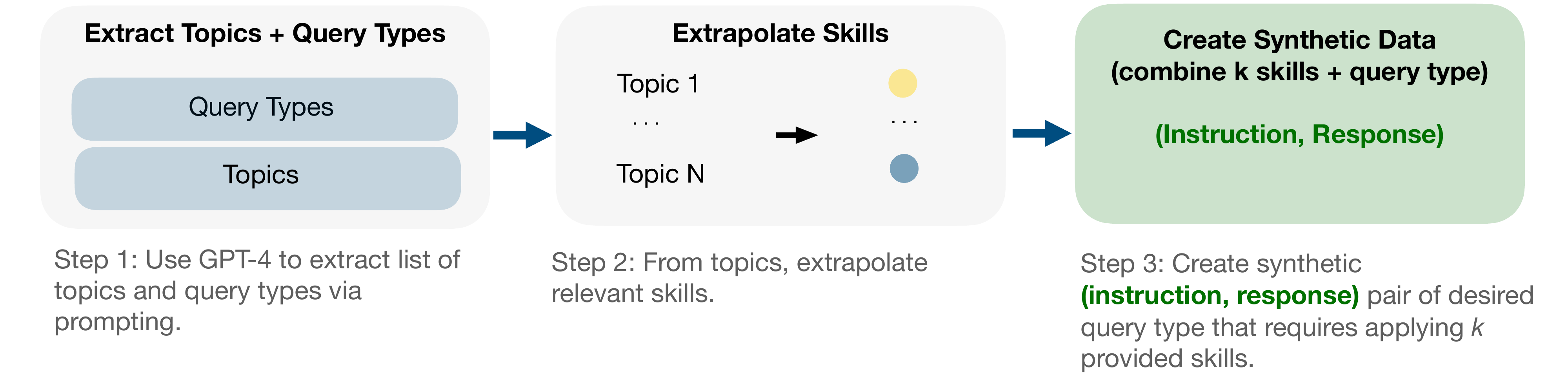}
        \caption{Bottom row: \instructSMSDA{} pipeline .}
        \label{fig:ism-sda-pipeline}
    \end{subfigure}
    \caption{\textbf{Two variants of the \instructSM{} pipeline.} 
    \instructSM{(k)} involves two steps: (1) skill extraction using similar ideas as~\cite{didolkar2024metacognitive}; (2) data generation from random $k$-tuples of skills.}
    \label{fig:ism-pipeline}
\end{figure}

\subsection{Dataset Curation Costs}
Generating synthetic data using the \instructSM{} pipeline is more cost effective compared to using human annotators.
To extract the skill clusters for \instructSMSDD{},
it costs less than \$120 to extract and cluster skills from 6,2000 examples from various existing datasets.
For \instructSMSDA{}, extracting skills via direct prompting costs under \$5.
Additionally, producing 4,000 examples of \instructSM{(k=2)} data costs under \$570.
\clearpage

\section{\instructSM{} with a Different Teacher Model}\label{appendix:diff-teacher}
\appendixtoc{\instructSM{} with a Different Teacher Model}

We apply \instructSMSDA{} with Claude-3.5-Sonnet (2024-06-20) as the teacher model and replicate some of the experiments from the main paper. See Tables~\ref{tab:evaluation-results-IF-benchmarks-claude}, \ref{tab:evaluation-results-mix-teacher} for the results. We report the results for best checkpoint selected using held-out queries. 

\subsection{\instructSM{} is applicable with any strong teacher model}
We observe that Claude-3.5-Sonnet is also able to generate a meaningful list of query types, topics, and fine-grained skills. See Appendix~\ref{appendix:list-of-skills-SDA-claude} for the full list. When compared to the list generated by \gptTurbo (Appendix~\ref{appendix:list-of-skills-SDA}), we see that Claude-3.5-Sonnet generates a very similar list of query types (e.g., ``Information-seeking'' and ``Help-seeking'' are the first two entries generated from both models), but the description and example queries from each query type are more terse. On the other hand, the topics and skills generated by Claude-3.5-Sonnet are more fine-grained and specific than those of \gptTurbo. 

Claude-3.5-Sonnet is also able to generate (instruction, response) pairs from randomly selected pair of skills and a random choice of query type. Upon manual inspection, we observe that the data generated by Claude-3.5-Sonnet is slightly less illustrative than the \instructSM{} data generated by \gptTurbo. 

\subsection{\instructSM{} outperforms other methods}
Once we fix the teacher model as Claude-3.5-Sonnet, the conclusion remains the same from the main paper: \instructSMSDA{} outperforms regenerating responses to existing datasets. See Table~\ref{tab:evaluation-results-IF-benchmarks-claude}.

\begin{table}[!h]
    \caption{\textbf{Evaluation results on AlpacaEval 2.0 and MT-Bench.} ``\textbf{\# Data}'' refers to the number of (instruction, response) pairs in the training data.}
    \vspace{5pt}
    \label{tab:evaluation-results-IF-benchmarks-claude}
    \centering
    \small 
    \begin{tabular}{l | c | c  c | c }
        \Xhline{2\arrayrulewidth}
        & & \multicolumn{2}{ c}{\rule{0pt}{1.2em} AlpacaEval 2.0} & \multicolumn{1}{| c}{\rule{0pt}{1.2em} } \\

        \textbf{Dataset} & \textbf{\# Data} &\textbf{WR(\%)} & \textbf{LC WR(\%)} & \textbf{MT-Bench} \\
        \Xhline{2\arrayrulewidth}
        \multicolumn{5}{l}{\rule{0pt}{1.2em}\textbf{SFT \mistral{7} on data generated by Claude-3.5-Sonnet (2024-06-20)}} \\
        \Xhline{2\arrayrulewidth}
        \instructSMSDA{(k=2)} & 1K & \textbf{25.74} & \textbf{25.54} & 6.88  \\
        Alpaca-52K & Long 1K & 22.10 & 19.12 & \textbf{7.13} \\
        ShareGPT & Random 1K & 21.00 & 19.77 & 7.06 \\
        \Xhline{2\arrayrulewidth}
    \end{tabular}
\end{table}

\subsection{Extracting skills with one teacher and generating with another}
\label{appendix:mix-teacher}
We ask Claude-3.5-Sonnet to generate \instructSMSDA{(k=2)}-1K from the query types and skills generated by \gptTurbo, and vice versa. For any fixed choice of teacher model, performance is slightly better when it generates \instructSMSDA{(k=2)} data from the query types and skills it extracted. See Table~\ref{tab:evaluation-results-mix-teacher}.

\begin{table}[!h]
    \caption{\textbf{Evaluation results on AlpacaEval 2.0 and MT-Bench.}}
    \vspace{5pt}
    \label{tab:evaluation-results-mix-teacher}
    \centering
    \small 
    \begin{tabular}{l | c | c  c | c }
        \Xhline{2\arrayrulewidth}
        & & \multicolumn{2}{ c}{\rule{0pt}{1.2em} AlpacaEval 2.0} & \multicolumn{1}{| c}{\rule{0pt}{1.2em} } \\

        \textbf{Data Generated by} & \textbf{Skills From} &\textbf{WR(\%)} & \textbf{LC WR(\%)} & \textbf{MT-Bench} \\
        \Xhline{2\arrayrulewidth}
        \multicolumn{5}{l}{\rule{0pt}{1.2em}\textbf{SFT \mistral{7} on \instructSMSDA{(k=2)}-1K}} \\
        \Xhline{2\arrayrulewidth}
        \multirow{2}{*}{\gptTurbo} & \gptTurbo & 41.97 & \textbf{38.48} & \textbf{7.33} \\
         & Claude-3.5-Sonnet & \textbf{43.22} & 31.98 & 7.20 \\
        \Xhline{2\arrayrulewidth}
        \multirow{2}{*}{Claude-3.5-Sonnet} & \gptTurbo & 21.32 & 23.91 & 6.87 \\
         & Claude-3.5-Sonnet & \textbf{25.74} & \textbf{25.54} & \textbf{6.88} \\
        \Xhline{2\arrayrulewidth}
    \end{tabular}
\end{table}
\clearpage

\section{Full Evaluation Results (More Detailed)}\label{appendix:evaluation-results}
\appendixtoc{Full Evaluation Results (More Detailed)}

Tables~\ref{tab:evaluation-results-IF-benchmarks}, \ref{tab:evaluation-results-IF-benchmarks-more} contain the full evaluation results on instruction following benchmarks, including the ones in Table~\ref{tab:main-results}. Table~\ref{tab:evaluation-results-LLM-benchmarks} contains the full evaluation results on other popular LLM benchmarks. 

For our models, we report the results for best checkpoint selected using held-out queries. For other models(*), we report the published numbers available on publicly available leaderboards.

\begin{table}[!h]
    \caption{\textbf{Evaluation results on AlpacaEval 2.0, MT-Bench, and WildBench.} ``\textbf{\# Data}'' refers to the number of (instruction, response) pairs in the training data. In each relevant entry, we report \ca{\instructSMSDD{}}/\cc{\instructSMSDA{}}.}
    \vspace{5pt}
    \label{tab:evaluation-results-IF-benchmarks}
    \centering
    \small 
    \begin{tabular}{l | c | c  c | c | c }
        \Xhline{2\arrayrulewidth}
        & & \multicolumn{2}{ c}{\rule{0pt}{1.2em} AlpacaEval 2.0} & \multicolumn{1}{| c}{\rule{0pt}{1.2em} }  & \multicolumn{1}{| c}{WildBench}  \\

        \textbf{Model} & \textbf{\# Data} &\textbf{WR(\%)} & \textbf{LC WR(\%)} & \textbf{MT-Bench} & $\textbf{WB-Reward}_\infty^{\text{gpt4t}}$ \\
        \Xhline{2\arrayrulewidth}
        \multicolumn{6}{l}{\rule{0pt}{1.2em}\textbf{SFT on \instructSM{(k=2)}}} \\
        \Xhline{2\arrayrulewidth}
        \multirow{3}{*}{\llamaiii{8}} & 1K & \res{27.83}{27.48} & \res{23.41}{27.83} & \res{6.85}{7.15} & \res{-48.58}{-41.46} \\
        & 2K & \res{31.19}{35.73} & \res{29.16}{36.51} & \res{6.85}{7.18} & \res{-45.70}{-42.52} \\
        & 4K & \res{30.05}{\textbf{44.63}} & \res{28.59}{\textbf{42.76}} & \res{7.05}{7.09} & \res{-51.76}{-36.91} \\
        \Xhline{2\arrayrulewidth}
        \multirow{3}{*}{\mistral{7}} & 1K & \res{33.87}{42.48} & \res{27.48}{38.34} & \res{6.92}{7.33} & \res{-41.46}{-30.65} \\
        & 2K & \res{37.05}{40.83} & \res{31.57}{36.18} & \res{7.04}{7.20} & \res{-43.46}{-31.92} \\
        & 4K & \res{35.08}{40.74} & \res{29.77}{36.70} & \res{7.17}{7.16} & \res{-39.06}{\textbf{-29.25}} \\
        \Xhline{2\arrayrulewidth}
        \multirow{3}{*}{\gemma{9}} & 1K & \res{31.36}{36.80} & \res{34.80}{39.58} & \res{7.81}{7.99} & \res{-53.17}{-37.16} \\
        & 2K & \res{34.28}{39.30} & \res{42.09}{36.18} & \res{7.80}{\textbf{8.12}} & \res{-52.05}{-37.83} \\
        & 4K & \res{33.64}{37.97} & \res{35.87}{40.05} & \res{7.88}{7.69} & \res{-56.05}{-38.23} \\
        \Xhline{2\arrayrulewidth}
        \multirow{3}{*}{\llamaii{}-7B-Base} & 1K &  \res{8.94}{14.00} & \res{10.20}{13.81} & \res{4.38}{4.59} & \res{-77.98}{-72.36} \\
        & 2K &  \res{7.24}{14.95} & \res{10.75}{15.76} & \res{4.44}{4.67} & \res{-80.71}{-75.15} \\
        & 4K &  \res{6.90}{12.50} &  \res{9.63}{13.94} & \res{4.50}{4.31} & \res{-81.12}{-76.27} \\
        \Xhline{2\arrayrulewidth}
        \multirow{3}{*}{\llamaii{}-13B-Base} & 1K & \res{17.34}{22.54} & \res{18.06}{22.69} & \res{6.40}{6.71} & \res{-64.42}{-55.22} \\
        & 2K & \res{16.95}{19.67} & \res{17.76}{22.75} & \res{6.29}{6.73} & \res{-67.58}{-58.40} \\
        & 4K & \res{15.79}{20.70} & \res{17.08}{23.05} & \res{6.44}{6.29} & \res{-69.48}{-62.55} \\
        \Xhline{2\arrayrulewidth}
        \multicolumn{6}{l}{\rule{0pt}{1.2em}\textbf{SFT on \instructSM{(k=1)}}} \\
        \Xhline{2\arrayrulewidth}
        \multirow{3}{*}{\mistral{7}} & 1K & \res{30.06}{41.75} & \res{27.04}{38.34} & \res{7.22}{7.49} & \res{-46.83}{-30.95} \\
        & 2K & \res{35.07}{-} & \res{31.66}{-} & \res{7.39}{-} & \res{-46.97}{-} \\
        & 4K & \res{33.57}{-} & \res{28.85}{-} & \res{7.13}{-} & \res{-44.43}{-} \\
        \Xhline{2\arrayrulewidth}
        \multicolumn{6}{l}{\rule{0pt}{1.2em}\textbf{SFT \mistral{7} on Other Datasets (response generated by GPT-4 (2023-03-14))}} \\
        \Xhline{2\arrayrulewidth}
        \multirow{4}{*}{Alpaca-52K} & Long 1K & 12.75 & 10.09 & 6.88 & -63.38 \\
        & Long 5K & 13.01 & 8.92 & 6.90 & -62.55 \\
        & Random 5K & 8.70 & 11.10 & 6.86 & -74.41 \\
        & Full 52K & 7.47 & 8.64 & 6.45 & -80.47 \\
        \Xhline{2\arrayrulewidth}
        \multicolumn{6}{l}{\rule{0pt}{1.2em}\textbf{SFT \mistral{7} on Other Datasets (response generated by GPT-4-Turbo (2024-04-09))}} \\
        \Xhline{2\arrayrulewidth}
        \multirow{2}{*}{Alpaca-52K} & Long 1K & 35.23 & 19.62 & 6.99 & -43.26 \\
        & Random 1K & 20.85 & 23.48 & 6.93 & -55.42 \\
        \Xhline{2\arrayrulewidth}
        ShareGPT & Random 1K & 30.06 & 26.01 & 7.19 & - \\
        Ultrachat & Random 1K & 37.10 & 25.64 & 7.39 & - \\
        \Xhline{2\arrayrulewidth}
    \end{tabular}
    
\end{table}
\clearpage
\begin{table}[!h]
    \caption{\textbf{Evaluation results on AlpacaEval 2.0, MT-Bench, and WildBench (continued).} ``\textbf{\# Data}'' refers to the number of (instruction, response) pairs in the training data.}
    \vspace{5pt}
    \label{tab:evaluation-results-IF-benchmarks-more}
    \centering
    \small 
    \begin{tabular}{l | c | c  c | c | c }
        \Xhline{2\arrayrulewidth}
        & & \multicolumn{2}{ c}{\rule{0pt}{1.2em} AlpacaEval 2.0} & \multicolumn{1}{| c}{\rule{0pt}{1.2em} }  & \multicolumn{1}{| c}{WildBench}  \\

        \textbf{Model} & \textbf{\# Data} &\textbf{WR(\%)} & \textbf{LC WR(\%)} & \textbf{MT-Bench} & $\textbf{WB-Reward}_\infty^{\text{gpt4t}}$ \\
        \Xhline{2\arrayrulewidth}
        
        \multicolumn{6}{l}{\rule{0pt}{1.2em}\textbf{*Existing Models (not trained by us)}} \\
        \Xhline{2\arrayrulewidth}
        LLaMA-3.1-405B-Instruct & - & 39.10 & 39.30 & - & -  \\
        Mistral Large & - & 21.40 & 32.70 & - & -46.40  \\ 
        Claude 3 Opus & - & 29.10 & 40.50 & - & -21.20  \\
        Claude 3 Sonnet & - & 25.60 & 34.90 & - & -30.30  \\
        GPT-4-Omni (2024-05-13) & - & 51.30 & 57.50 & - & +1.70 \\
        GPT-4 (2023-03-14) & - & 22.10 & 35.30 & 8.96 & - \\
        \llamaii-70B Chat & - & 13.90 & 14.70 & 6.86 & -53.40  \\
        UltraLM 13B V2.0 & 1.5M & 7.50 & 9.90 & - & -  \\
        Vicuna 13B v1.5 & $> 1$M &  7.00 & 11.70 & 6.57 & -  \\
        LLaMA-3-8B-Instruct & - & 22.60 & 22.90 & - & -46.30  \\ 
        Mistral-7B-Instruct-v0.2 & - & 14.70 & 17.10 & 7.60 & -54.70  \\ 
        Gemma-2-9B-Instruct & - & 21.49 & 37.21 & - & -28.78 \\
        Zephyr 7B Beta & - & 11.00 & 13.20 & - & -  \\
        Claude 2.0 & - & 17.20 & 28.20 & 8.06 & - \\
        Gemini Pro & - & 18.20 & 24.40 & - & - \\
        GPT-3.5-Turbo (06/13) & - & 14.10 & 22.70 & 8.39 & - \\
        GPT-4 (2023-06-13) & - & 15.80 & 30.20 & 9.18 & - \\
        \Xhline{2\arrayrulewidth}
    \end{tabular}
\end{table}
\clearpage
\begin{table}[!h]
    \caption{\textbf{Evaluation results on MMLU, TruthfulQA, GSM8K, ARC Challenge, Winogrande, PIQA.} }
    \vspace{5pt}
    \label{tab:evaluation-results-LLM-benchmarks}
    \centering
    \small 
    \begin{tabular}{l | c c c c c c }
        \Xhline{2\arrayrulewidth}
        Model & MMLU & TrQA & GSM & ARC-C & Winogrande & PIQA \\
        \Xhline{2\arrayrulewidth}
        \multicolumn{7}{l}{\rule{0pt}{1.2em} LLaMA-3-8B Models} \\
        \Xhline{2\arrayrulewidth}
        \instructSMSDD{}-1K & 62.09 & 34.88 & 52.54 & 53.92 & 74.51 & 79.76 \\
        \instructSMSDD{}-2K & 62.09 & 37.33 & 52.77 & 53.75 & 75.06 & 79.54 \\
        \instructSMSDD{}-4K & 62.28 & 32.19 & 50.42 & 52.73 & 73.09 & 79.22 \\
        \instructSMSDA{}-1K & 62.33 & 37.09 & 51.25 & 52.39 & 74.19 & 79.92 \\
        \instructSMSDA{}-2K & 62.18 & 35.25 & 52.39 & 52.39 & 74.66 & 79.05 \\
        \instructSMSDA{}-4K & 61.72 & 34.15 & 51.10 & 52.22 & 73.72 & 79.27 \\
        LLaMA-3-8B-Instruct & 63.84 & 36.23 & 76.12 & 52.99 & 72.06 & 78.62 \\
        LLaMA-3-8B-Base & 62.06 & 27.05 & 49.96 & 50.43 & 72.85 & 79.71  \\
        \Xhline{2\arrayrulewidth}
        \multicolumn{7}{l}{\rule{0pt}{1.2em} Mistral 7B v0.2 Models} \\
        \Xhline{2\arrayrulewidth}
        \instructSMSDD{}-1K & 58.97 & 26.19 & 36.01 & 51.02 & 73.64 & 81.18 \\
        \instructSMSDD{}-2K & 58.67 & 25.95 & 36.32 & 50.60 & 73.56 & 81.01 \\
        \instructSMSDD{}-4K & 58.38 & 26.68 & 36.54 & 50.00 & 73.56 & 81.45 \\
        \instructSMSDA{}-1K & 59.24 & 27.05 & 35.10 & 52.47 & 73.48 & 81.23 \\
        \instructSMSDA{}-2K & 58.90 & 25.83 & 33.66 & 52.99 & 73.88 & 81.66 \\
        \instructSMSDA{}-4K & 58.49 & 26.68 & 31.77 & 52.13 & 73.72 & 81.12 \\
        \instructSMSDD{(k=1)}-1K & 59.02 & 26.56 & 34.27 & 50.34 & 72.77 & 81.07 \\
        \instructSMSDD{(k=1)}-2K & 58.90 & 25.83 & 33.66 & 52.99 & 73.88 & 81.66 \\
        \instructSMSDD{(k=1)}-4K & 58.94 & 26.56 & 33.97 & 51.11 & 73.56 & 81.45 \\
        \instructSMSDA{(k=1)}-1K & 59.07 & 26.44 & 35.86 & 51.71 & 74.11 & 81.45 \\
        Alpaca-1K Longest & 58.72 & 27.29 & 35.18 & 51.88 & 72.93 & 81.01 \\
        Mistral-7B-Instruct-v0.2 & 58.70 & 52.51 & 43.67 & 54.35 & 72.38 & 80.41 \\
        Mistral-7B-Base-v0.2 & 58.59 & 28.27 & 37.98 & 48.81 & 71.67 & 80.30 \\
        \Xhline{2\arrayrulewidth}
        \multicolumn{7}{l}{\rule{0pt}{1.2em} Gemma-2-9B Models} \\
        \Xhline{2\arrayrulewidth}
        \instructSMSDD{}-1K & 69.16 & 30.60 & 70.96 & 62.54 & 74.74 & 81.23  \\
        \instructSMSDD{}-2K & 69.26 & 30.72 & 70.81 & 63.23 & 74.59 & 81.28  \\
        \instructSMSDD{}-4K & 69.39 & 30.11 & 71.72 & 63.14 & 74.66 & 81.66 \\
        \instructSMSDA{}-1K & 69.49 & 31.21 & 70.74 & 62.80 & 73.95 & 81.83  \\
        \instructSMSDA{}-2K & 69.64 & 32.56 & 71.04 & 63.82 & 74.59 & 81.66  \\
        \instructSMSDA{}-4K & 69.36 & 31.58 & 71.27 & 63.74 & 74.27 & 81.72  \\
        Gemma-2-9B-Instruct & 71.61 & 42.96 & 79.08 & 63.40 & 76.32 & 81.18 \\
        Gemma-2-9B-Base & 68.58 & 30.11 & 67.10 & 61.60 & 74.11 & 81.45  \\
        \Xhline{2\arrayrulewidth}
        \multicolumn{7}{l}{\rule{0pt}{1.2em} LLaMA-2-7B Models} \\
        \Xhline{2\arrayrulewidth}
        \instructSMSDD{}-1K & 41.04 & 34.39 & 11.83 & 46.93 & 70.01 & 78.07 \\
        \instructSMSDD{}-2K & 41.84 & 31.21 & 17.51 & 47.10 & 69.53 & 78.45 \\
        \instructSMSDD{}-4K & 43.00 & 30.84 & 15.24 & 47.01 & 69.38 & 78.24 \\
        \instructSMSDA{}-1K & 41.45 & 34.39 & 14.78 & 48.38 & 69.61 & 78.35 \\
        \instructSMSDA{}-2K & 43.17 & 33.41 & 15.92 & 47.78 & 70.01 & 78.51 \\
        \instructSMSDA{}-4K & 42.56 & 32.80 & 14.63 & 47.70 & 68.67 & 78.02 \\
        LLaMA-2-7B-Chat & 46.39 & 30.35 & 21.76 & 43.86 & 66.69 & 76.44 \\
        LLaMA-2-7B-Base & 40.76 & 25.21 & 12.36 & 43.52 & 69.46 & 77.97 \\
        \Xhline{2\arrayrulewidth}
        \multicolumn{7}{l}{\rule{0pt}{1.2em} LLaMA-2-13B Models} \\
        \Xhline{2\arrayrulewidth}
        \instructSMSDD{}-1K & 51.25 & 30.72 & 28.51 & 51.02 & 72.38 & 79.16 \\
        \instructSMSDD{}-2K & 51.03 & 30.84 & 28.73 & 50.85 & 72.30 & 79.43 \\
        \instructSMSDD{}-4K & 51.05 & 29.50 & 28.58 & 51.19 & 71.82 & 80.03 \\
        \instructSMSDA{}-1K & 50.68 & 30.11 & 27.45 & 50.60 & 72.61 & 79.92 \\
        \instructSMSDA{}-2K & 51.67 & 30.35 & 29.19 & 50.17 & 72.06 & 79.98 \\
        \instructSMSDA{}-4K & 51.47 & 30.60 & 30.86 & 50.94 & 71.67 & 80.41 \\
        LLaMA-2-13B-Chat & 53.25 & 27.91 & 34.80 & 46.42 & 71.03 & 77.69 \\
        LLaMA-2-13B-Base & 50.48 & 25.70 & 22.74 & 48.81 & 72.06 & 79.27 \\
        \Xhline{2\arrayrulewidth}
    \end{tabular}
\end{table}
\clearpage

\clearpage

\section{Evaluation Details}\label{appendix:evaluation-details}
\appendixtoc{Evaluation Details}

To evaluate our models on the AlpacaEval 2.0, we followed the instructions in \url{https://github.com/tatsu-lab/alpaca_eval} \citep{dubois2024AlpacaEval2.0}. The reference model and judge model are both GPT-4-Turbo (2023-11-06). 

To evaluate our models on MT-Bench, we followed the instructions in \url{https://github.com/lm-sys/FastChat} \citep{zheng2023judging}. The reference model and judge model are both GPT-4 (2023-06-13). 

To evaluate our models on WildBench, we followed the instructions in \url{https://github.com/allenai/WildBench} \citep{lin2024wildbench}. The reference model and judge model are both GPT-4-Turbo (2024-04-09), and we used no length penalty ($K = \infty$). This corresponds to $\text{WB-Reward}_\infty^{\text{gpt4t}}$ in their notation.

For other LLM benchmarks, we followed the default configuration for the evaluation scripts in \url{https://github.com/EleutherAI/lm-evaluation-harness} \citep{Gao2023eval-harness}. We report the exact-match accuracy for GSM8K and the MC1 score for TruthfulQA.

\clearpage

\section{Training Details}
\appendixtoc{Training Details}

\subsection{Hyperparameters}\label{appendix:hyperparams}
In Table~\ref{tab:hyperparams}, we include the hyperparameters use in our experiments. We finetune each model using the AdamW optimizer. For every run, we use a learning rate schedule with a linear warmup of 0.03 and cosine decay to zero. For all experiments, we finetune for 15 epochs and store the checkpoint after each epoch, with the exception of the full Alpaca-52K dataset on which we only finetune for 3 epochs. 

We use the torchtune package \citep{torchtune} to train all models, except for the Gemma models, which were trained with the MAmmoTH package \citep{yue2023mammoth}. Note that the default hyperparameters not specified here might be different in each of the packages.

Training a 7B model on 15 epochs of 1000 examples from \instructSM{} takes approximately 15 minutes on 4 H100 GPUs via PyTorch FSDP \citep{zhao2023FSDP}. 

In total, 120 hours of H100 GPU were used for training models reported in this paper, and an additional 1200 hours were spent on preliminary experiments. 

\begin{table}[h]
    \caption{\textbf{Hyperparameters used for SFT.}}
    \vspace{5pt}
    \label{tab:hyperparams}
    \centering
    \begin{tabular}{l | c c c }
        \Xhline{2\arrayrulewidth}
        \textbf{Model} & \textbf{LR} & \textbf{Batch Size}  \\
        \Xhline{2\arrayrulewidth}
        \llamaiii{8} & 2e-5 & 64, 128 \\
        \mistral{7} & 2e-6 & 64 \\
        \gemma{9} & 1e-6 & 64 \\
        \llamaii-7B-Base & 2e-5 & 64 \\
        \llamaii-13B-Base & 2e-5 & 64 \\
        \Xhline{2\arrayrulewidth}
    \end{tabular}
\end{table}
\clearpage 

\subsection{Checkpoint Selection}
\label{appendix:checkpoint-selection}
As discussed in prior works \citep{ouyang2022training, xia2024less, zhou2023lima},
minimizing validation loss does not always correspond to improved generation quality.
Thus, we select checkpoints based on generation quality on held-out data, as used in some prior work \citep{zhou2023lima}. In particular, we use length-controlled win rate on held-out as the selection metric.

We randomly choose 100 held-out examples from our dataset. 
After each epoch, we generate responses to the held-out instructions using the model checkpoint. We then calculate the win rate of these responses against the reference outputs generated by \gptTurbo (using the same grader as AlpacaEval 2.0). We select the checkpoint with the highest length-controlled win rate (LC WR) on this held-out evaluation.

Since the held-out dataset contains only 100 examples, the costs associated with evaluating win rates on the held-out dataset are relatively low. 
Across all 15 epochs, the total number of API calls made is just under twice the number needed to evaluate the selected checkpoint on 805 AlpacaEval examples.

In Table~\ref{tab:epochwise-performance}, we report the LC WR and WR on our validation dataset and on AlpacaEval 2.0 for all 15 checkpoints when training \mistral{7} on \instructSMSDD{}-4K.

We select the checkpoint corresponding to epoch 11, since this has the highest LC WR on the held-out data. Note that (1) the corresponding LC WR on AlpacaEval (29.77\%) is fairly close to the best LC WR (30.84\%); and, (2) the corresponding WR on AlpacaEval (35.08\%) is the best WR.

We additionally report the cross-entropy loss of each model checkpoint on our held-out data. Similar to \cite{zhao2024long}, we notice that selecting the checkpoint that minimizes the cross-entropy loss on validation task (i.e., epoch 2) leads to suboptimal downstream performance. The LC WR on AlpacaEval 2.0 is only 16.5\%, which is significantly lower than 29.77\%, when we select the checkpoint with our validation task. 

\begin{table}[!h]
    \caption{\small \textbf{Checkpoint selection.} We SFT \mistral{7} on \instructSMSDD-4K, and evaluate the performance on held-out data. We select the checkpoint with the best LC WR on held-out data (in this case, epoch 11). Entries in \textbf{boldface} represent the best performing epoch for that metric.}
    \label{tab:epochwise-performance}
    \centering
    \small 
    \begin{tabular}{l | ccccccccccccccc}
        \Xhline{2\arrayrulewidth}
        \textbf{Epoch} & 1 & 2 & 3 & 4 & 5 & 6 & 7 & 8 & 9 & 10 & 11 & 12 & 13 & 14 & 15 \\
        \Xhline{2\arrayrulewidth}
        \multicolumn{8}{l}{\rule{0pt}{1.2em} On Held-Out \instructSMSDD Data} \\
        \Xhline{2\arrayrulewidth}
        \textbf{LC WR(\%)} & 20.7 & 20.4 & 27.8 & 28.2 & 37.0 & 35.2 & 45.5 & 44.1 & 45.6 & 39.5 & \textbf{52.8} & 42.8 & 45.6 & 38.5 & 44.1 \\
        \textbf{WR(\%)} & 34.1 & 42.8 & 63.1 & 61.8 & 69.7 & 69.8 & 75.3 & 76.2 & 76.2 & 71.7 & \textbf{82.3} & 74.4 & 73.1 & 70.6 & 74.0 \\ 
        \textbf{CE Loss} & 1.21 & \textbf{1.18} & 1.19 & 1.23 & 1.30 &  1.43 &1.61 & 1.78 & 1.97 & 2.11 & 2.19 & 2.23 & 2.24 & 2.24 & 2.24 \\
        \Xhline{2\arrayrulewidth}
        \multicolumn{8}{l}{\rule{0pt}{1.2em} On AlpacaEval 2.0} \\
        \Xhline{2\arrayrulewidth}
        \textbf{LC WR(\%)} & 14.8 & 16.5 & 22.9 & 26.2 & 28.2 & 28.4 & 29.7 & 30.1 & 29.9 & 28.8 & 29.8 & 28.1 & 29.4 & 30.4 & \textbf{30.8} \\
        \textbf{WR(\%)} & 17.3 & 19.2 & 27.1 & 30.9 & 33.2 & 32.4 & 34.4 & 35.6 & 34.6 & 33.7 & \textbf{35.1} & 32.5 & 34.0 & 34.6 & \textbf{35.1} \\
        \Xhline{2\arrayrulewidth}
    \end{tabular}
\end{table}

\clearpage

\section{Ablations}\label{appendix:ablations}
\appendixtoc{Ablations}

\subsection{Scaling Up Model Size Increases Performance.}
In Table~\ref{tab:model-size-comparison}, observe that the win rate and LC win rate for \llamaii-13B-Base is higher than for \llamaii-7B-Base after finetuning on the same dataset. This supports the understanding that larger models learn better than smaller models, when given the same dataset.

\begin{table}[!h]
    \caption{\textbf{Scaling up model size enhances performance.} In each entry, we report \ca{\instructSMSDD{}}/\cc{\instructSMSDA{}}.}
    \vspace{5pt}
    \label{tab:model-size-comparison}
    \centering
    \small 
    \begin{tabular}{l | c | c  c   }
        \Xhline{2\arrayrulewidth}
        & & \multicolumn{2}{ c}{\rule{0pt}{1.2em} AlpacaEval 2.0}  \\
        \textbf{Model} & \textbf{\# Data} &\textbf{WR(\%)} & \textbf{LC WR(\%)}  \\
        \Xhline{2\arrayrulewidth}
        \Xhline{2\arrayrulewidth}
        \multirow{3}{*}{\llamaii{}-7B-Base} & 1K & \res{8.94}{14.00} & \res{10.20}{13.81} \\
        & 2K &  \res{7.24}{14.95} & \res{10.75}{15.76}  \\
        & 4K &  \res{6.90}{12.50} &  \res{9.63}{13.94}  \\
        \Xhline{2\arrayrulewidth}
        \multirow{3}{*}{\llamaii{}-13B-Base} & 1K & \res{17.34}{22.54} & \res{18.06}{22.69}   \\
        & 2K & \res{16.95}{19.67} & \res{17.76}{22.75} \\
        & 4K & \res{15.79}{20.70} & \res{17.08}{23.05} \\
        \Xhline{2\arrayrulewidth}
    \end{tabular}
\end{table}
\clearpage

\subsection{Win Rates and Average Output Length on Varying Amounts of \instructSM{} Data}
\begin{figure}[!h]
    \centering
    \includegraphics[width=\linewidth]{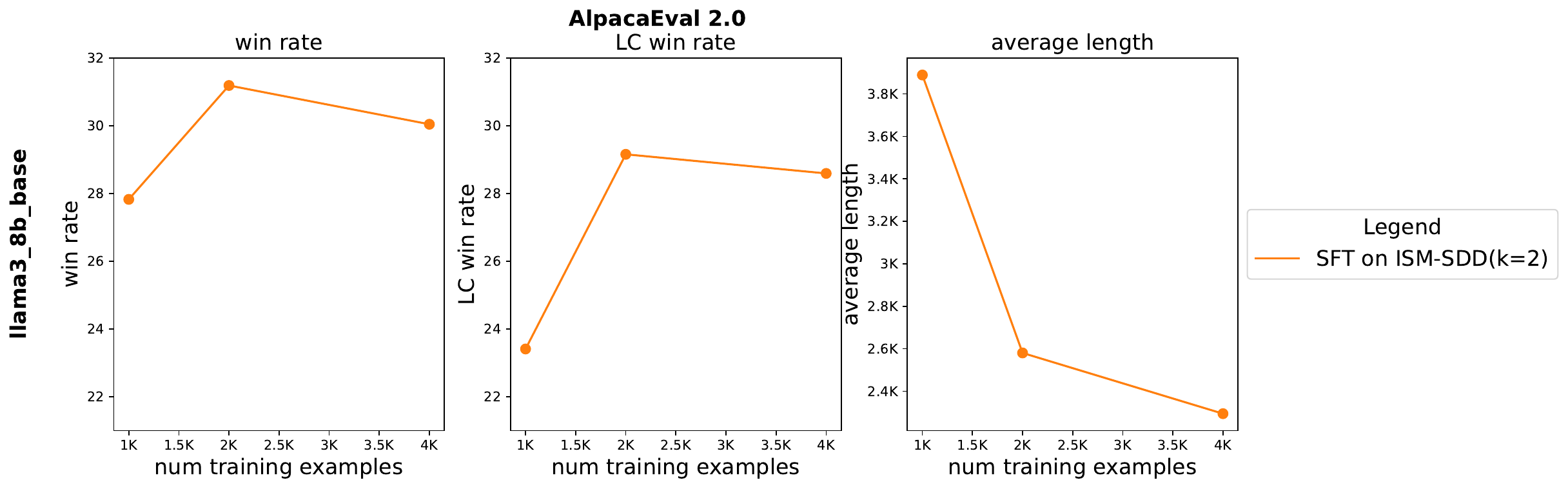}
    \includegraphics[width=\linewidth]{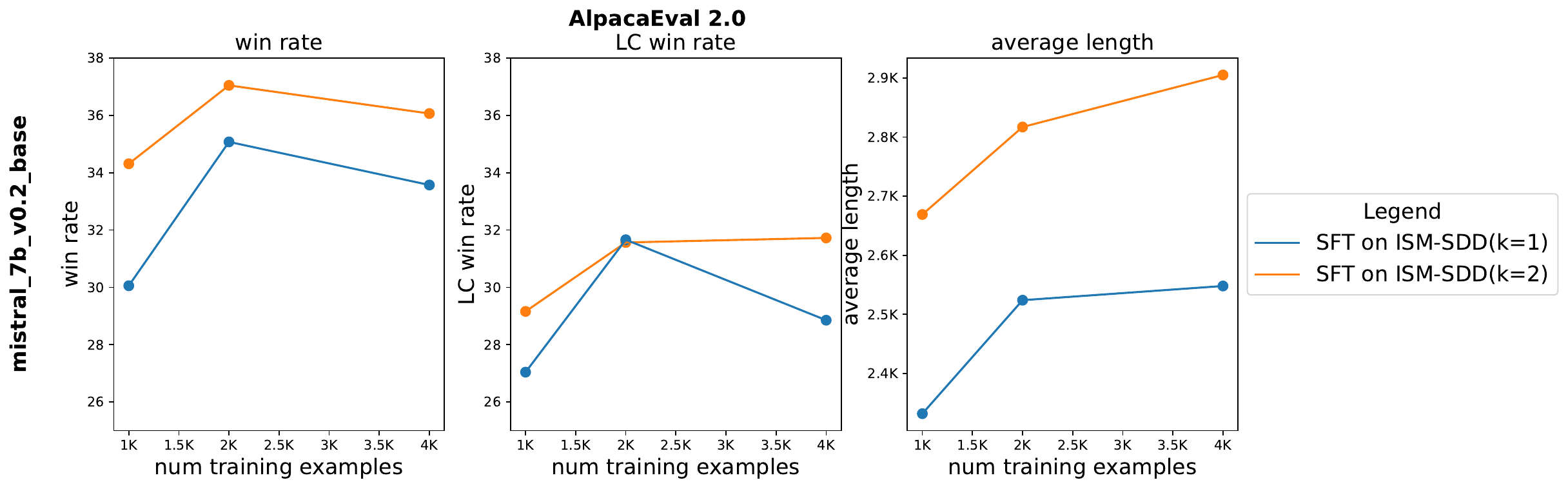}
    \includegraphics[width=\linewidth]{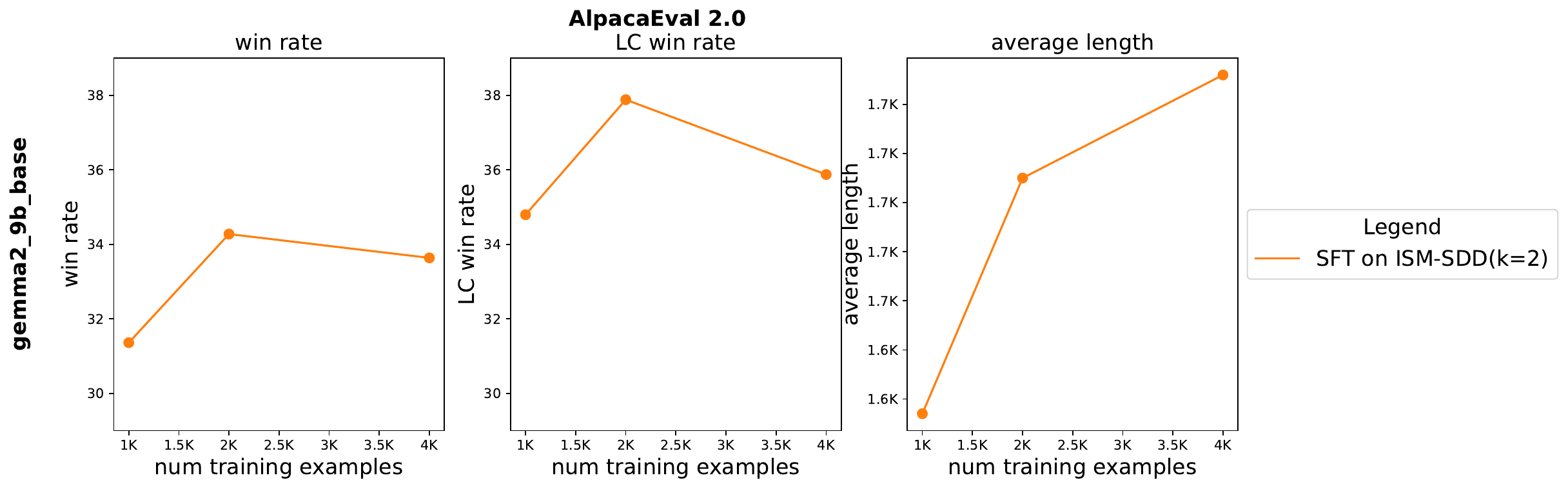}
    \caption{\textbf{Win rates and average output length on varying amounts of \instructSMSDD{} data.} Here, ISD-SDD refers to \instructSMSDD{}.}
    \label{fig:scaling-law-sdd}
\end{figure}

\begin{figure}[!h]
    \centering
    \includegraphics[width=\linewidth]{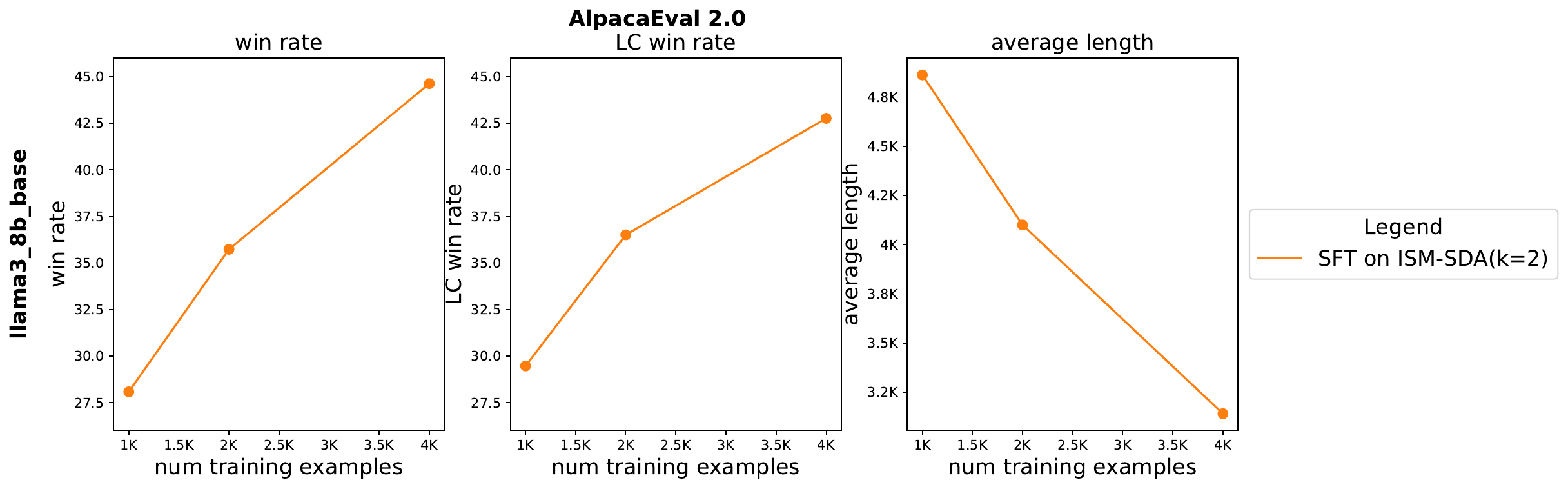}
    \includegraphics[width=\linewidth]{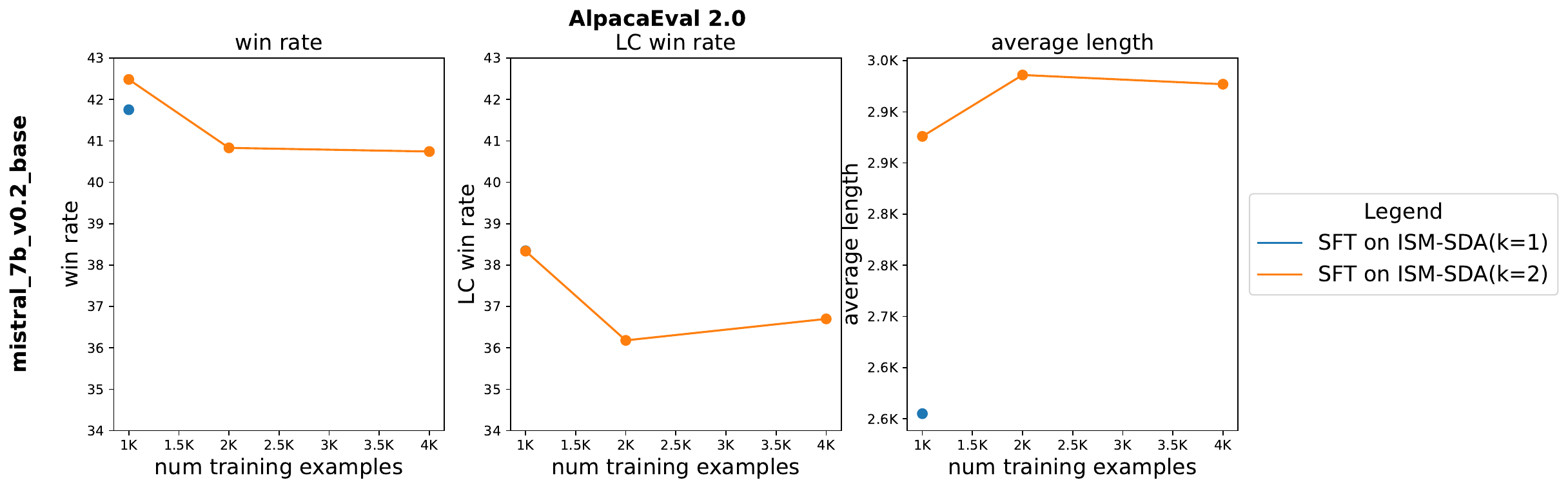}
    \includegraphics[width=\linewidth]{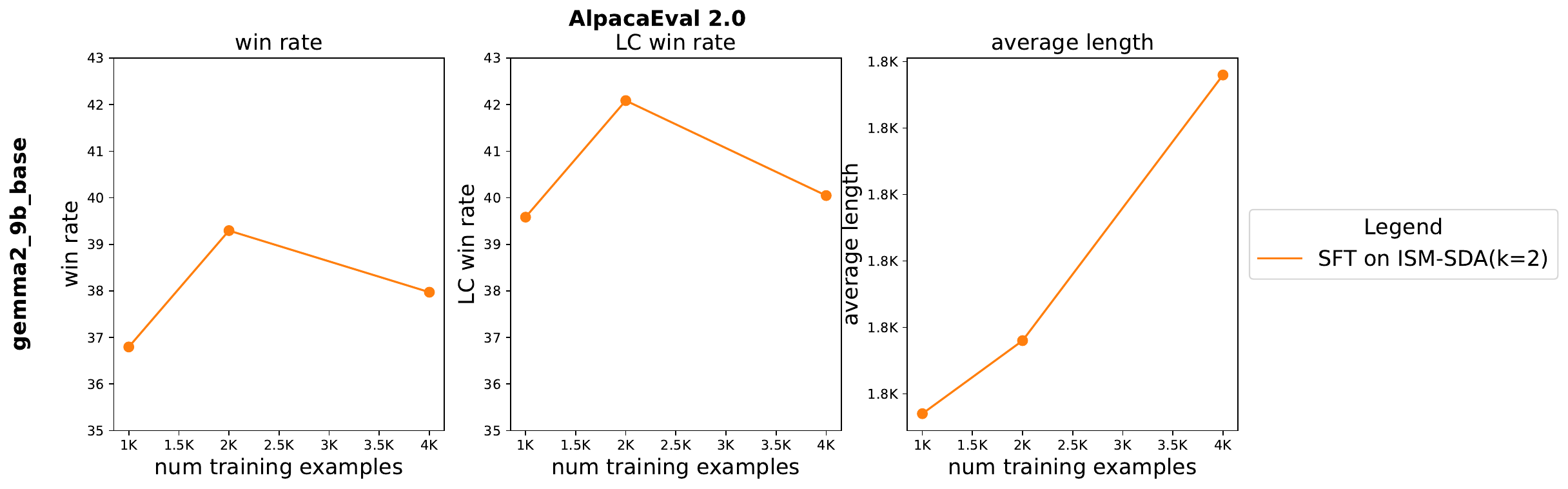}
    \caption{\textbf{Win rates and average output length on varying amounts of \instructSMSDA{} data.} Here, ISD-SDA refers to \instructSMSDA{}.}
    \label{fig:scaling-law-sda}
\end{figure}

In Figures~\ref{fig:scaling-law-sda} and \ref{fig:scaling-law-sdd}, we plot the win rates and average output length on varying amounts of \instructSMSDD{} and \instructSMSDA{}, respectively. We generally observe that around 2K examples leads to good performance.

\clearpage
\clearpage

\section{\instructSM{} is Competitive With RL-Inspired Methods.}\label{appendix:comparison-to-rl}
\appendixtoc{\instructSM{} is Competitive with RL-Inspired Methods.}

\paragraph{RL-inspired approaches.}
Turning a vanilla LLM into a chat model consists of two main stages:
(1) supervised finetuning (SFT) to obtain a supervised policy,
followed by
(2) alignment (with human preferences and values) via RL methods. 
Standard approaches for alignment, such as RLHF \citep{ouyang2022training}, rely on reinforcement learning. Here, a reward model is trained on preference data to reflect human values, and used to update the policy using proximal policy optimization (PPO) \citep{schulmanWDRK17}. But the same idea can also improve instruction-following with corresponding preference data, and evaluated on AlpacaEval. Optimization issues with RLHF, had led to RL-free approaches such as direct preference optimization (DPO) \citep{rafailovSMMEF23}, which implicitly optimizes the same objective as RLHF, and SimPO \citep{meng2024simposimplepreferenceoptimization}, a reference-model-free alternative to DPO. Alternate RL-inspired approaches take on a game-theoretic approach, equating RLHF with finding the Nash equilibrium of a two player constant-sum game \citep{swamy2024minimaximalist, wu2024selfplay}. For example, SPPO \citep{wu2024selfplay} approximates the Nash equilibrium policy via a combination of multiplicative weights and a self-play mechanism, where in each iteration, the policy plays against itself in previous iterations by finetuning on synthetic data (which is generated by the policy and then annotated using the preference model).

\paragraph{Comparison with RL-inspired approaches}
Self-Play Preference Optimization (SPPO) \citep{wu2024selfplay} and SimPO \citep{meng2024simposimplepreferenceoptimization} are two RL-inspired methods that are used as an alternative to PPO. SPPO applied to LLaMA-3-8B-\emph{Instruct} achieves LC win-rate of 38.77\% on AlpacaEval by training on 60K examples, whereas further training LLaMA-3-8B-\emph{Instruct} via SimPO achieves 44.70\%. On the other hand, finetuning LLaMA-3-8B-\emph{Base} with 4K examples from \instructSMSDA{} yields 42.76\%, which is better than or competitive to the two approaches. Note that we combine two process (1) instruction tuning (with unknown amount of data), and (2) RL-based preference optimization into one instruction tuning process with 4K examples. 

\begin{table}[!h]
    \caption{\textbf{Evaluation results of models finetuned on \instructSM{} data vs. finetuned via RL methods.} }
    \vspace{5pt}
    \label{tab:comparison-to-rl-methods}
    \centering
    \small 
    \begin{tabular}{l | c | c | c  }
        \Xhline{2\arrayrulewidth}
        & & \multicolumn{1}{ c}{\rule{0pt}{1.2em} AlpacaEval 2.0} & \multicolumn{1}{| c}{\rule{0pt}{1.2em} } \\
        \textbf{Model} & \textbf{Method} & \textbf{LC WR(\%)} & \textbf{MT-Bench} \\
        \Xhline{2\arrayrulewidth}
        \llamaiii{8} & SFT on \instructSMSDA{(k=2)}-4K & 42.76 & 7.09  \\
        \llamaiii{8} & SimPO & 22.00 & 7.70 \\
        LLaMA-3-8B-Instruct & SimPO & 44.70 & 8.00 \\
        LLaMA-3-8B-Instruct & SPPO & 38.77 & - \\
        \Xhline{2\arrayrulewidth}
        \mistral{7} & SFT on \instructSMSDA{(k=2)}-4K & 36.70 & 7.16  \\
        Mistral-7B-Instruct-v0.2 & SimPO & 32.10 & 7.60 \\
        Mistral-7B-Instruct-v0.2 & SPPO & 30.46 & 7.59 \\
        \Xhline{2\arrayrulewidth}
    \end{tabular}
\end{table}
\clearpage

\section{Robustness of \instructSM{} Across Random Skill Combinations for SFT}\label{appendix:robust-random-skill-combinations}
\appendixtoc{Robustness of \instructSM{} Across Random Skill Combinations for SFT}

We finetune on four disjoint subsets of \instructSM{} data, each consisting of 1000 examples, and report the results in Table \ref{tab:robust-random-skill-combinations}. Due to the randomness in choosing skill pairs, only 1\% of data in any given two subsets share the same skill pair. Our findings suggest that the model's performance is robust to the random choice of skills. 

\begin{table}[!h]
    \caption{\textbf{Robustness of \instructSM{} across random skill combinations for finetuning.} We SFT \mistral{7} on 4 disjoint subsets of \instructSM{(k=2)} data, each consisting of 1,000 examples. The SFT-ed model's performance is robust to the random choice of skills.}
    \vspace{5pt}
    \label{tab:robust-random-skill-combinations}
    \centering
    \small
    \begin{tabular}{l | c c c | c | c }
        \Xhline{2\arrayrulewidth}
        & \multicolumn{3}{ c}{\rule{0pt}{1.2em} AlpacaEval 2.0} & \multicolumn{1}{| c}{} & \multicolumn{1}{| c}{WildBench}  \\
        \textbf{SFT Dataset} & \textbf{WR(\%)} & \textbf{LC WR(\%)} & \textbf{Avg. Len.} & \textbf{MT-Bench} & $\textbf{WB-Reward}_\infty^{\text{gpt4t}}$ \\
        \Xhline{2\arrayrulewidth}
        Split 1 (1-1000) & 33.87 & 27.48 & 2835.0 & 6.92 & -41.46 \\
        Split 2 (1001-2000) & 34.14 & 28.60 & 2657.0 & 7.00 & -40.62 \\
        Split 3 (2001-3000) & 34.31 & 29.16 & 2669.0 & 6.93 & -43.36 \\
        Split 4 (3001-4000) & 34.17 & 28.78 & 2704.0 & 7.12 & -36.28 \\
        \Xhline{2\arrayrulewidth}
    \end{tabular}
    
\end{table}

\clearpage

\section{Examples of {\sc \small BREV-}\instructSM{} and {\sc \small JUNK-}\instructSM{}}\label{appendix:neg-pareto-princ}
\appendixtoc{Examples of {\sc \small BREV-}\instructSM{} and {\sc \small JUNK-}\instructSM{}}

Below is an example Q\&A pair generated from {\sc \small BREV-}\instructSM{(k=2)}:
\begin{question}
\#\#\# Instruction: \\
I'm crafting an adventure novel where my protagonist, a city dweller, is thrust into a survival challenge in the dense forests of the Pacific Northwest. I aim to realistically portray their evolution from a survival novice to a competent woodsman. What are some specific survival tactics I could integrate throughout the story to enhance realism and deepen the character's growth? Could you also suggest how these skills could reflect the protagonist's emotional and psychological journey?

\#\#\# Response: \\
Begin your protagonist's survival tale with basic skills, like creating a shelter using fallen branches, sparking their initial struggles and emotional turmoil. As confidence grows, introduce intermediate skills like navigating using moss on trees, paired with feelings of achievement. Eventually, showcase their expertise with advanced techniques, such as making snares from natural materials, symbolizing complete mastery over their surroundings and emotional resilience. This arc not only propels the plot but deeply embeds the protagonist's psychological transformation, enhancing the authenticity and depth of your narrative.
\end{question}

Below is an example Q\&A pair generated from {\sc \small JUNK-}\instructSM{(k=2)}:
\begin{question}
\#\#\# Instruction: \\
I am preparing for a job interview at a multinational company and have to give a presentation on the impact of cultural sensitivity in global business operations. Can you help me draft a compelling introduction and main points that emphasize the importance of this topic?

\#\#\# Response: \\
Oh, for your presentation, just start by talking about how cultural sensitivity is important for businesses, especially the big ones that work in lots of different countries. Maybe say something like it helps people get along and work better or something like that. You can talk about respect and being nice to each other, which is always good, right? Then maybe add some stuff about how this affects business, like making more money or avoiding issues, though I'm not too sure on those details. Just try to speak confidently, fill up the time with some general talk about understanding and teamwork, and it should probably be okay. Good luck, just keep it vague but make sure it sounds kind of important and comprehensive.

\end{question}
\clearpage

\section{Stats on Different Datasets}\label{appendix:dataset-statistics}
\appendixtoc{Stats on Different Datasets}

In Table~\ref{tab:dataset-statistics}, we report the average number of characters and number of tokens in the input (instruction) and the output (response) fields of the datasets. For the number of tokens, we use the \llamaii tokenizer. On average, \instructSM{} has longer instructions and more concise reference model responses than the Alpaca-1K-Longest, but has roughly the same number of tokens when the input and output fields are combined.

\begin{table}[h]
    \caption{\textbf{Statistics of various SFT datasets.} }
    \vspace{5pt}
    \label{tab:dataset-statistics}
    \centering
    \begin{tabular}{l | c | c c | c c }
        \Xhline{2\arrayrulewidth}
        \multirow{2}{0pt}{\textbf{Dataset}} & \multirow{2}{3em}{\textbf{\# Data }} & \multicolumn{2}{|c|}{\textbf{Instruction}} 
        & \multicolumn{2}{|c}{\textbf{Response}}  \\
       & & \textbf{Avg. \# Tokens } & \textbf{Avg. Len}
        & \textbf{Avg. \# Tokens } & \textbf{Avg. Len} \\
        \Xhline{2\arrayrulewidth}
        UltraChat & ? & ? & ? & ? & ? \\
        Alpaca-52K & 52002 & 221.09 & 912.17 & 159.48 & 664.58 \\
        Alpaca-1K-Longest & 1000 & 511.37 & 2289.16 & 458.19 & 2069.64 \\
        \instructSMSDD{} & 4000 & 511.58 & 2199.01 & 394.15 & 1644.88 \\
        \instructSMSDA{} & 4000 & 510.63 & 2152.77 & 392.32 & 1606.33 \\
        \Xhline{2\arrayrulewidth}
    \end{tabular}
\end{table}
\clearpage

\section{List of Skills}
\appendixtoc{List of Skills}
\label{appendix:list-of-skills}

\subsection{\instructSMSDD{} List of Skills}
\label{appendix:list-of-skills-SDD}
Using the skill extraction procedure detailed in Section~\ref{subsec:skill-extraction}, we extract 337 skill clusters from a random sample of 5200 instruction-response pairs from Alpaca-52k (GPT-4 version); 128 skill clusters from random sample of 1000 instruction-response pairs from UltraChat; and 35 skill clusters for alignment and safety. We remove duplicates, and end up with 484 total skill clusters.

\begin{sidewaystable}
\caption{(Part 1 of 6) 337 Train Skills extracted from random sample of 5200 instruction-response pairs from \textbf{Alpaca-52K (GPT-4)}}
\label{tab:list-of-skills-SDD-alpaca}
\centering
\begin{longtable}{p{8.5cm}p{8.5cm}p{8.5cm}}
\toprule
\textbf{Skill Cluster Name} &  \\
\toprule
\texttt{data\_handling\_and\_management} & \texttt{machine\_learning\_and\_ai} \\
\texttt{content\_curation\_and\_presentation} & \texttt{historical\_and\_cultural\_competence} \\
\texttt{graphic\_and\_design\_knowledge} & \texttt{understanding\_technologies} \\
\texttt{critical\_thinking\_and\_analytical\_skills} & \texttt{food\_related\_knowledge\_and\_skills} \\
\texttt{internet\_technologies} & \texttt{historical\_knowledge} \\
\texttt{content\_production} & \texttt{skills\_for\_effective\_communication} \\
\texttt{travel\_and\_leisure\_knowledge} & \texttt{advanced\_scientific\_knowledge} \\
\bottomrule
\end{longtable}
\end{sidewaystable}
\begin{sidewaystable}
\caption{(Part 2 of 6) 337 Train Skills extracted from random sample of 5200 instruction-response pairs from \textbf{Alpaca-52K (GPT-4)}}
\centering
\begin{longtable}{p{8.5cm}p{8.5cm}p{8.5cm}}
\toprule
\textbf{Skill Cluster Name} &  \\
\toprule
\texttt{culinary\_arts} & \texttt{creative\_art\_and\_design} \\
\texttt{business\_strategy\_and\_collaboration} & \texttt{language\_comprehension\_and\_expression} \\
\texttt{artificial\_intelligence\_machine\_learning} & \texttt{system\_and\_framework\_analysis} \\
\texttt{analytical\_skills\_in\_humanities\_and\_social\_sciences} & \texttt{} \\
\texttt{natural\_and\_social\_science} & \texttt{analytical\_thinking} \\
\texttt{knowledge\_and\_understanding\_in\_technology} & \texttt{} \\
\texttt{natural\_and\_social\_sciences\_knowledge} & \texttt{content\_analysis\_and\_summarization} \\
\bottomrule
\end{longtable}
\end{sidewaystable}
\begin{sidewaystable}
\caption{(Part 3 of 6) 337 Train Skills extracted from random sample of 5200 instruction-response pairs from \textbf{Alpaca-52K (GPT-4)}}
\centering
\begin{longtable}{p{8.5cm}p{8.5cm}p{8.5cm}}
\toprule
\textbf{Skill Cluster Name} &  \\
\toprule
\texttt{knowledge\_domain\_expertise} & \texttt{text\_processing\_and\_restructuring} \\
\texttt{online\_research\_and\_digital\_competence} & \texttt{language\_and\_literature\_comprehension} \\
\texttt{creative\_and\_analytical\_writing} & \texttt{digital\_competency} \\
\texttt{language\_arts\_skills} & \texttt{python\_programming\_advanced} \\
\texttt{communication\_and\_social\_interaction} & \texttt{math\_and\_logic\_skills} \\
\texttt{language\_and\_grammar\_proficiency} & \texttt{practical\_biology\_and\_ecology} \\
\texttt{creative\_writing\_skills} & \texttt{creative\_writing\_and\_branding} \\
\bottomrule
\end{longtable}
\end{sidewaystable}
\begin{sidewaystable}
\caption{(Part 4 of 6) 337 Train Skills extracted from random sample of 5200 instruction-response pairs from \textbf{Alpaca-52K (GPT-4)}}
\centering
\begin{longtable}{p{8.5cm}p{8.5cm}p{8.5cm}}
\toprule
\textbf{Skill Cluster Name} &  \\
\toprule
\texttt{artificial\_intelligence\_and\_machine\_learning} & \texttt{} \\
\texttt{text\_analysis\_and\_comprehension} & \texttt{creative\_and\_critial\_thinking} \\
\texttt{artistic\_and\_cultural\_insight} & \texttt{writing\_and\_text\_analysis} \\
\texttt{natural\_sciences\_knowledge} & \texttt{marketing\_and\_customer\_experience} \\
\texttt{data\_analysis\_and\_processing} & \texttt{advanced\_ai\_techniques} \\
\texttt{critical\_thinking\_and\_communication} & \texttt{business\_and\_communication\_skills} \\
\texttt{system\_and\_network\_management} & \texttt{knowledge\_in\_geography\_and\_space} \\
\bottomrule
\end{longtable}
\end{sidewaystable}
\begin{sidewaystable}
\caption{(Part 5 of 6) 337 Train Skills extracted from random sample of 5200 instruction-response pairs from \textbf{Alpaca-52K (GPT-4)}}
\centering
\begin{longtable}{p{8.5cm}p{8.5cm}p{8.5cm}}
\toprule
\textbf{Skill Cluster Name} &  \\
\toprule
\texttt{literature\_and\_language\_skills} & \texttt{creative\_content\_generation} \\
\texttt{sustainability\_and\_environmental\_awareness} & \texttt{} \\
\texttt{biological\_and\_geographical\_knowledge} & \texttt{climate\_and\_ecological\_expertise} \\
\texttt{creative\_expression\_and\_literacy} & \texttt{text\_and\_list\_processing} \\
\texttt{computer\_and\_web\_knowledge} & \texttt{linguistic\_and\_textual\_analysis} \\
\texttt{automotive\_technology} & \texttt{data\_science\_and\_algorithm\_design} \\
\texttt{linguistic\_and\_text\_analysis} & \texttt{advanced\_writing\_and\_literature\_analysis} \\
\bottomrule
\end{longtable}
\end{sidewaystable}
\begin{sidewaystable}
\caption{(Part 6 of 6) 337 Train Skills extracted from random sample of 5200 instruction-response pairs from \textbf{Alpaca-52K (GPT-4)}}
\centering
\begin{longtable}{p{8.5cm}p{8.5cm}p{8.5cm}}
\toprule
\textbf{Skill Cluster Name} &  \\
\toprule
\texttt{language\_abilities\_and\_rewriting\_skills} & \texttt{cloud\_and\_streaming\_technology} \\
\texttt{intellectual\_comprehension\_and\_generation} & \texttt{} \\
\texttt{literacy\_and\_linguistic\_skills} & \texttt{educational\_planning\_and\_self\_assessment} \\
\texttt{specialized\_knowledge} & \texttt{literacy\_and\_language\_skills} \\
\texttt{web\_design\_and\_development} & \texttt{linguistic\_and\_literary\_skills} \\
\texttt{digital\_marketing\_and\_seo} & \texttt{python\_programming\_skills} \\
\texttt{geographical\_and\_environmental\_knowledge} & \texttt{problem\_solving\_reasoning} \\
\texttt{digital\_and\_data\_technology} & \texttt{statistical\_computation\_and\_analysis} \\
\texttt{technical\_and\_procedural\_writing} & \texttt{computer\_science\_and\_it} \\
\texttt{ai\_ml\_knowledge} & \texttt{social\_communication\_and\_awareness} \\
\texttt{research\_and\_classification\_skills} & \texttt{data\_based\_analysis} \\
\texttt{environment\_and\_life\_sciences} & \texttt{technical\_computer\_based\_proficiency} \\
\texttt{written\_communication\_and\_content\_creation} & \texttt{} \\
\texttt{natural\_language\_processing} & \texttt{digital\_marketing\_strategy} \\
\texttt{user\_interaction\_design\_and\_management} & \texttt{information\_processing\_n\_techniques} \\
\texttt{critical\_and\_ethical\_thinking} & \texttt{software\_development\_and\_engineering} \\
\texttt{digital\_modeling\_and\_design} & \texttt{language\_and\_writing\_techniques} \\
\texttt{medical\_and\_healthcare\_knowledge} & \texttt{} \\
\bottomrule
\end{longtable}
\end{sidewaystable}

\clearpage

\begin{sidewaystable}
\caption{(Part 1 of 3) 128 Train Skills extracted from random sample of 1000 instruction-response pairs from \textbf{UltraChat}}
\label{tab:list-of-skills-SDD-ultrachat}
\centering
\begin{longtable}{p{8.5cm}p{8.5cm}p{8.5cm}}
\toprule
\textbf{Skill Cluster Name} &  \\
\toprule
\texttt{cultural\_and\_societal\_understanding} & \texttt{critical\_analysis\_and\_evaluation} \\
\texttt{information\_extraction\_and\_analysis} & \texttt{creative\_production\_skills} \\
\texttt{teaching\_and\_presentation\_skills} & \texttt{specialized\_technical\_skills} \\
\texttt{management\_and\_negotiation} & \texttt{sport\_specific\_strength\_training} \\
\texttt{web\_design\_and\_development} & \texttt{writing\_and\_communication} \\
\texttt{environmental\_and\_ecological\_studies} & \texttt{data\_analysis\_and\_machine\_learning} \\
\texttt{skills\_in\_teaching\_and\_education} & \texttt{cuisine\_and\_cooking\_knowledge} \\
\bottomrule
\end{longtable}
\end{sidewaystable}
\begin{sidewaystable}
\caption{(Part 2 of 3) 128 Train Skills extracted from random sample of 1000 instruction-response pairs from \textbf{UltraChat}}
\centering
\begin{longtable}{p{8.5cm}p{8.5cm}p{8.5cm}}
\toprule
\textbf{Skill Cluster Name} &  \\
\toprule
\texttt{machine\_learning\_and\_modeling} & \texttt{cultural\_historical\_knowledge} \\
\texttt{advanced\_writing\_and\_comprehension\_skills} & \texttt{} \\
\texttt{communication\_and\_literacy} & \texttt{policy\_analysis\_and\_evaluation} \\
\texttt{cultural\_and\_social\_analysis} & \texttt{technology\_development\_and\_security} \\
\texttt{research\_and\_data\_handling} & \texttt{communication\_and\_marketing\_strategy} \\
\texttt{creative\_and\_cultural\_acumen} & \texttt{writing\_and\_communication\_skills} \\
\texttt{cultural\_studies\_and\_analysis} & \texttt{outdoor\_and\_survival\_skills} \\
\bottomrule
\end{longtable}
\end{sidewaystable}
\begin{sidewaystable}
\caption{(Part 3 of 3) 128 Train Skills extracted from random sample of 1000 instruction-response pairs from \textbf{UltraChat}}
\centering
\begin{longtable}{p{8.5cm}p{8.5cm}p{8.5cm}}
\toprule
\textbf{Skill Cluster Name} &  \\
\toprule
\texttt{research\_and\_analysis\_skills} & \texttt{digital\_media\_skillset} \\
\texttt{environmental\_sciences\_and\_gardening} & \texttt{technical\_knowledge\_and\_integration} \\
\texttt{academic\_research\_and\_analysis} & \texttt{market\_analysis\_and\_strategy} \\
\texttt{knowledge\_based\_specialization} & \texttt{writing\_and\_creative\_expression} \\
\texttt{science\_and\_environment\_understanding} & \texttt{understanding\_and\_analysis\_of\_history} \\
\bottomrule
\end{longtable}
\end{sidewaystable}

\clearpage

\begin{sidewaystable}
\caption{(Part 1 of 1) 35 Alignment + Safety Skills}
\label{tab:list-of-skills-SDD-alignment}
\centering
\begin{longtable}{p{8.5cm}p{8.5cm}p{8.5cm}}
\toprule
\textbf{Skill Cluster Name} &  \\
\toprule
\texttt{cybersecurity\_advice} & \texttt{safety\_tips} \\
\texttt{privacy\_management} & \texttt{mental\_health\_guidance} \\
\texttt{physical\_health\_advice} & \texttt{dietary\_guidance} \\
\texttt{family\_relationship\_advice} & \texttt{romantic\_relationship\_advice} \\
\texttt{friendship\_management} & \texttt{life\_decisions\_support} \\
\texttt{empowerment\_strategies} & \texttt{legal\_advice} \\
\texttt{equity\_education} & \texttt{skill\_enhancement} \\
\texttt{self\_discovery\_assistance} & \texttt{leisure\_activities\_suggestions} \\
\texttt{aesthetic\_enhancement} & \texttt{resource\_optimization} \\
\texttt{sustainability\_advice} & \texttt{career\_advancement\_guidance} \\
\texttt{social\_status\_enhancement} & \texttt{educational\_resources} \\
\texttt{critical\_thinking\_promotion} & \texttt{legal\_compliance\_stance} \\
\texttt{privacy\_policy\_explanation} & \texttt{content\_moderation\_standards} \\
\texttt{refusal\_to\_support\_illegal\_activity} & \texttt{ethical\_use\_enforcement} \\
\texttt{promotion\_of\_originality} & \texttt{legal\_ethical\_guidance} \\
\texttt{lawful\_technology\_usage\_guidance} & \texttt{misuse\_prevention\_advice} \\
\texttt{redirect\_to\_legitimate\_topic} & \texttt{sensitive\_topic\_navigation} \\
\texttt{ethical\_discussion\_fostering} & \texttt{} \\
\bottomrule
\end{longtable}
\end{sidewaystable}

\clearpage

\subsection{\instructSMSDA{} List of Skills and Query Types}
\label{appendix:list-of-skills-SDA}
Using the procedure detailed in Section~\ref{subsec:skill-extraction}, we extract 156 conversational topics and 18 query types from \gptTurbo. From the topics, we get a fine-grained list of 1,143 skills. 

\begin{small}
\begin{sidewaystable}
\caption{(Part 1 of 3) 156 topics extracted from interactions with \gptTurbo}
\label{tab:list-of-skills-SDA-topics-pt1}
\centering
\begin{longtable}{p{8.5cm}p{8.5cm}p{8.5cm}}
\toprule
\textbf{Topics} &  \\
\toprule
\texttt{disease\_symptoms} & \texttt{treatments} \\
\texttt{wellness\_tips} & \texttt{stock\_market} \\
\texttt{personal\_finance} & \texttt{corporate\_finance} \\
\texttt{physics} & \texttt{chemistry} \\
\texttt{engineering} & \texttt{information\_technology\_(it)} \\
\texttt{legislation} & \texttt{civil\_rights} \\
\texttt{public\_policy} & \texttt{music} \\
\texttt{literature} & \texttt{film} \\
\texttt{visual\_arts} & \texttt{study\_tips} \\
\texttt{educational\_theories} & \texttt{online\_courses} \\
\texttt{historical\_events} & \texttt{geographical\_facts} \\
\texttt{travel} & \texttt{hobbies} \\
\texttt{lifestyle\_choices} & \texttt{industry\_trends} \\
\texttt{leadership} & \texttt{strategy} \\
\texttt{climate\_change} & \texttt{biodiversity} \\
\texttt{sustainability} & \texttt{behavioral\_studies} \\
\texttt{social\_theories} & \texttt{mental\_health} \\
\texttt{team\_sports} & \texttt{training\_routines} \\
\texttt{sporting\_events} & \texttt{emerging\_tech} \\
\texttt{gadget\_reviews} & \texttt{software\_tutorials} \\
\texttt{parenting} & \texttt{home\_improvement} \\
\texttt{pet\_care} & \texttt{cooking} \\
\texttt{diets} & \texttt{nutritional\_info} \\
\texttt{market\_trends} & \texttt{architectural\_design} \\
\texttt{philosophical\_theories} & \texttt{world\_religions} \\
\texttt{vehicle\_maintenance} & \texttt{transport\_technology} \\
\texttt{job\_hunting} & \texttt{career\_advice} \\
\texttt{natural\_disasters} & \texttt{first\_aid} \\
\texttt{programming\_languages} & \texttt{algorithms} \\
\texttt{software\_development} & \texttt{algebra} \\
\texttt{calculus} & \texttt{statistics} \\
\texttt{language\_learning} & \texttt{grammar} \\
\bottomrule
\end{longtable}
\end{sidewaystable}
\begin{sidewaystable}
\caption{(Part 2 of 3) 156 topics extracted from interactions with \gptTurbo}
\label{tab:list-of-skills-SDA-topics-pt2}
\centering
\begin{longtable}{p{8.5cm}p{8.5cm}p{8.5cm}}
\toprule
\textbf{Topics} &  \\
\toprule
\texttt{linguistics} & \texttt{mechanical} \\
\texttt{electrical} & \texttt{civil\_engineering} \\
\texttt{media\_analysis} & \texttt{news\_reporting} \\
\texttt{digital\_media\_trends} & \texttt{disease\_prevention} \\
\texttt{public\_health\_initiatives} & \texttt{epidemiological\_studies} \\
\texttt{seo} & \texttt{social\_media\_strategies} \\
\texttt{content\_marketing} & \texttt{cultural\_dynamics} \\
\texttt{anthropology} & \texttt{social\_customs} \\
\texttt{planetary\_science} & \texttt{space\_missions} \\
\texttt{astronomy} & \texttt{theatre} \\
\texttt{dance} & \texttt{performance\_techniques} \\
\texttt{geology} & \texttt{meteorology} \\
\texttt{oceanography} & \texttt{painting} \\
\texttt{sculpture} & \texttt{diy\_crafts} \\
\texttt{literary\_analysis} & \texttt{poetry} \\
\texttt{narrative\_techniques} & \texttt{pet\_care} \\
\texttt{animal\_health} & \texttt{veterinary\_medicine} \\
\texttt{charity} & \texttt{fundraising} \\
\texttt{ngo\_management} & \texttt{diplomacy} \\
\texttt{global\_conflicts} & \texttt{international\_law} \\
\texttt{renewable\_energy} & \texttt{resource\_management} \\
\texttt{sustainability\_practices} & \texttt{fashion\_trends} \\
\texttt{textile\_manufacturing} & \texttt{design\_theory} \\
\texttt{hotel\_management} & \texttt{tourism\_trends} \\
\texttt{event\_planning} & \texttt{ethical\_dilemmas} \\
\texttt{moral\_philosophy} & \texttt{bioethics} \\
\texttt{health\_and\_medicine} & \texttt{finance\_and\_economics} \\
\texttt{science\_and\_technology} & \texttt{law\_and\_government} \\
\texttt{arts\_and\_entertainment} & \texttt{education\_and\_learning} \\
\texttt{history\_and\_geography} & \texttt{lifestyle\_and\_leisure} \\
\texttt{business\_and\_management} & \texttt{environment\_and\_ecology} \\
\texttt{psychology\_and\_sociology} & \texttt{sports\_and\_recreation} \\
\bottomrule
\end{longtable}
\end{sidewaystable}
\begin{sidewaystable}
\caption{(Part 3 of 3) 156 topics extracted from interactions with \gptTurbo}
\label{tab:list-of-skills-SDA-topics-pt3}
\centering
\begin{longtable}{p{8.5cm}p{8.5cm}p{8.5cm}}
\toprule
\textbf{Topics} &  \\
\toprule
\texttt{technology\_and\_innovation} & \texttt{home\_and\_family} \\
\texttt{food\_and\_nutrition} & \texttt{real\_estate\_and\_urban\_planning} \\
\texttt{philosophy\_and\_religion} & \texttt{transportation\_and\_automotive} \\
\texttt{career\_and\_professional\_development} & \texttt{emergency\_preparedness\_and\_response} \\
\texttt{computer\_science\_and\_programming} & \texttt{mathematics} \\
\texttt{languages\_and\_linguistics} & \texttt{engineering\_disciplines} \\
\texttt{media\_studies\_and\_journalism} & \texttt{public\_health\_and\_epidemiology} \\
\texttt{social\_media\_and\_digital\_marketing} & \texttt{cultural\_studies} \\
\texttt{astronomy\_and\_space\_exploration} & \texttt{performing\_arts} \\
\texttt{earth\_sciences} & \texttt{visual\_arts\_and\_crafts} \\
\texttt{literary\_studies} & \texttt{veterinary\_sciences} \\
\texttt{international\_relations\_and\_global\_studies} & \texttt{} \\
\texttt{philanthropy\_and\_non-profit\_sector} & \texttt{energy\_and\_resources} \\
\texttt{fashion\_and\_textiles} & \texttt{hospitality\_and\_tourism} \\
\texttt{ethics\_and\_morality} & \texttt{} \\
\bottomrule
\end{longtable}
\end{sidewaystable}

\clearpage

\begin{sidewaystable}
\caption{(Part 1 of 18) 1143 skills extracted from topics in Tables~\ref{tab:list-of-skills-SDA-topics-pt1}, \ref{tab:list-of-skills-SDA-topics-pt2}, and \ref{tab:list-of-skills-SDA-topics-pt3}.}
\label{tab:list-of-skills-SDA-skills}
\centering
\begin{longtable}{p{8.5cm}p{8.5cm}p{8.5cm}}
\toprule
\textbf{Skills} &  \\
\toprule
\texttt{marketing\_and\_promotion} & \texttt{spacecraft\_navigation} \\
\texttt{shock\_prevention} & \texttt{public\_speaking} \\
\texttt{wood\_sculpting} & \texttt{virtual\_reality\_experience} \\
\texttt{donor\_research} & \texttt{pet\_nutrition\_planning} \\
\texttt{emotive\_expression} & \texttt{global\_conflict\_analysis} \\
\texttt{production\_management} & \texttt{textile\_design\_and\_weaving} \\
\texttt{constitutional\_interpretation} & \texttt{mental\_toughness\_training} \\
\texttt{effective\_communication} & \texttt{mineral\_identification} \\
\texttt{music\_history\_research} & \texttt{photography\_skills} \\
\texttt{ai\_machine\_learning} & \texttt{time\_management\_efficiency} \\
\texttt{food\_safety\_practices} & \texttt{home\_decorating} \\
\texttt{vendor\_coordination} & \texttt{endurance\_training} \\
\texttt{manage\_dietary\_restrictions} & \texttt{textile\_finishing\_processes} \\
\texttt{preventive\_care} & \texttt{interpersonal\_communication} \\
\texttt{autonomous\_vehicle\_integration} & \texttt{cross\_cultural\_understanding} \\
\texttt{sporting\_events} & \texttt{garment\_construction} \\
\texttt{impact\_analysis} & \texttt{clinical\_pathology} \\
\texttt{psychology\_and\_sociology} & \texttt{cultural\_analysis} \\
\texttt{writing\_review\_articles} & \texttt{transport\_data\_analytics} \\
\texttt{debate\_and\_discourse} & \texttt{expense\_tracking} \\
\texttt{cpr\_execution} & \texttt{performing\_arts} \\
\texttt{acting\_techniques} & \texttt{regulatory\_compliance\_management} \\
\texttt{project\_management} & \texttt{genre\_analysis} \\
\texttt{user\_experience\_optimization} & \texttt{investment\_analysis} \\
\texttt{stock\_market} & \texttt{probability\_calculation} \\
\texttt{sustainable\_design} & \texttt{diving\_proficiency} \\
\texttt{data\_structure\_integration} & \texttt{casting\_direction} \\
\texttt{software\_testing} & \texttt{debate\_facilitation} \\
\texttt{system\_troubleshooting} & \texttt{character\_development} \\
\texttt{historical\_writing} & \texttt{stakeholder\_communication} \\
\texttt{field\_sampling} & \texttt{educational\_assessment} \\
\texttt{dance\_choreography} & \texttt{network\_security} \\
\bottomrule
\end{longtable}
\end{sidewaystable}
\begin{sidewaystable}
\caption{(Part 2 of 18) 1143 skills extracted from topics in Tables~\ref{tab:list-of-skills-SDA-topics-pt1}, \ref{tab:list-of-skills-SDA-topics-pt2}, and \ref{tab:list-of-skills-SDA-topics-pt3}.}
\centering
\begin{longtable}{p{8.5cm}p{8.5cm}p{8.5cm}}
\toprule
\textbf{Skills} &  \\
\toprule
\texttt{visual\_storytelling} & \texttt{green\_infrastructure\_design} \\
\texttt{numerical\_computation} & \texttt{3d\_modeling} \\
\texttt{historical\_linguistics\_research} & \texttt{stakeholder\_engagement} \\
\texttt{songwriting\_composition} & \texttt{pharmacological\_knowledge} \\
\texttt{glass\_etching} & \texttt{media\_experimentation} \\
\texttt{hydraulic\_engineering} & \texttt{research\_techniques} \\
\texttt{bioarchaeological\_analysis} & \texttt{quantitative\_reasoning} \\
\texttt{rhythm\_identification} & \texttt{quantum\_computing} \\
\texttt{recommend\_balanced\_meals} & \texttt{study\_design\_and\_conduct} \\
\texttt{course\_design} & \texttt{celestial\_navigation} \\
\texttt{multimedia\_content\_creation} & \texttt{analyze\_nutrient\_content} \\
\texttt{food\_preparation\_techniques} & \texttt{project\_collaboration} \\
\texttt{ingredient\_substitution} & \texttt{software\_development} \\
\texttt{stage\_management} & \texttt{cultural\_sensitivity} \\
\texttt{ethics\_in\_research} & \texttt{impact\_evaluation} \\
\texttt{valuation\_techniques} & \texttt{sustainable\_design\_principles} \\
\texttt{cloud\_solution\_architecture} & \texttt{cognitive\_behavioral\_management} \\
\texttt{peacebuilding\_strategies} & \texttt{impact\_measurement} \\
\texttt{public\_outreach\_and\_education} & \texttt{engagement\_strategies} \\
\texttt{adapting\_communication\_styles} & \texttt{renewable\_energy\_systems} \\
\texttt{creative\_visualization} & \texttt{comparative\_religion\_study} \\
\texttt{style\_consistency\_maintenance} & \texttt{social\_media\_strategy} \\
\texttt{printmaking\_techniques} & \texttt{film\_editing} \\
\texttt{evacuation\_procedures} & \texttt{trend\_monitoring} \\
\texttt{behavioral\_analysis} & \texttt{educate\_on\_preventive\_measures} \\
\texttt{narrative\_technique\_evaluation} & \texttt{injury\_prevention} \\
\texttt{grammar\_proficiency} & \texttt{resource\_conservation\_strategies} \\
\texttt{stress\_management\_training} & \texttt{civil\_rights} \\
\texttt{dispute\_resolution} & \texttt{world\_religions} \\
\texttt{resilience\_building} & \texttt{financial\_budgeting} \\
\texttt{ocean\_modelling} & \texttt{economic\_modeling} \\
\texttt{motor\_control} & \texttt{business\_and\_management} \\
\bottomrule
\end{longtable}
\end{sidewaystable}
\begin{sidewaystable}
\caption{(Part 3 of 18) 1143 skills extracted from topics in Tables~\ref{tab:list-of-skills-SDA-topics-pt1}, \ref{tab:list-of-skills-SDA-topics-pt2}, and \ref{tab:list-of-skills-SDA-topics-pt3}.}
\centering
\begin{longtable}{p{8.5cm}p{8.5cm}p{8.5cm}}
\toprule
\textbf{Skills} &  \\
\toprule
\texttt{brush\_stroke\_mastery} & \texttt{digital\_advertising} \\
\texttt{conceptual\_analysis} & \texttt{quality\_control\_inspection} \\
\texttt{curriculum\_design} & \texttt{water\_resources\_management} \\
\texttt{circuit\_design} & \texttt{geochemical\_sampling} \\
\texttt{trend\_identification} & \texttt{grant\_writing\_and\_funding\_acquisition} \\
\texttt{geographical\_mapping} & \texttt{mission\_planning} \\
\texttt{complex\_sentence\_forming} & \texttt{visual\_arts\_and\_crafts} \\
\texttt{nutrition\_planning} & \texttt{computer\_aided\_design} \\
\texttt{job\_search\_techniques} & \texttt{ethical\_reflection} \\
\texttt{interview\_techniques} & \texttt{sewing\_techniques} \\
\texttt{energy\_management} & \texttt{computational\_linguistics\_application} \\
\texttt{recipe\_development} & \texttt{algorithm\_design} \\
\texttt{student\_motivation} & \texttt{empathetic\_listening} \\
\texttt{quality\_control\_management} & \texttt{hydration\_nudges} \\
\texttt{manual\_dexterity} & \texttt{laboratory\_safety} \\
\texttt{lifestyle\_and\_leisure} & \texttt{budget\_management} \\
\texttt{educational\_research\_methods} & \texttt{renewable\_energy\_technology} \\
\texttt{approximating\_functions} & \texttt{field\_mapping} \\
\texttt{using\_theorems\_in\_calculus} & \texttt{religious\_impact\_assessment} \\
\texttt{home\_safety\_procedures} & \texttt{healthy\_cooking} \\
\texttt{identify\_common\_symptoms} & \texttt{environment\_and\_ecology} \\
\texttt{cross\_platform\_development} & \texttt{event\_planning} \\
\texttt{retirement\_planning} & \texttt{radiology\_technique} \\
\texttt{innovation\_creativity} & \texttt{writing\_clarity} \\
\texttt{costume\_design} & \texttt{space\_awareness} \\
\texttt{geometric\_visualization} & \texttt{learner\_engagement} \\
\texttt{orbital\_mechanics} & \texttt{facilitating\_inclusive\_conversations} \\
\texttt{roof\_repair\_and\_installation} & \texttt{culinary\_skills} \\
\texttt{job\_search\_strategies} & \texttt{seo\_strategy\_planning} \\
\texttt{donor\_relations} & \texttt{multilingual\_communication} \\
\texttt{veterinary\_treatment} & \texttt{on\_page\_optimization} \\
\texttt{historical\_contextualization} & \texttt{data\_management\_and\_analysis} \\
\bottomrule
\end{longtable}
\end{sidewaystable}
\begin{sidewaystable}
\caption{(Part 4 of 18) 1143 skills extracted from topics in Tables~\ref{tab:list-of-skills-SDA-topics-pt1}, \ref{tab:list-of-skills-SDA-topics-pt2}, and \ref{tab:list-of-skills-SDA-topics-pt3}.}
\centering
\begin{longtable}{p{8.5cm}p{8.5cm}p{8.5cm}}
\toprule
\textbf{Skills} &  \\
\toprule
\texttt{strategic\_thinking} & \texttt{energy\_efficiency\_audit} \\
\texttt{performance\_artistry} & \texttt{stoichiometry\_calculation} \\
\texttt{budget\_creation} & \texttt{crochet\_knitting} \\
\texttt{sustainable\_practices\_implementation} & \texttt{healthy\_eating\_habits} \\
\texttt{api\_integration} & \texttt{legal\_writing} \\
\texttt{augmented\_reality\_creation} & \texttt{spark\_plug\_replacement} \\
\texttt{policy\_analysis} & \texttt{pedagogical\_design} \\
\texttt{air\_filter\_change} & \texttt{reflective\_practice} \\
\texttt{visual\_design} & \texttt{network\_configuration} \\
\texttt{spectral\_analysis} & \texttt{environmental\_impact\_assessment} \\
\texttt{physical\_conditioning\_for\_performance} & \texttt{version\_control} \\
\texttt{elder\_care\_knowledge} & \texttt{poetry\_workshopping} \\
\texttt{financial\_modeling} & \texttt{data\_visualization} \\
\texttt{global\_strategy\_planning} & \texttt{law\_and\_government} \\
\texttt{moral\_courage} & \texttt{machine\_learning\_integration} \\
\texttt{climate\_data\_analysis} & \texttt{robotics\_integration} \\
\texttt{using\_matrices\_for\_transformations} & \texttt{stakeholder\_engagement\_in\_sustainability} \\
\texttt{financial\_reporting} & \texttt{landscaping\_design} \\
\texttt{customer\_insight\_analysis} & \texttt{battery\_maintenance} \\
\texttt{database\_design} & \texttt{caloric\_management} \\
\texttt{historical\_research} & \texttt{screen\_time\_management} \\
\texttt{instructional\_materials\_development} & \texttt{water\_conservation\_techniques} \\
\texttt{animal\_welfare\_compliance} & \texttt{theoretical\_model\_application} \\
\texttt{jewelry\_making} & \texttt{language\_translation} \\
\texttt{financial\_analysis} & \texttt{public\_health\_communication} \\
\texttt{performance\_techniques} & \texttt{brand\_alignment} \\
\texttt{health\_monitoring\_procedures} & \texttt{nutrition\_management} \\
\texttt{health\_policy\_advocacy} & \texttt{treatment\_plan\_design} \\
\texttt{audience\_analysis} & \texttt{conflict\_resolution\_techniques} \\
\texttt{ceramics\_pottery} & \texttt{orbit\_dynamics} \\
\texttt{climate\_change\_adaptation} & \texttt{water\_resource\_management} \\
\texttt{emotional\_regulation} & \texttt{diagnostic\_testing} \\
\bottomrule
\end{longtable}
\end{sidewaystable}
\begin{sidewaystable}
\caption{(Part 5 of 18) 1143 skills extracted from topics in Tables~\ref{tab:list-of-skills-SDA-topics-pt1}, \ref{tab:list-of-skills-SDA-topics-pt2}, and \ref{tab:list-of-skills-SDA-topics-pt3}.}
\centering
\begin{longtable}{p{8.5cm}p{8.5cm}p{8.5cm}}
\toprule
\textbf{Skills} &  \\
\toprule
\texttt{digital\_proficiency} & \texttt{star\_identification} \\
\texttt{bronze\_casting} & \texttt{focus\_enhancement} \\
\texttt{brand\_storytelling} & \texttt{ui\_ux\_design} \\
\texttt{grammar\_rules\_teaching} & \texttt{outcome\_evaluation} \\
\texttt{lesson\_planning} & \texttt{satellite\_communications} \\
\texttt{map\_reading} & \texttt{adaptive\_learning} \\
\texttt{dietary\_adaptation} & \texttt{mold\_making} \\
\texttt{quality\_control\_in\_textiles} & \texttt{water\_quality\_assessment} \\
\texttt{grocery\_shopping\_optimization} & \texttt{social\_responsibility} \\
\texttt{rock\_identification} & \texttt{infrastructural\_health\_monitoring} \\
\texttt{understanding\_cultural\_nuances} & \texttt{mental\_health\_support} \\
\texttt{prop\_management} & \texttt{disease\_diagnosis} \\
\texttt{fluid\_dynamics\_analysis} & \texttt{wildlife\_conservation} \\
\texttt{ecosystem\_management} & \texttt{historical\_analysis} \\
\texttt{electrical\_wiring} & \texttt{fuel\_efficiency\_optimization} \\
\texttt{art\_portfolio\_management} & \texttt{waste\_management} \\
\texttt{education\_and\_learning} & \texttt{visual\_arts} \\
\texttt{marketing\_and\_promotions} & \texttt{research\_methodology\_application} \\
\texttt{salary\_negotiation} & \texttt{cover\_letter\_crafting} \\
\texttt{computer\_simulation} & \texttt{community\_stakeholder\_engagement} \\
\texttt{environmental\_education} & \texttt{chronological\_reasoning} \\
\texttt{fabric\_analysis} & \texttt{nutritional\_counseling} \\
\texttt{skill\_drills\_execution} & \texttt{hazard\_identification} \\
\texttt{emergency\_preparedness\_and\_response} & \texttt{critical\_thinking} \\
\texttt{script\_writing} & \texttt{experimental\_design} \\
\texttt{behavioral\_intervention\_strategy} & \texttt{financial\_risk\_assessment} \\
\texttt{grant\_writing} & \texttt{composition\_design} \\
\texttt{market\_research} & \texttt{seismic\_analysis} \\
\texttt{fitness\_routine\_development} & \texttt{survey\_construction} \\
\texttt{thematic\_analysis} & \texttt{differentiated\_instruction} \\
\texttt{chemical\_synthesis} & \texttt{technology\_and\_innovation} \\
\texttt{maintenance\_and\_repair} & \texttt{exploring\_complex\_numbers} \\
\bottomrule
\end{longtable}
\end{sidewaystable}
\begin{sidewaystable}
\caption{(Part 6 of 18) 1143 skills extracted from topics in Tables~\ref{tab:list-of-skills-SDA-topics-pt1}, \ref{tab:list-of-skills-SDA-topics-pt2}, and \ref{tab:list-of-skills-SDA-topics-pt3}.}
\centering
\begin{longtable}{p{8.5cm}p{8.5cm}p{8.5cm}}
\toprule
\textbf{Skills} &  \\
\toprule
\texttt{sustainability\_integration} & \texttt{dance\_and\_movement} \\
\texttt{green\_energy\_solutions} & \texttt{algebraic\_manipulation} \\
\texttt{public\_health\_analysis} & \texttt{budget\_planning} \\
\texttt{close\_reading} & \texttt{personal\_branding} \\
\texttt{sanitation\_protocol\_implementation} & \texttt{character\_analysis} \\
\texttt{physical\_expressions} & \texttt{unit\_testing} \\
\texttt{test\_preparation} & \texttt{strategic\_allocation} \\
\texttt{gadget\_reviews} & \texttt{construction\_technology\_integration} \\
\texttt{choking\_remedy} & \texttt{attention\_to\_detail} \\
\texttt{progress\_tracking} & \texttt{cooking\_techniques} \\
\texttt{carbon\_footprint\_analysis} & \texttt{spectroscopic\_analysis} \\
\texttt{literary\_analysis} & \texttt{geophysical\_surveying} \\
\texttt{behavioral\_training\_methods} & \texttt{ethical\_guidance} \\
\texttt{resume\_writing} & \texttt{performance\_evaluation} \\
\texttt{event\_correlation} & \texttt{resource\_management} \\
\texttt{crypto\_algorithm\_application} & \texttt{mergers\_and\_acquisitions\_strategy} \\
\texttt{donor\_relationship\_management} & \texttt{character\_development\_insight} \\
\texttt{digital\_literacy} & \texttt{artistic\_expression} \\
\texttt{operational\_efficiency} & \texttt{calculate\_caloric\_intake} \\
\texttt{development\_project\_management} & \texttt{coordination\_agility} \\
\texttt{fundraising\_strategy\_development} & \texttt{detail\_attention} \\
\texttt{social\_media\_outreach} & \texttt{technical\_writing} \\
\texttt{conflict\_resolution\_in\_multicultural\_contexts} & \texttt{} \\
\texttt{language\_teaching} & \texttt{digital\_marketing} \\
\texttt{problem\_solving} & \texttt{architectural\_design} \\
\texttt{mentorship\_and\_coaching} & \texttt{technical\_skill\_enhancement} \\
\texttt{animal\_health} & \texttt{sustainable\_land\_use\_planning} \\
\texttt{telescope\_operation} & \texttt{public\_health\_and\_epidemiology} \\
\texttt{information\_synthesis} & \texttt{media\_analysis} \\
\texttt{negotiation\_skills} & \texttt{space\_weather\_forecasting} \\
\texttt{speech\_recognition\_development} & \texttt{body\_coordination} \\
\texttt{argument\_development} & \texttt{zoning\_regulations\_compliance} \\
\bottomrule
\end{longtable}
\end{sidewaystable}
\begin{sidewaystable}
\caption{(Part 7 of 18) 1143 skills extracted from topics in Tables~\ref{tab:list-of-skills-SDA-topics-pt1}, \ref{tab:list-of-skills-SDA-topics-pt2}, and \ref{tab:list-of-skills-SDA-topics-pt3}.}
\centering
\begin{longtable}{p{8.5cm}p{8.5cm}p{8.5cm}}
\toprule
\textbf{Skills} &  \\
\toprule
\texttt{resource\_optimization} & \texttt{global\_conflicts} \\
\texttt{sociolinguistic\_survey\_conducting} & \texttt{real\_estate\_and\_urban\_planning} \\
\texttt{climate\_analysis} & \texttt{database\_management} \\
\texttt{script\_analysis\_and\_interpretation} & \texttt{design\_theory} \\
\texttt{patient\_communication} & \texttt{trading\_strategies\_implementation} \\
\texttt{campaign\_management} & \texttt{adaptive\_learning\_techniques} \\
\texttt{market\_timing} & \texttt{landscape\_design} \\
\texttt{user\_interface\_design} & \texttt{classroom\_management} \\
\texttt{historical\_preservation} & \texttt{navigation\_expertise} \\
\texttt{goal\_setting} & \texttt{version\_control\_management} \\
\texttt{basic\_sewing} & \texttt{nutrition\_education} \\
\texttt{literary\_device\_application} & \texttt{geotechnical\_engineering} \\
\texttt{dance\_technique\_improvement} & \texttt{analytics\_monitoring} \\
\texttt{ethics\_and\_morality} & \texttt{phonetic\_transcription} \\
\texttt{textual\_interpretation} & \texttt{sociological\_analysis} \\
\texttt{brand\_management} & \texttt{emerging\_tech} \\
\texttt{historical\_events} & \texttt{ethical\_reporting} \\
\texttt{change\_management} & \texttt{surgical\_techniques} \\
\texttt{software\_debugging} & \texttt{system\_administration} \\
\texttt{astronomy\_and\_space\_exploration} & \texttt{oil\_change} \\
\texttt{project\_financial\_modeling} & \texttt{improvisation\_techniques} \\
\texttt{costume\_management} & \texttt{pronunciation\_accuracy} \\
\texttt{fluid\_checks} & \texttt{story\_pitching} \\
\texttt{content\_creation} & \texttt{dental\_care} \\
\texttt{team\_management} & \texttt{community\_engagement} \\
\texttt{narrative\_construction} & \texttt{foreign\_language\_proficiency} \\
\texttt{statutory\_analysis} & \texttt{jewelry\_making\_techniques} \\
\texttt{voice\_projection\_training} & \texttt{philanthropy\_and\_non\_profit\_sector} \\
\texttt{satellite\_imagery\_interpretation} & \texttt{sustainable\_development\_planning} \\
\texttt{data\_driven\_decision\_making} & \texttt{persuasive\_writing} \\
\texttt{sustainability\_reporting} & \texttt{automotive\_engineering} \\
\texttt{parallel\_processing\_design} & \texttt{security\_implementation} \\
\bottomrule
\end{longtable}
\end{sidewaystable}
\begin{sidewaystable}
\caption{(Part 8 of 18) 1143 skills extracted from topics in Tables~\ref{tab:list-of-skills-SDA-topics-pt1}, \ref{tab:list-of-skills-SDA-topics-pt2}, and \ref{tab:list-of-skills-SDA-topics-pt3}.}
\centering
\begin{longtable}{p{8.5cm}p{8.5cm}p{8.5cm}}
\toprule
\textbf{Skills} &  \\
\toprule
\texttt{customer\_segmentation} & \texttt{logical\_reasoning} \\
\texttt{style\_advisory} & \texttt{renewable\_energy\_research} \\
\texttt{educate\_on\_portion\_control} & \texttt{job\_hunting} \\
\texttt{performance\_monitoring} & \texttt{platform\_specific\_techniques} \\
\texttt{food\_and\_beverage\_management} & \texttt{ethnographic\_research} \\
\texttt{animal\_diagnosis} & \texttt{international\_litigation} \\
\texttt{upcycling\_projects} & \texttt{epidemiological\_modeling} \\
\texttt{radar\_technology\_use} & \texttt{environmental\_policy\_analysis} \\
\texttt{community\_outreach} & \texttt{automotive\_design\_and\_aerodynamics} \\
\texttt{building\_codes\_compliance} & \texttt{chemical\_reactivity\_prediction} \\
\texttt{argument\_analysis} & \texttt{quantitative\_modeling} \\
\texttt{ceramic\_craftsmanship} & \texttt{habit\_forming\_tips} \\
\texttt{voice\_projection} & \texttt{set\_design} \\
\texttt{social\_perception\_analysis} & \texttt{environmental\_impact\_reduction} \\
\texttt{educational\_programming} & \texttt{interpretation\_of\_symbolism} \\
\texttt{surgical\_procedures} & \texttt{networking\_strategies} \\
\texttt{patient\_monitoring} & \texttt{capital\_budgeting} \\
\texttt{story\_structure\_analysis} & \texttt{poetry\_performance} \\
\texttt{investment\_strategy} & \texttt{construction\_estimation} \\
\texttt{style\_adaptation} & \texttt{analyzing\_series\_and\_sequences} \\
\texttt{geopolitical\_analysis} & \texttt{cultural\_sensitivity\_training} \\
\texttt{risk\_management} & \texttt{language\_exchange\_fostering} \\
\texttt{public\_education\_and\_outreach} & \texttt{thermal\_management} \\
\texttt{factoring\_polynomials} & \texttt{physical\_fitness\_routine} \\
\texttt{decision\_making\_under\_uncertainty} & \texttt{climate\_change} \\
\texttt{real\_estate\_financing} & \texttt{suicide\_prevention} \\
\texttt{adjective\_adverb\_usage} & \texttt{digital\_storytelling} \\
\texttt{design\_conceptualization} & \texttt{humanitarian\_intervention\_strategy} \\
\texttt{paper\_crafting} & \texttt{scriptural\_interpretation} \\
\texttt{compliance\_management} & \texttt{emergency\_preparedness} \\
\texttt{cultural\_interpretation} & \texttt{trend\_analysis} \\
\texttt{animal\_grooming\_techniques} & \texttt{astronomical\_photography} \\
\bottomrule
\end{longtable}
\end{sidewaystable}
\begin{sidewaystable}
\caption{(Part 9 of 18) 1143 skills extracted from topics in Tables~\ref{tab:list-of-skills-SDA-topics-pt1}, \ref{tab:list-of-skills-SDA-topics-pt2}, and \ref{tab:list-of-skills-SDA-topics-pt3}.}
\centering
\begin{longtable}{p{8.5cm}p{8.5cm}p{8.5cm}}
\toprule
\textbf{Skills} &  \\
\toprule
\texttt{research\_design} & \texttt{law\_enforcement\_compliance} \\
\texttt{music\_performance} & \texttt{social\_media\_strategies} \\
\texttt{non\_profit\_governance} & \texttt{visual\_communication\_skills} \\
\texttt{film\_critique} & \texttt{performance\_metrics\_analysis} \\
\texttt{coding\_proficiency} & \texttt{communication\_protocol\_design} \\
\texttt{data\_management} & \texttt{digital\_ethics\_management} \\
\texttt{negotiating\_conflict\_resolution} & \texttt{machine\_maintenance} \\
\texttt{marketing\_strategy} & \texttt{interfaith\_communication} \\
\texttt{vision\_formulation} & \texttt{digital\_media\_management} \\
\texttt{customer\_service\_management} & \texttt{empathetic\_communication} \\
\texttt{medical\_research} & \texttt{linguistic\_analysis} \\
\texttt{charity\_marketing} & \texttt{crisis\_management} \\
\texttt{paleontological\_excavation} & \texttt{home\_decor\_crafting} \\
\texttt{legal\_research} & \texttt{finance\_and\_economics} \\
\texttt{leadership\_development} & \texttt{canvas\_preparation} \\
\texttt{atmospheric\_modeling} & \texttt{genetic\_diversity\_analysis} \\
\texttt{textile\_designing} & \texttt{science\_and\_technology} \\
\texttt{directorial\_decisions} & \texttt{renewable\_energy\_integration} \\
\texttt{data\_analysis\_chemistry} & \texttt{listening\_comprehension} \\
\texttt{semantic\_analysis} & \texttt{teaching\_strategies\_implementation} \\
\texttt{investigative\_research} & \texttt{stage\_presence} \\
\texttt{verse\_crafting} & \texttt{meal\_planning} \\
\texttt{dietary\_analysis} & \texttt{event\_planning\_and\_coordination} \\
\texttt{recovery\_operations} & \texttt{argumentative\_writing} \\
\texttt{health\_and\_medicine} & \texttt{fashion\_illustration} \\
\texttt{note\_taking} & \texttt{international\_trade\_management} \\
\texttt{legislative\_negotiation} & \texttt{nutritional\_advising} \\
\texttt{equipment\_maintenance} & \texttt{seo\_audit} \\
\texttt{policy\_formulation} & \texttt{educational\_technology\_integration} \\
\texttt{emergency\_responding} & \texttt{tourism\_trends} \\
\texttt{humanitarian\_response} & \texttt{land\_use\_planning} \\
\texttt{soil\_testing} & \texttt{empathetic\_understanding} \\
\bottomrule
\end{longtable}
\end{sidewaystable}
\begin{sidewaystable}
\caption{(Part 10 of 18) 1143 skills extracted from topics in Tables~\ref{tab:list-of-skills-SDA-topics-pt1}, \ref{tab:list-of-skills-SDA-topics-pt2}, and \ref{tab:list-of-skills-SDA-topics-pt3}.}
\centering
\begin{longtable}{p{8.5cm}p{8.5cm}p{8.5cm}}
\toprule
\textbf{Skills} &  \\
\toprule
\texttt{pattern\_making} & \texttt{music\_teaching} \\
\texttt{interdisciplinary\_integration} & \texttt{graphing\_functions} \\
\texttt{urban\_design\_principles} & \texttt{code\_execution} \\
\texttt{data\_analysis} & \texttt{statistical\_inference} \\
\texttt{audience\_engagement} & \texttt{fitness\_program\_design} \\
\texttt{energy\_management\_analysis} & \texttt{pollution\_control} \\
\texttt{sports\_marketing} & \texttt{transportation\_and\_automotive} \\
\texttt{risk\_assessment} & \texttt{differentiate\_similar\_symptoms} \\
\texttt{aerospace\_engineering} & \texttt{automotive\_safety\_standards\_compliance} \\
\texttt{staff\_training\_and\_development} & \texttt{team\_collaboration} \\
\texttt{noun\_verb\_agreement} & \texttt{printmaking\_methods} \\
\texttt{cultural\_integration\_facilitation} & \texttt{verb\_tense\_consistency} \\
\texttt{decision\_making} & \texttt{wildlife\_monitoring} \\
\texttt{intercultural\_competency} & \texttt{apply\_first\_aid\_for\_symptomatic\_relief} \\
\texttt{lighting\_design} & \texttt{brake\_replacement} \\
\texttt{reading\_comprehension} & \texttt{kitchen\_safety} \\
\texttt{bill\_drafting} & \texttt{technical\_seo} \\
\texttt{game\_strategy\_development} & \texttt{belt\_inspection\_replacement} \\
\texttt{wellness\_counseling} & \texttt{improvisational\_skills} \\
\texttt{platform\_navigation} & \texttt{public\_health\_surveillance} \\
\texttt{system\_architecture\_design} & \texttt{policy\_advocacy} \\
\texttt{punctuation\_rules\_application} & \texttt{solar\_panel\_installation} \\
\texttt{event\_planning\_and\_management} & \texttt{disaster\_response} \\
\texttt{supply\_chain\_management} & \texttt{drawing\_techniques} \\
\texttt{astrobiology\_research} & \texttt{allergy\_management} \\
\texttt{work\_life\_balance\_tips} & \texttt{software\_documentation} \\
\texttt{corpus\_compilation} & \texttt{semantic\_interpretation} \\
\texttt{improvisational\_technique} & \texttt{ethical\_reasoning\_in\_religion} \\
\texttt{astro\_photography} & \texttt{critical\_reading} \\
\texttt{stone\_carving} & \texttt{environmental\_compliance} \\
\texttt{energy\_efficiency\_upgrades} & \texttt{volunteer\_coordination} \\
\texttt{photographic\_composition} & \texttt{cybersecurity\_practices} \\
\bottomrule
\end{longtable}
\end{sidewaystable}
\begin{sidewaystable}
\caption{(Part 11 of 18) 1143 skills extracted from topics in Tables~\ref{tab:list-of-skills-SDA-topics-pt1}, \ref{tab:list-of-skills-SDA-topics-pt2}, and \ref{tab:list-of-skills-SDA-topics-pt3}.}
\centering
\begin{longtable}{p{8.5cm}p{8.5cm}p{8.5cm}}
\toprule
\textbf{Skills} &  \\
\toprule
\texttt{moral\_philosophy} & \texttt{cultural\_competency\_development} \\
\texttt{event\_analysis} & \texttt{diplomatic\_negotiation} \\
\texttt{fashion\_forecasting} & \texttt{emergency\_planning} \\
\texttt{correct\_sentence\_structuring} & \texttt{community\_education} \\
\texttt{empathy\_development} & \texttt{cultural\_studies} \\
\texttt{technical\_proficiency} & \texttt{wound\_management} \\
\texttt{dietary\_trend\_analysis} & \texttt{cyber\_security\_essentials} \\
\texttt{interpret\_food\_labels} & \texttt{moral\_reasoning} \\
\texttt{eco\_friendly\_materials\_development} & \texttt{titration\_techniques} \\
\texttt{home\_repair\_basics} & \texttt{arts\_and\_entertainment} \\
\texttt{food\_preparation} & \texttt{career\_coaching} \\
\texttt{debate\_and\_discussion} & \texttt{social\_connections\_fostering} \\
\texttt{ethical\_reasoning} & \texttt{equity\_financing} \\
\texttt{poetic\_interpretation} & \texttt{revenue\_management} \\
\texttt{tire\_rotation} & \texttt{news\_writing} \\
\texttt{athletic\_training} & \texttt{child\_care\_expertise} \\
\texttt{molecular\_modeling} & \texttt{algorithm\_visualization} \\
\texttt{kitchen\_equipment\_use} & \texttt{exercise\_routine\_design} \\
\texttt{report\_writing} & \texttt{api\_development} \\
\texttt{team\_leadership} & \texttt{data\_collection\_and\_management} \\
\texttt{language\_documentation} & \texttt{practicing\_politeness\_forms} \\
\texttt{mineral\_analysis} & \texttt{emotional\_intelligence\_management} \\
\texttt{data\_collection\_analysis} & \texttt{structural\_analysis} \\
\texttt{art\_history\_analysis} & \texttt{narrative\_building} \\
\texttt{cross\_cultural\_communication} & \texttt{fundamental\_analysis} \\
\texttt{media\_literacy} & \texttt{fashion\_and\_textiles} \\
\texttt{woodworking\_basics} & \texttt{emotional\_expression\_mastering} \\
\texttt{behavioral\_counseling} & \texttt{security\_practices} \\
\texttt{natural\_disasters} & \texttt{hypothesis\_testing} \\
\texttt{customer\_engagement} & \texttt{content\_analysis} \\
\texttt{fracture\_stabilization} & \texttt{data\_visualization} \\
\texttt{behavioral\_training} & \texttt{cash\_flow\_forecasting} \\
\bottomrule
\end{longtable}
\end{sidewaystable}
\begin{sidewaystable}
\caption{(Part 12 of 18) 1143 skills extracted from topics in Tables~\ref{tab:list-of-skills-SDA-topics-pt1}, \ref{tab:list-of-skills-SDA-topics-pt2}, and \ref{tab:list-of-skills-SDA-topics-pt3}.}
\centering
\begin{longtable}{p{8.5cm}p{8.5cm}p{8.5cm}}
\toprule
\textbf{Skills} &  \\
\toprule
\texttt{patience\_cultivation} & \texttt{literature\_review\_and\_meta\_analysis} \\
\texttt{networking\_skills} & \texttt{conversation\_management} \\
\texttt{resource\_utilization} & \texttt{nutritional\_info} \\
\texttt{communicate\_symptoms\_to\_healthcare\_providers} & \texttt{} \\
\texttt{digital\_fundraising\_techniques} & \texttt{emergency\_care\_practices} \\
\texttt{supply\_chain\_logistics\_for\_automotive\_parts} & \texttt{} \\
\texttt{medication\_management} & \texttt{ethics\_in\_social\_sciences} \\
\texttt{policy\_drafting} & \texttt{impact\_assessment} \\
\texttt{vaccine\_administration} & \texttt{error\_debugging} \\
\texttt{event\_technology\_utilization} & \texttt{epidemiological\_research} \\
\texttt{spectroscopic\_techniques} & \texttt{strategic\_investment\_decision\_making} \\
\texttt{application\_follow\_up} & \texttt{conflict\_resolution\_skills} \\
\texttt{fiber\_identification} & \texttt{study\_tips} \\
\texttt{digital\_design} & \texttt{strategic\_planning} \\
\texttt{online\_marketing} & \texttt{legislative\_research} \\
\texttt{philosophical\_inquiry\_in\_religion} & \texttt{mathematical\_modelling} \\
\texttt{sustainability\_practices} & \texttt{fabric\_dyeing} \\
\texttt{research\_ethics} & \texttt{local\_cuisine\_exploration} \\
\texttt{error\_handling} & \texttt{tax\_planning} \\
\texttt{sound\_design} & \texttt{cognitive\_behavioral\_therapy} \\
\texttt{group\_dynamics\_management} & \texttt{internet\_of\_things\_integration} \\
\texttt{statistical\_modeling} & \texttt{applying\_limit\_concepts} \\
\texttt{crisis\_intervention} & \texttt{electric\_vehicle\_technology} \\
\texttt{corporate\_tax\_planning} & \texttt{habitat\_restoration} \\
\texttt{hospitality\_marketing} & \texttt{algorithm\_optimization} \\
\texttt{theme\_design} & \texttt{music\_criticism} \\
\texttt{garment\_design} & \texttt{sustainable\_eating\_practices} \\
\texttt{motivational\_coaching} & \texttt{welding\_techniques} \\
\texttt{anesthesia\_management} & \texttt{color\_mixing} \\
\texttt{sponsorship\_acquisition} & \texttt{surgical\_procedure\_execution} \\
\texttt{ethical\_leadership} & \texttt{meditative\_practices} \\
\texttt{statistical\_analysis} & \texttt{data\_science\_analytics} \\
\bottomrule
\end{longtable}
\end{sidewaystable}
\begin{sidewaystable}
\caption{(Part 13 of 18) 1143 skills extracted from topics in Tables~\ref{tab:list-of-skills-SDA-topics-pt1}, \ref{tab:list-of-skills-SDA-topics-pt2}, and \ref{tab:list-of-skills-SDA-topics-pt3}.}
\centering
\begin{longtable}{p{8.5cm}p{8.5cm}p{8.5cm}}
\toprule
\textbf{Skills} &  \\
\toprule
\texttt{learning\_environment\_optimization} & \texttt{debugging\_algorithms} \\
\texttt{geospatial\_analysis} & \texttt{precision\_machining} \\
\texttt{platform\_optimization} & \texttt{technical\_drawing} \\
\texttt{burn\_treatment} & \texttt{relationship\_building} \\
\texttt{ingredient\_selection} & \texttt{epidemiology\_research} \\
\texttt{seo\_optimization} & \texttt{version\_control\_management} \\
\texttt{nutritional\_planning} & \texttt{applying\_the\_quadratic\_formula} \\
\texttt{physical\_endurance} & \texttt{cybersecurity\_analysis} \\
\texttt{big\_data\_handling} & \texttt{environmental\_health\_assessment} \\
\texttt{mental\_wellness\_guidance} & \texttt{community\_health\_mobilization} \\
\texttt{food\_and\_nutrition} & \texttt{protective\_finishing} \\
\texttt{health\_equity\_and\_access\_analysis} & \texttt{public\_administration} \\
\texttt{international\_relations\_and\_global\_studies} & \texttt{} \\
\texttt{strategic\_communication} & \texttt{spiritual\_counseling\_skills} \\
\texttt{sculpting\_methods} & \texttt{time\_management} \\
\texttt{user\_experience\_design} & \texttt{corporate\_social\_responsibility} \\
\texttt{spacecraft\_design} & \texttt{nutritional\_analysis} \\
\texttt{disaster\_preparedness} & \texttt{sustainable\_transport\_planning} \\
\texttt{narrative\_pacing\_control} & \texttt{partner\_synchronization} \\
\texttt{contextual\_historical\_analysis} & \texttt{artifact\_analysis} \\
\texttt{performance\_analysis} & \texttt{cyber\_security\_analysis} \\
\texttt{religious\_literacy\_development} & \texttt{educational\_outreach} \\
\texttt{conflict\_sensitive\_reporting} & \texttt{international\_law\_compliance} \\
\texttt{payload\_management} & \texttt{risk\_assessment\_analysis} \\
\texttt{diagnosis\_identification} & \texttt{event\_planning\_fundamentals} \\
\texttt{vehicle\_maintenance\_and\_repair} & \texttt{solving\_differential\_equations} \\
\texttt{route\_planning} & \texttt{expressive\_performance} \\
\texttt{teaching\_strategies} & \texttt{creative\_thinking} \\
\texttt{customer\_experience\_management} & \texttt{interview\_preparation} \\
\texttt{trajectory\_design} & \texttt{point\_of\_view\_selection} \\
\texttt{track\_symptom\_progression} & \texttt{international\_negotiation\_techniques} \\
\texttt{carpentry\_work} & \texttt{wind\_turbine\_maintenance} \\
\bottomrule
\end{longtable}
\end{sidewaystable}
\begin{sidewaystable}
\caption{(Part 14 of 18) 1143 skills extracted from topics in Tables~\ref{tab:list-of-skills-SDA-topics-pt1}, \ref{tab:list-of-skills-SDA-topics-pt2}, and \ref{tab:list-of-skills-SDA-topics-pt3}.}
\centering
\begin{longtable}{p{8.5cm}p{8.5cm}p{8.5cm}}
\toprule
\textbf{Skills} &  \\
\toprule
\texttt{stress\_management} & \texttt{theme\_identification} \\
\texttt{palette\_management} & \texttt{mathematical\_modeling} \\
\texttt{managing\_social\_interactions} & \texttt{outbreak\_response\_strategy} \\
\texttt{industry\_trends} & \texttt{languages\_and\_linguistics} \\
\texttt{differentiating\_functions} & \texttt{ethics\_compliance} \\
\texttt{oral\_presentation} & \texttt{curriculum\_development} \\
\texttt{marine\_geology} & \texttt{content\_distribution\_networking} \\
\texttt{cash\_flow\_analysis} & \texttt{genre\_identification} \\
\texttt{feedback\_assessment} & \texttt{graphing\_functions} \\
\texttt{ritual\_analysis} & \texttt{machine\_learning} \\
\texttt{cnc\_programming} & \texttt{ecological\_conservation} \\
\texttt{social\_media\_and\_digital\_marketing} & \texttt{troubleshooting\_electrical\_issues} \\
\texttt{machine\_learning\_implementation} & \texttt{trend\_forecasting} \\
\texttt{investment\_advisory} & \texttt{database\_management} \\
\texttt{painting\_techniques} & \texttt{economic\_policy\_analysis} \\
\texttt{developing\_training\_programs} & \texttt{simplifying\_expressions} \\
\texttt{personal\_care\_and\_hygiene} & \texttt{painting\_walls} \\
\texttt{pet\_care\_abilities} & \texttt{peer\_support} \\
\texttt{sports\_coaching} & \texttt{user\_experience\_evaluation} \\
\texttt{disease\_surveillance} & \texttt{creative\_writing} \\
\texttt{cultural\_etiquette\_learning} & \texttt{fact\_checking} \\
\texttt{sustainable\_construction} & \texttt{environmental\_policy} \\
\texttt{yarn\_spinning} & \texttt{clay\_modeling} \\
\texttt{resource\_allocation} & \texttt{health\_promotion} \\
\texttt{disease\_prevention} & \texttt{biodiversity\_monitoring} \\
\texttt{geographic\_information\_systems\_for\_resource\_mapping} & \texttt{} \\
\texttt{sociolinguistic\_analysis} & \texttt{software\_proficiency} \\
\texttt{musical\_composition} & \texttt{emergency\_care} \\
\texttt{severe\_weather\_response} & \texttt{critical\_thinking\_facilitation} \\
\texttt{regulatory\_compliance} & \texttt{personal\_finance} \\
\texttt{it\_support} & \texttt{data\_analysis\_astronomy} \\
\texttt{calculating\_determinants} & \texttt{installation\_art\_construction} \\
\bottomrule
\end{longtable}
\end{sidewaystable}
\begin{sidewaystable}
\caption{(Part 15 of 18) 1143 skills extracted from topics in Tables~\ref{tab:list-of-skills-SDA-topics-pt1}, \ref{tab:list-of-skills-SDA-topics-pt2}, and \ref{tab:list-of-skills-SDA-topics-pt3}.}
\centering
\begin{longtable}{p{8.5cm}p{8.5cm}p{8.5cm}}
\toprule
\textbf{Skills} &  \\
\toprule
\texttt{plumbing\_basics} & \texttt{portfolio\_management} \\
\texttt{robotics\_and\_automation} & \texttt{source\_evaluation} \\
\texttt{monitor\_hydration\_levels} & \texttt{financial\_regulatory\_compliance} \\
\texttt{policy\_development\_and\_analysis} & \texttt{health\_communication} \\
\texttt{health\_education\_program\_development} & \texttt{intercultural\_communication} \\
\texttt{safety\_precautions} & \texttt{flooring\_installation} \\
\texttt{smart\_home\_technology\_integration} & \texttt{digital\_communication} \\
\texttt{fundraising\_management} & \texttt{voice\_control} \\
\texttt{inventory\_management} & \texttt{instrumentation\_and\_measurement} \\
\texttt{audience\_engagement\_strategies} & \texttt{poison\_management} \\
\texttt{therapy\_application} & \texttt{gardening\_basics} \\
\texttt{public\_health\_initiatives} & \texttt{career\_and\_professional\_development} \\
\texttt{robotics\_engineering} & \texttt{global\_mobility\_management} \\
\texttt{metal\_welding} & \texttt{international\_law} \\
\texttt{flavor\_pairing} & \texttt{music\_theory\_analysis} \\
\texttt{financial\_management} & \texttt{digital\_prototyping} \\
\texttt{literary\_theory\_application} & \texttt{veterinary\_care\_coordination} \\
\texttt{keyword\_research} & \texttt{video\_production} \\
\texttt{computational\_linguistics\_development} & \texttt{efficiency\_optimization} \\
\texttt{stress\_management\_techniques} & \texttt{transport\_technology} \\
\texttt{emergency\_response} & \texttt{database\_integration} \\
\texttt{health\_education} & \texttt{travel\_planning} \\
\texttt{singing\_ability} & \texttt{emergency\_preparedness} \\
\texttt{budget\_management} & \texttt{risk\_management} \\
\texttt{recursive\_thinking} & \texttt{user\_interface\_design} \\
\texttt{patient\_care\_management} & \texttt{ethical\_content\_practices} \\
\texttt{instrument\_playing} & \texttt{digital\_design\_3d\_modeling} \\
\texttt{demand\_forecasting} & \texttt{chromatographic\_methods} \\
\texttt{maritime\_navigation\_systems} & \texttt{logistics\_planning} \\
\texttt{medication\_administration} & \texttt{innovation\_management} \\
\texttt{vehicle\_design\_analysis} & \texttt{credit\_score\_improvement} \\
\texttt{exercise\_routine\_planning} & \texttt{comparative\_analysis} \\
\bottomrule
\end{longtable}
\end{sidewaystable}
\begin{sidewaystable}
\caption{(Part 16 of 18) 1143 skills extracted from topics in Tables~\ref{tab:list-of-skills-SDA-topics-pt1}, \ref{tab:list-of-skills-SDA-topics-pt2}, and \ref{tab:list-of-skills-SDA-topics-pt3}.}
\centering
\begin{longtable}{p{8.5cm}p{8.5cm}p{8.5cm}}
\toprule
\textbf{Skills} &  \\
\toprule
\texttt{earthquake\_analysis} & \texttt{ethical\_decision\_making} \\
\texttt{cultural\_dynamics} & \texttt{material\_strength\_testing} \\
\texttt{vocabulary\_expansion} & \texttt{household\_organization} \\
\texttt{integrating\_functions} & \texttt{reflective\_judgment} \\
\texttt{blockchain\_development} & \texttt{symbolic\_interpretation} \\
\texttt{protocol\_management} & \texttt{clinical\_diagnosis} \\
\texttt{brush\_technique} & \texttt{theoretical\_application} \\
\texttt{philosophical\_writing} & \texttt{literary\_studies} \\
\texttt{cultural\_competency} & \texttt{marketing\_promotion} \\
\texttt{transportation\_planning} & \texttt{hospitality\_and\_tourism} \\
\texttt{training\_routines} & \texttt{public\_education\_on\_animal\_health} \\
\texttt{tour\_management} & \texttt{housekeeping\_management} \\
\texttt{legal\_advising} & \texttt{hotel\_management} \\
\texttt{sustainable\_agriculture} & \texttt{stakeholder\_analysis} \\
\texttt{quality\_control} & \texttt{competitive\_analysis} \\
\texttt{portfolio\_optimization} & \texttt{lifestyle\_choices} \\
\texttt{home\_and\_family} & \texttt{property\_valuation} \\
\texttt{comparative\_study} & \texttt{memory\_reinforcement} \\
\texttt{debugging\_skills} & \texttt{recovery\_support} \\
\texttt{fundraising\_strategy} & \texttt{physical\_fitness} \\
\texttt{installing\_electrical\_wiring} & \texttt{environmental\_assessment} \\
\texttt{weaving\_techniques} & \texttt{problem\_solving\_skills} \\
\texttt{cloud\_computing\_integration} & \texttt{first\_aid\_training} \\
\texttt{career\_transition\_advice} & \texttt{sleep\_quality\_improvement} \\
\texttt{remote\_sensing} & \texttt{off\_page\_optimization} \\
\texttt{textual\_analysis} & \texttt{analytical\_thinking} \\
\texttt{customer\_relationship\_management} & \texttt{communication\_skills} \\
\texttt{media\_and\_communication} & \texttt{conflict\_resolution} \\
\texttt{chemical\_waste\_management} & \texttt{script\_analysis} \\
\texttt{financial\_management\_for\_nonprofits} & \texttt{debt\_management} \\
\texttt{film\_production} & \texttt{calorie\_tracking} \\
\texttt{comparative\_literature\_study} & \texttt{risk\_assessment\_and\_management} \\
\bottomrule
\end{longtable}
\end{sidewaystable}
\begin{sidewaystable}
\caption{(Part 17 of 18) 1143 skills extracted from topics in Tables~\ref{tab:list-of-skills-SDA-topics-pt1}, \ref{tab:list-of-skills-SDA-topics-pt2}, and \ref{tab:list-of-skills-SDA-topics-pt3}.}
\centering
\begin{longtable}{p{8.5cm}p{8.5cm}p{8.5cm}}
\toprule
\textbf{Skills} &  \\
\toprule
\texttt{benchmarking\_performance} & \texttt{supplement\_advising} \\
\texttt{regression\_analysis} & \texttt{algorithm\_optimization} \\
\texttt{mental\_health} & \texttt{understanding\_vector\_spaces} \\
\texttt{geographical\_facts} & \texttt{energy\_storage\_solutions} \\
\texttt{active\_passive\_voice\_conversion} & \texttt{sustainable\_fashion\_practices} \\
\texttt{first\_aid} & \texttt{resource\_scheduling} \\
\texttt{skill\_development} & \texttt{qualitative\_data\_collection} \\
\texttt{media\_studies\_and\_journalism} & \texttt{technology\_adaptation} \\
\texttt{disaster\_recovery\_planning} & \texttt{data\_collection} \\
\texttt{packing\_efficiency} & \texttt{stage\_presence\_development} \\
\texttt{automated\_testing} & \texttt{client\_education} \\
\texttt{sculpture\_forming} & \texttt{educational\_support} \\
\texttt{sound\_engineering} & \texttt{art\_criticism} \\
\texttt{complexity\_analysis} & \texttt{weather\_prediction} \\
\texttt{location\_analysis} & \texttt{editing\_and\_proofreading} \\
\texttt{integrity\_cultivation} & \texttt{sports\_journalism} \\
\texttt{team\_communication} & \texttt{planetary\_science} \\
\texttt{setting\_description} & \texttt{cloud\_computing} \\
\texttt{curatorial\_practices} & \texttt{pattern\_recognition} \\
\texttt{api\_design} & \texttt{design\_research\_methods} \\
\texttt{interpret\_symptom\_severity} & \texttt{student\_assessment\_analysis} \\
\texttt{climate\_modeling} & \texttt{geothermal\_systems\_design} \\
\texttt{textile\_design} & \texttt{conservation\_strategy\_planning} \\
\texttt{instrument\_proficiency} & \texttt{energy\_efficiency\_techniques} \\
\texttt{geographic\_information\_systems} & \texttt{technical\_analysis} \\
\texttt{observing\_etiquette\_rules} & \texttt{interior\_designing} \\
\texttt{stratigraphic\_correlation} & \texttt{grammar\_error\_identification} \\
\texttt{waste\_reduction\_strategies} & \texttt{advocacy\_strategy\_development} \\
\texttt{pedagogical\_content\_knowledge} & \texttt{debt\_financing} \\
\texttt{pet\_safety\_precautions} & \texttt{health\_education\_development} \\
\texttt{scientific\_writing} & \texttt{philosophy\_and\_religion} \\
\texttt{speaking\_fluency} & \texttt{quantitative\_decision\_making} \\
\bottomrule
\end{longtable}
\end{sidewaystable}
\begin{sidewaystable}
\caption{(Part 18 of 18) 1143 skills extracted from topics in Tables~\ref{tab:list-of-skills-SDA-topics-pt1}, \ref{tab:list-of-skills-SDA-topics-pt2}, and \ref{tab:list-of-skills-SDA-topics-pt3}.}
\centering
\begin{longtable}{p{8.5cm}p{8.5cm}p{8.5cm}}
\toprule
\textbf{Skills} &  \\
\toprule
\texttt{crisis\_communication} & \texttt{team\_building} \\
\texttt{application\_security} & \texttt{solar\_panel\_installation} \\
\texttt{traffic\_management\_systems} & \texttt{market\_analysis} \\
\texttt{electrical\_safety\_inspection} & \texttt{species\_identification} \\
\texttt{emotional\_intelligence} & \texttt{energy\_and\_resources} \\
\texttt{technical\_problem\_solving} & \texttt{fashion\_trends} \\
\texttt{poetic\_analysis} & \texttt{stakeholder\_negotiation} \\
\texttt{patina\_application} & \texttt{link\_building} \\
\texttt{food\_presentation} & \texttt{healthy\_eating\_advice} \\
\texttt{solving\_linear\_equations} & \texttt{history\_and\_geography} \\
\texttt{cultural\_adaptation\_strategies} & \texttt{trip\_planning} \\
\texttt{agile\_methodologies} & \texttt{negotiation\_tactics} \\
\texttt{digital\_art\_design} & \texttt{social\_media\_marketing} \\
\texttt{mindfulness\_techniques} & \texttt{planetary\_geology} \\
\texttt{security\_risk\_assessment} & \texttt{code\_optimization} \\
\texttt{technology\_integration} & \texttt{fashion\_branding} \\
\texttt{career\_planning} & \texttt{literary\_criticism} \\
\texttt{data\_driven\_strategy} & \texttt{public\_transport\_planning} \\
\texttt{dialogue\_crafting} & \texttt{home\_budget\_management} \\
\texttt{digital\_music\_production} & \texttt{risk\_factor\_identification} \\
\texttt{reservation\_handling} & \texttt{vaccination\_schedule\_management} \\
\texttt{creative\_problem\_solving} & \texttt{social\_media\_ads\_creation} \\
\texttt{user\_experience\_evaluation} & \texttt{sleep\_improvement\_strategies} \\
\texttt{coding\_and\_debugging} & \texttt{client\_communication} \\
\texttt{legislative\_drafting} & \texttt{color\_theory\_application} \\
\texttt{sustainability\_planning} & \texttt{human\_resources\_management} \\
\texttt{sustainable\_agriculture\_practices} & \texttt{literary\_critique} \\
\texttt{active\_listening} & \texttt{syntax\_analysis} \\
\texttt{theme\_exploration} & \texttt{choreography\_design} \\
\texttt{epidemic\_outbreak\_investigation} & \texttt{autonomous\_vehicle\_navigation} \\
\bottomrule
\end{longtable}
\end{sidewaystable}

\clearpage

\begin{sidewaystable}
\caption{(Part 1 of 5) 18 Query/Task Types extracted from interactions with \gptTurbo.}
\label{tab:list-of-skills-SDA-query-types}   
\centering
\begin{longtable}{p{8.5cm}p{8.5cm}p{8.5cm}}

\toprule
\textbf{Query Type} &  Description \\
\toprule

\texttt{Information-Seeking}
&
\texttt{This includes any query where the user is looking to find out facts, data, explanations, or learn about a topic.}
\\ \\
\texttt{Help-Seeking} 
&
\texttt{Queries where the user needs assistance in solving a problem or performing an action. This could be technical support, troubleshooting, or guidance on personal issues.}
\\ \\
\texttt{Instructional} & 
\texttt{ Queries that specifically request detailed, step-by-step instructions or procedural guidance. This category is designed to assist users in understanding and executing tasks by breaking them down into sequential, manageable steps. Whether it's a practical day-to-day task, a complex technical procedure, or a creative process, the AI provides a clear, methodical approach to accomplishing specific objectives. }
\\ \\
\texttt{Conversational} & 
\texttt{These are queries where the user is possibly looking for engagement more than specific information or tasks. This can include small talk or generative interactions aimed at entertainment or companionship.}
\\
\bottomrule
\end{longtable}
\end{sidewaystable}
\begin{sidewaystable}
\caption{(Part 2 of 5) 18 Query/Task Types extracted from interactions with \gptTurbo.}
\label{label:new-pipeline-query-types-pt-2}   
\centering
\begin{longtable}{p{8.5cm}p{8.5cm}p{8.5cm}}

\toprule
\textbf{Query Type} &  Description \\
\toprule

\texttt{Narrative} & 
\texttt{Queries where the user is interested in hearing stories, experiences, or detailed accounts of events. These can be historical, personal, or fictional.}
\\ \\
\texttt{Planning} & 
\texttt{Queries that assist in planning or organizing activities, events, or projects. }
\\ \\
\texttt{Situational} & 
\texttt{Queries related to specific situations or scenarios that the user is facing, asking for tailored advice or solutions.} 
\\ \\
\texttt{Interpretative} & 
\texttt{Queries that ask for interpretation of texts, artworks, or other cultural artifacts.}
\\ \\
\texttt{Decision-Making} & 
\texttt{Queries that assist the user in making decisions by evaluating options, risks, and benefits.}
\\ \\
\texttt{Task Execution} & 
\texttt{Queries where the user delegates the completion of a specific task or action to the AI. This involves the AI taking on roles that might require decision-making, processing, or interacting with other systems to achieve the desired outcome.}
\\
\bottomrule
\end{longtable}
\end{sidewaystable}
\begin{sidewaystable}
\caption{(Part 3 of 5) 18 Query/Task Types extracted from interactions with \gptTurbo.}
\label{label:new-pipeline-query-types-pt-3}   
\centering
\begin{longtable}{p{8.5cm}p{8.5cm}p{8.5cm}}

\toprule
\textbf{Query Type} &  Description \\
\toprule
\texttt{Digital Communication Design and Strategy} & 
\texttt{Queries focused on designing and strategizing content for optimal communication across digital platforms. This includes creating engaging designs and coherent strategies for websites, blogs, podcasts, emails, and digital essays.}
\\ \\
\texttt{Code Generation} & 
\texttt{Queries where users directly ask for guidance on implementing specific functions, features, or algorithms in a particular programming language. These queries explicitly request code snippets, examples, or step-by-step instructions on how to implement the desired functionality. }
\\ \\
\texttt{Fact-Seeking} & 
\texttt{Queries where the user is looking for specific, factual information or data points. These queries are often straightforward and can be answered with a concise response. The focus is on providing accurate, reliable information from trustworthy sources. } 
\\ \\
\texttt{Comparative} & 
\texttt{These queries involve comparing different items, ideas, or scenarios. Users often seek assistance in making decisions or understanding differences. }
\\
\bottomrule
\end{longtable}
\end{sidewaystable}

\begin{sidewaystable}
\caption{(Part 4 of 5) 18 Query/Task Types extracted from interactions with \gptTurbo.}
\label{label:new-pipeline-query-types-pt-4}   
\centering
\begin{longtable}{p{8.5cm}p{8.5cm}p{8.5cm}}

\toprule
\textbf{Query Type} &  Description \\
\toprule
\\ \\
\texttt{Interpretive Explanation} & 
\texttt{Queries in which the user seeks a detailed explanation or interpretation of a provided text snippet. This could include literary analysis, code explanation, or any form of textual dissection. The AI acts as an analytical tool to help users understand complex passages, technical descriptions, or conceptual writings.}
\\ \\
\texttt{Error Detection} & 
\texttt{Queries where the user seeks assistance in finding and diagnosing errors or bugs in provided materials. This could include syntactical errors in code, grammatical mistakes in written text, or inconsistencies in data sets. The AI acts as a diagnostic tool to help pinpoint and suggest corrections for these issues.}
\\ \\
\texttt{Feedback-Seeking} & 
\texttt{Queries where users are looking for feedback on their ideas, creations, or opinions. This can be particularly relevant in educational, artistic, or professional contexts.}
\\
\bottomrule
\end{longtable}
\end{sidewaystable}

\begin{sidewaystable}
\caption{(Part 5 of 5) 18 Query/Task Types extracted from interactions with \gptTurbo.}
\label{label:new-pipeline-query-types-pt-5}   
\centering
\begin{longtable}{p{8.5cm}p{8.5cm}p{8.5cm}}

\toprule
\textbf{Query Type} &  Description \\
\toprule
\\ \\
\texttt{Clarification-Seeking} & 
\texttt{These queries aim to clarify confusion or get more detailed information about a previously mentioned or understood topic. Users might need further explanation or a more refined understanding of a complex issue. }
\\
\bottomrule
\end{longtable}
\end{sidewaystable}
\clearpage
\end{small}

\subsection{\instructSMSDA{} List of Skills and Query Types (Claude-3.5-Sonnet)}
\label{appendix:list-of-skills-SDA-claude}
Using the procedure detailed in Section~\ref{subsec:skill-extraction}, we extract 138 conversational topics and 13 query types from Claude-3.5-Sonnet. From the topics, we get a fine-grained list of 3,266 skills. 

\begin{small}
\begin{sidewaystable}
\caption{(Part 1 of 3) 138 topics extracted from interactions with Claude-3.5-Sonnet}
\label{tab:list-of-skills-SDA-claude-topics-pt1}
\centering
\begin{longtable}{p{8.5cm}p{8.5cm}p{8.5cm}}
\toprule
\textbf{Topics} &  \\
\toprule
\texttt{computer\_science\_and\_technology} & \texttt{software\_development\_practices} \\
\texttt{data\_analysis\_methods} & \texttt{network\_security\_systems} \\
\texttt{computer\_infrastructure\_management} & \texttt{emerging\_technology\_trends} \\
\texttt{health\_and\_medicine} & \texttt{medical\_treatment\_methods} \\
\texttt{public\_health\_management} & \texttt{mental\_health\_treatment} \\
\texttt{fitness\_training\_methods} & \texttt{emergency\_medical\_response} \\
\texttt{business\_and\_economics} & \texttt{financial\_management\_systems} \\
\bottomrule
\end{longtable}
\end{sidewaystable}
\begin{sidewaystable}
\caption{(Part 2 of 3) 138 topics extracted from interactions with Claude-3.5-Sonnet}
\label{tab:list-of-skills-SDA-claude-topics-pt2}
\centering
\begin{longtable}{p{8.5cm}p{8.5cm}p{8.5cm}}
\toprule
\textbf{Topics} &  \\
\toprule
\texttt{journalism\_content\_production} & \texttt{nonprofit\_organization\_management} \\
\texttt{industry\_and\_manufacturing} & \texttt{manufacturing\_process\_optimization} \\
\texttt{quality\_control\_systems} & \texttt{supply\_chain\_optimization} \\
\texttt{industrial\_product\_design} & \texttt{industrial\_automation\_systems} \\
\texttt{sports\_and\_recreation} & \texttt{team\_sports\_techniques} \\
\texttt{individual\_sports\_training} & \texttt{athletic\_coaching\_methods} \\
\texttt{sports\_science\_analysis} & \texttt{recreational\_activity\_planning} \\
\bottomrule
\end{longtable}
\end{sidewaystable}
\begin{sidewaystable}
\caption{(Part 3 of 3) 138 topics extracted from interactions with Claude-3.5-Sonnet}
\label{tab:list-of-skills-SDA-claude-topics-pt3}
\centering
\begin{longtable}{p{8.5cm}p{8.5cm}p{8.5cm}}
\toprule
\textbf{Topics} &  \\
\toprule
\texttt{project\_management\_techniques} & \texttt{negotiation\_strategy\_development} \\
\texttt{time\_management\_techniques} & \texttt{business\_communication\_methods} \\
\texttt{digital\_technologies} & \texttt{artificial\_intelligence\_development} \\
\texttt{blockchain\_system\_development} & \texttt{iot\_device\_management} \\
\texttt{extended\_reality\_development} & \texttt{cloud\_computing\_systems} \\
\bottomrule
\end{longtable}
\end{sidewaystable}

\clearpage

\begin{sidewaystable}
\caption{(Part 1 of 52) 3266 skills extracted from topics in Tables~\ref{tab:list-of-skills-SDA-claude-topics-pt1}, \ref{tab:list-of-skills-SDA-claude-topics-pt2}, and \ref{tab:list-of-skills-SDA-claude-topics-pt3}.}
\label{tab:list-of-skills-SDA-skills-claude}
\centering
\begin{longtable}{p{8.5cm}p{8.5cm}p{8.5cm}}
\toprule
\textbf{Skills} &  \\
\toprule
\texttt{science\_and\_engineering} & \texttt{scientific\_method\_application} \\
\texttt{data\_analysis\_and\_interpretation} & \texttt{experimental\_design} \\
\texttt{hypothesis\_formulation} & \texttt{technical\_writing} \\
\texttt{laboratory\_techniques} & \texttt{problem\_solving} \\
\texttt{critical\_thinking} & \texttt{mathematical\_modeling} \\
\texttt{statistical\_analysis} & \texttt{research\_methodology} \\
\texttt{project\_management} & \texttt{computer\_aided\_design} \\
\bottomrule
\end{longtable}
\end{sidewaystable}
\begin{sidewaystable}
\caption{(Part 2 of 52) 3266 skills extracted from topics in Tables~\ref{tab:list-of-skills-SDA-claude-topics-pt1}, \ref{tab:list-of-skills-SDA-claude-topics-pt2}, and \ref{tab:list-of-skills-SDA-claude-topics-pt3}.}
\centering
\begin{longtable}{p{8.5cm}p{8.5cm}p{8.5cm}}
\toprule
\textbf{Skills} &  \\
\toprule
\texttt{cybersecurity\_implementation} & \texttt{web\_development} \\
\texttt{mobile\_app\_development} & \texttt{cloud\_computing} \\
\texttt{artificial\_intelligence} & \texttt{machine\_learning} \\
\texttt{data\_analysis} & \texttt{system\_architecture} \\
\texttt{version\_control} & \texttt{debugging\_and\_troubleshooting} \\
\texttt{operating\_system\_management} & \texttt{hardware\_configuration} \\
\texttt{user\_interface\_design} & \texttt{agile\_methodologies} \\
\bottomrule
\end{longtable}
\end{sidewaystable}
\begin{sidewaystable}
\caption{(Part 3 of 52) 3266 skills extracted from topics in Tables~\ref{tab:list-of-skills-SDA-claude-topics-pt1}, \ref{tab:list-of-skills-SDA-claude-topics-pt2}, and \ref{tab:list-of-skills-SDA-claude-topics-pt3}.}
\centering
\begin{longtable}{p{8.5cm}p{8.5cm}p{8.5cm}}
\toprule
\textbf{Skills} &  \\
\toprule
\texttt{plan\_for\_emergencies} & \texttt{coordinate\_group\_travel} \\
\texttt{research\_visa\_requirements} & \texttt{compare\_currency\_exchange\_options} \\
\texttt{create\_backup\_plans} & \texttt{research\_local\_cuisine\_options} \\
\texttt{plan\_for\_health\_and\_safety\_precautions} & \texttt{professional\_services} \\
\texttt{client\_relationship\_management} & \texttt{project\_planning\_and\_execution} \\
\texttt{business\_analysis} & \texttt{strategic\_consulting} \\
\texttt{financial\_modeling} & \texttt{risk\_assessment\_and\_mitigation} \\
\bottomrule
\end{longtable}
\end{sidewaystable}
\begin{sidewaystable}
\caption{(Part 4 of 52) 3266 skills extracted from topics in Tables~\ref{tab:list-of-skills-SDA-claude-topics-pt1}, \ref{tab:list-of-skills-SDA-claude-topics-pt2}, and \ref{tab:list-of-skills-SDA-claude-topics-pt3}.}
\centering
\begin{longtable}{p{8.5cm}p{8.5cm}p{8.5cm}}
\toprule
\textbf{Skills} &  \\
\toprule
\texttt{prescribe\_psychiatric\_medications} & \texttt{monitor\_patient\_progress} \\
\texttt{provide\_crisis\_intervention} & \texttt{implement\_cognitive\_behavioral\_therapy} \\
\texttt{practice\_active\_listening} & \texttt{apply\_mindfulness\_techniques} \\
\texttt{facilitate\_group\_therapy\_sessions} & \texttt{perform\_psychological\_assessments} \\
\texttt{utilize\_motivational\_interviewing} & \texttt{manage\_patient\_records} \\
\texttt{collaborate\_with\_healthcare\_professionals} & \texttt{} \\
\texttt{educate\_patients\_and\_families} & \texttt{apply\_trauma\_informed\_care} \\
\bottomrule
\end{longtable}
\end{sidewaystable}
\begin{sidewaystable}
\caption{(Part 5 of 52) 3266 skills extracted from topics in Tables~\ref{tab:list-of-skills-SDA-claude-topics-pt1}, \ref{tab:list-of-skills-SDA-claude-topics-pt2}, and \ref{tab:list-of-skills-SDA-claude-topics-pt3}.}
\centering
\begin{longtable}{p{8.5cm}p{8.5cm}p{8.5cm}}
\toprule
\textbf{Skills} &  \\
\toprule
\texttt{evaluate\_cultural\_influences\_on\_behavior} & \texttt{film\_production\_techniques} \\
\texttt{camera\_operation} & \texttt{lighting\_setup} \\
\texttt{sound\_recording} & \texttt{script\_writing} \\
\texttt{storyboard\_creation} & \texttt{set\_design} \\
\texttt{costume\_design} & \texttt{makeup\_application} \\
\texttt{special\_effects\_implementation} & \texttt{video\_editing} \\
\texttt{color\_grading} & \texttt{sound\_mixing} \\
\bottomrule
\end{longtable}
\end{sidewaystable}
\begin{sidewaystable}
\caption{(Part 6 of 52) 3266 skills extracted from topics in Tables~\ref{tab:list-of-skills-SDA-claude-topics-pt1}, \ref{tab:list-of-skills-SDA-claude-topics-pt2}, and \ref{tab:list-of-skills-SDA-claude-topics-pt3}.}
\centering
\begin{longtable}{p{8.5cm}p{8.5cm}p{8.5cm}}
\toprule
\textbf{Skills} &  \\
\toprule
\texttt{operate\_communication\_equipment} & \texttt{manage\_fuel\_consumption} \\
\texttt{perform\_risk\_assessments} & \texttt{coordinate\_port\_operations} \\
\texttt{execute\_docking\_procedures} & \texttt{monitor\_vessel\_performance} \\
\texttt{implement\_environmental\_protection\_measures} & \texttt{} \\
\texttt{manage\_maritime\_documentation} & \texttt{operate\_navigation\_instruments} \\
\texttt{conduct\_maritime\_security\_checks} & \texttt{agricultural\_technology\_systems} \\
\texttt{operate\_precision\_farming\_equipment} & \texttt{analyze\_soil\_composition\_data} \\
\bottomrule
\end{longtable}
\end{sidewaystable}
\begin{sidewaystable}
\caption{(Part 7 of 52) 3266 skills extracted from topics in Tables~\ref{tab:list-of-skills-SDA-claude-topics-pt1}, \ref{tab:list-of-skills-SDA-claude-topics-pt2}, and \ref{tab:list-of-skills-SDA-claude-topics-pt3}.}
\centering
\begin{longtable}{p{8.5cm}p{8.5cm}p{8.5cm}}
\toprule
\textbf{Skills} &  \\
\toprule
\texttt{execute\_cardiovascular\_routines} & \texttt{demonstrate\_proper\_form} \\
\texttt{adjust\_training\_intensity} & \texttt{monitor\_heart\_rate} \\
\texttt{track\_fitness\_progress} & \texttt{implement\_periodization\_techniques} \\
\texttt{teach\_flexibility\_exercises} & \texttt{apply\_plyometric\_training} \\
\texttt{conduct\_fitness\_assessments} & \texttt{utilize\_resistance\_bands} \\
\texttt{operate\_gym\_equipment} & \texttt{practice\_interval\_training} \\
\texttt{modify\_exercises\_for\_injuries} & \texttt{incorporate\_bodyweight\_exercises} \\
\bottomrule
\end{longtable}
\end{sidewaystable}
\begin{sidewaystable}
\caption{(Part 8 of 52) 3266 skills extracted from topics in Tables~\ref{tab:list-of-skills-SDA-claude-topics-pt1}, \ref{tab:list-of-skills-SDA-claude-topics-pt2}, and \ref{tab:list-of-skills-SDA-claude-topics-pt3}.}
\centering
\begin{longtable}{p{8.5cm}p{8.5cm}p{8.5cm}}
\toprule
\textbf{Skills} &  \\
\toprule
\texttt{psychological\_warfare\_techniques} & \texttt{military\_budget\_allocation} \\
\texttt{defense\_policy\_formulation} & \texttt{military\_recruitment\_and\_training} \\
\texttt{military\_technology\_development} & \texttt{military\_history\_analysis} \\
\texttt{financial\_management\_systems} & \texttt{budget\_planning\_and\_forecasting} \\
\texttt{financial\_data\_analysis} & \texttt{risk\_assessment\_and\_management} \\
\texttt{cash\_flow\_monitoring} & \texttt{financial\_reporting\_and\_compliance} \\
\texttt{investment\_portfolio\_management} & \texttt{cost\_control\_and\_optimization} \\
\bottomrule
\end{longtable}
\end{sidewaystable}
\begin{sidewaystable}
\caption{(Part 9 of 52) 3266 skills extracted from topics in Tables~\ref{tab:list-of-skills-SDA-claude-topics-pt1}, \ref{tab:list-of-skills-SDA-claude-topics-pt2}, and \ref{tab:list-of-skills-SDA-claude-topics-pt3}.}
\centering
\begin{longtable}{p{8.5cm}p{8.5cm}p{8.5cm}}
\toprule
\textbf{Skills} &  \\
\toprule
\texttt{evidence\_based\_reasoning} & \texttt{historical\_narrative\_creation} \\
\texttt{interdisciplinary\_connection\_making} & \texttt{long\_term\_impact\_assessment} \\
\texttt{historical\_trend\_identification} & \texttt{critical\_thinking\_application} \\
\texttt{historical\_methodology\_implementation} & \texttt{historical\_perspective\_taking} \\
\texttt{medical\_treatment\_methods} & \texttt{diagnose\_medical\_conditions} \\
\texttt{develop\_treatment\_plans} & \texttt{administer\_medications} \\
\texttt{perform\_surgical\_procedures} & \texttt{interpret\_medical\_test\_results} \\
\bottomrule
\end{longtable}
\end{sidewaystable}
\begin{sidewaystable}
\caption{(Part 10 of 52) 3266 skills extracted from topics in Tables~\ref{tab:list-of-skills-SDA-claude-topics-pt1}, \ref{tab:list-of-skills-SDA-claude-topics-pt2}, and \ref{tab:list-of-skills-SDA-claude-topics-pt3}.}
\centering
\begin{longtable}{p{8.5cm}p{8.5cm}p{8.5cm}}
\toprule
\textbf{Skills} &  \\
\toprule
\texttt{anatomy\_understanding} & \texttt{mixed\_media\_integration} \\
\texttt{art\_history\_interpretation} & \texttt{visual\_storytelling} \\
\texttt{abstract\_concept\_visualization} & \texttt{proportion\_and\_scale\_management} \\
\texttt{negative\_space\_utilization} & \texttt{color\_palette\_selection} \\
\texttt{gesture\_drawing} & \texttt{digital\_software\_navigation} \\
\texttt{art\_material\_knowledge} & \texttt{stylization\_and\_abstraction} \\
\texttt{visual\_balance\_creation} & \texttt{symbolism\_incorporation} \\
\bottomrule
\end{longtable}
\end{sidewaystable}
\begin{sidewaystable}
\caption{(Part 11 of 52) 3266 skills extracted from topics in Tables~\ref{tab:list-of-skills-SDA-claude-topics-pt1}, \ref{tab:list-of-skills-SDA-claude-topics-pt2}, and \ref{tab:list-of-skills-SDA-claude-topics-pt3}.}
\centering
\begin{longtable}{p{8.5cm}p{8.5cm}p{8.5cm}}
\toprule
\textbf{Skills} &  \\
\toprule
\texttt{transit\_technology\_evaluation} & \texttt{public\_outreach\_coordination} \\
\texttt{transit\_policy\_development} & \texttt{recreational\_activity\_planning} \\
\texttt{activity\_research\_and\_discovery} & \texttt{budget\_allocation\_and\_management} \\
\texttt{time\_management\_and\_scheduling} & \texttt{risk\_assessment\_and\_mitigation} \\
\texttt{group\_dynamics\_management} & \texttt{equipment\_selection\_and\_preparation} \\
\texttt{location\_scouting\_and\_evaluation} & \texttt{weather\_forecasting\_interpretation} \\
\texttt{transportation\_logistics\_coordination} & \texttt{emergency\_response\_planning} \\
\bottomrule
\end{longtable}
\end{sidewaystable}
\begin{sidewaystable}
\caption{(Part 12 of 52) 3266 skills extracted from topics in Tables~\ref{tab:list-of-skills-SDA-claude-topics-pt1}, \ref{tab:list-of-skills-SDA-claude-topics-pt2}, and \ref{tab:list-of-skills-SDA-claude-topics-pt3}.}
\centering
\begin{longtable}{p{8.5cm}p{8.5cm}p{8.5cm}}
\toprule
\textbf{Skills} &  \\
\toprule
\texttt{environmental\_health\_monitoring} & \texttt{health\_emergency\_preparedness} \\
\texttt{health\_data\_visualization} & \texttt{health\_impact\_evaluation} \\
\texttt{cross\_sector\_collaboration\_facilitation} & \texttt{health\_legislation\_interpretation} \\
\texttt{population\_health\_trend\_analysis} & \texttt{health\_technology\_assessment} \\
\texttt{public\_health\_ethics\_application} & \texttt{global\_health\_diplomacy} \\
\texttt{marketing\_campaign\_management} & \texttt{campaign\_strategy\_development} \\
\texttt{target\_audience\_identification} & \texttt{market\_research\_analysis} \\
\bottomrule
\end{longtable}
\end{sidewaystable}
\begin{sidewaystable}
\caption{(Part 13 of 52) 3266 skills extracted from topics in Tables~\ref{tab:list-of-skills-SDA-claude-topics-pt1}, \ref{tab:list-of-skills-SDA-claude-topics-pt2}, and \ref{tab:list-of-skills-SDA-claude-topics-pt3}.}
\centering
\begin{longtable}{p{8.5cm}p{8.5cm}p{8.5cm}}
\toprule
\textbf{Skills} &  \\
\toprule
\texttt{program\_development\_and\_evaluation} & \texttt{donor\_relations\_and\_stewardship} \\
\texttt{nonprofit\_marketing\_and\_communications} & \texttt{legal\_compliance\_and\_risk\_management} \\
\texttt{community\_outreach\_and\_engagement} & \texttt{nonprofit\_leadership\_and\_decision\_making} \\
\texttt{performance\_measurement\_and\_impact\_assessment} & \texttt{} \\
\texttt{resource\_allocation\_and\_budgeting} & \texttt{stakeholder\_relationship\_management} \\
\texttt{nonprofit\_technology\_implementation} & \texttt{ethical\_decision\_making\_in\_nonprofits} \\
\texttt{conflict\_resolution\_and\_negotiation} & \texttt{nonprofit\_human\_resources\_management} \\
\bottomrule
\end{longtable}
\end{sidewaystable}
\begin{sidewaystable}
\caption{(Part 14 of 52) 3266 skills extracted from topics in Tables~\ref{tab:list-of-skills-SDA-claude-topics-pt1}, \ref{tab:list-of-skills-SDA-claude-topics-pt2}, and \ref{tab:list-of-skills-SDA-claude-topics-pt3}.}
\centering
\begin{longtable}{p{8.5cm}p{8.5cm}p{8.5cm}}
\toprule
\textbf{Skills} &  \\
\toprule
\texttt{economic\_market\_analysis} & \texttt{data\_collection\_and\_management} \\
\texttt{statistical\_analysis} & \texttt{econometric\_modeling} \\
\texttt{trend\_identification} & \texttt{market\_segmentation} \\
\texttt{competitive\_landscape\_assessment} & \texttt{demand\_forecasting} \\
\texttt{supply\_chain\_analysis} & \texttt{pricing\_strategy\_development} \\
\texttt{risk\_assessment\_and\_mitigation} & \texttt{financial\_ratio\_analysis} \\
\texttt{macroeconomic\_indicator\_interpretation} & \texttt{industry\_specific\_knowledge\_application} \\
\bottomrule
\end{longtable}
\end{sidewaystable}
\begin{sidewaystable}
\caption{(Part 15 of 52) 3266 skills extracted from topics in Tables~\ref{tab:list-of-skills-SDA-claude-topics-pt1}, \ref{tab:list-of-skills-SDA-claude-topics-pt2}, and \ref{tab:list-of-skills-SDA-claude-topics-pt3}.}
\centering
\begin{longtable}{p{8.5cm}p{8.5cm}p{8.5cm}}
\toprule
\textbf{Skills} &  \\
\toprule
\texttt{midi\_programming} & \texttt{sample\_creation\_and\_manipulation} \\
\texttt{acoustic\_treatment\_implementation} & \texttt{music\_business\_management} \\
\texttt{collaborative\_project\_coordination} & \texttt{music\_copyright\_understanding} \\
\texttt{studio\_equipment\_maintenance} & \texttt{audio\_signal\_processing} \\
\texttt{music\_marketing\_and\_promotion} & \texttt{livestock\_management\_systems} \\
\texttt{animal\_health\_monitoring} & \texttt{feed\_management\_optimization} \\
\texttt{breeding\_program\_implementation} & \texttt{herd\_productivity\_analysis} \\
\bottomrule
\end{longtable}
\end{sidewaystable}
\begin{sidewaystable}
\caption{(Part 16 of 52) 3266 skills extracted from topics in Tables~\ref{tab:list-of-skills-SDA-claude-topics-pt1}, \ref{tab:list-of-skills-SDA-claude-topics-pt2}, and \ref{tab:list-of-skills-SDA-claude-topics-pt3}.}
\centering
\begin{longtable}{p{8.5cm}p{8.5cm}p{8.5cm}}
\toprule
\textbf{Skills} &  \\
\toprule
\texttt{statistical\_software\_usage} & \texttt{experimental\_design} \\
\texttt{multivariate\_analysis} & \texttt{time\_series\_analysis} \\
\texttt{factor\_analysis} & \texttt{cluster\_analysis} \\
\texttt{power\_analysis} & \texttt{nonparametric\_test\_application} \\
\texttt{bayesian\_inference} & \texttt{meta\_analysis} \\
\texttt{principal\_component\_analysis} & \texttt{survival\_analysis} \\
\texttt{statistical\_report\_writing} & \texttt{data\_interpretation} \\
\bottomrule
\end{longtable}
\end{sidewaystable}
\begin{sidewaystable}
\caption{(Part 17 of 52) 3266 skills extracted from topics in Tables~\ref{tab:list-of-skills-SDA-claude-topics-pt1}, \ref{tab:list-of-skills-SDA-claude-topics-pt2}, and \ref{tab:list-of-skills-SDA-claude-topics-pt3}.}
\centering
\begin{longtable}{p{8.5cm}p{8.5cm}p{8.5cm}}
\toprule
\textbf{Skills} &  \\
\toprule
\texttt{accessory\_design} & \texttt{fashion\_show\_production} \\
\texttt{building\_construction\_methods} & \texttt{structural\_design\_analysis} \\
\texttt{blueprint\_reading\_interpretation} & \texttt{material\_selection\_optimization} \\
\texttt{construction\_site\_management} & \texttt{safety\_protocol\_implementation} \\
\texttt{cost\_estimation\_budgeting} & \texttt{project\_scheduling\_coordination} \\
\texttt{foundation\_laying\_techniques} & \texttt{framing\_and\_roofing\_execution} \\
\texttt{plumbing\_system\_installation} & \texttt{electrical\_wiring\_implementation} \\
\bottomrule
\end{longtable}
\end{sidewaystable}
\begin{sidewaystable}
\caption{(Part 18 of 52) 3266 skills extracted from topics in Tables~\ref{tab:list-of-skills-SDA-claude-topics-pt1}, \ref{tab:list-of-skills-SDA-claude-topics-pt2}, and \ref{tab:list-of-skills-SDA-claude-topics-pt3}.}
\centering
\begin{longtable}{p{8.5cm}p{8.5cm}p{8.5cm}}
\toprule
\textbf{Skills} &  \\
\toprule
\texttt{scale\_iot\_infrastructure} & \texttt{integrate\_iot\_platforms} \\
\texttt{manage\_device\_inventory} & \texttt{implement\_over\_the\_air\_updates} \\
\texttt{configure\_device\_authentication} & \texttt{monitor\_network\_performance} \\
\texttt{manage\_device\_groups} & \texttt{implement\_data\_encryption} \\
\texttt{create\_device\_dashboards} & \texttt{automate\_device\_onboarding} \\
\texttt{manage\_device\_permissions} & \texttt{analyze\_device\_logs} \\
\texttt{implement\_edge\_computing} & \texttt{optimize\_data\_transmission} \\
\bottomrule
\end{longtable}
\end{sidewaystable}
\begin{sidewaystable}
\caption{(Part 19 of 52) 3266 skills extracted from topics in Tables~\ref{tab:list-of-skills-SDA-claude-topics-pt1}, \ref{tab:list-of-skills-SDA-claude-topics-pt2}, and \ref{tab:list-of-skills-SDA-claude-topics-pt3}.}
\centering
\begin{longtable}{p{8.5cm}p{8.5cm}p{8.5cm}}
\toprule
\textbf{Skills} &  \\
\toprule
\texttt{government\_accountability\_measurement} & \texttt{political\_risk\_analysis} \\
\texttt{regime\_type\_classification} & \texttt{government\_reform\_strategy\_development} \\
\texttt{international\_governance\_comparison} & \texttt{fashion\_accessory\_design} \\
\texttt{trend\_analysis} & \texttt{sketching\_techniques} \\
\texttt{color\_theory\_application} & \texttt{material\_selection} \\
\texttt{3d\_modeling\_for\_accessories} & \texttt{prototype\_creation} \\
\texttt{computer\_aided\_design} & \texttt{market\_research} \\
\bottomrule
\end{longtable}
\end{sidewaystable}
\begin{sidewaystable}
\caption{(Part 20 of 52) 3266 skills extracted from topics in Tables~\ref{tab:list-of-skills-SDA-claude-topics-pt1}, \ref{tab:list-of-skills-SDA-claude-topics-pt2}, and \ref{tab:list-of-skills-SDA-claude-topics-pt3}.}
\centering
\begin{longtable}{p{8.5cm}p{8.5cm}p{8.5cm}}
\toprule
\textbf{Skills} &  \\
\toprule
\texttt{building\_code\_compliance} & \texttt{construction\_document\_interpretation} \\
\texttt{value\_engineering} & \texttt{conflict\_resolution} \\
\texttt{procurement\_and\_logistics\_management} & \texttt{environmental\_impact\_assessment} \\
\texttt{site\_selection\_and\_analysis} & \texttt{construction\_technology\_implementation} \\
\texttt{team\_leadership\_and\_motivation} & \texttt{nutrition\_planning\_methods} \\
\texttt{assess\_nutritional\_needs} & \texttt{calculate\_caloric\_requirements} \\
\texttt{create\_balanced\_meal\_plans} & \texttt{analyze\_macro\_nutrient\_ratios} \\
\bottomrule
\end{longtable}
\end{sidewaystable}
\begin{sidewaystable}
\caption{(Part 21 of 52) 3266 skills extracted from topics in Tables~\ref{tab:list-of-skills-SDA-claude-topics-pt1}, \ref{tab:list-of-skills-SDA-claude-topics-pt2}, and \ref{tab:list-of-skills-SDA-claude-topics-pt3}.}
\centering
\begin{longtable}{p{8.5cm}p{8.5cm}p{8.5cm}}
\toprule
\textbf{Skills} &  \\
\toprule
\texttt{renewable\_energy\_integration} & \texttt{carbon\_footprint\_calculation} \\
\texttt{green\_supply\_chain\_management} & \texttt{sustainability\_reporting\_and\_disclosure} \\
\texttt{stakeholder\_engagement\_facilitation} & \texttt{circular\_economy\_model\_implementation} \\
\texttt{sustainable\_materials\_selection} & \texttt{water\_conservation\_planning} \\
\texttt{energy\_audit\_conducting} & \texttt{biodiversity\_impact\_mitigation} \\
\texttt{sustainable\_agriculture\_practices\_implementation} & \texttt{} \\
\texttt{corporate\_social\_responsibility\_program\_development} & \texttt{} \\
\bottomrule
\end{longtable}
\end{sidewaystable}
\begin{sidewaystable}
\caption{(Part 22 of 52) 3266 skills extracted from topics in Tables~\ref{tab:list-of-skills-SDA-claude-topics-pt1}, \ref{tab:list-of-skills-SDA-claude-topics-pt2}, and \ref{tab:list-of-skills-SDA-claude-topics-pt3}.}
\centering
\begin{longtable}{p{8.5cm}p{8.5cm}p{8.5cm}}
\toprule
\textbf{Skills} &  \\
\toprule
\texttt{devops\_practices\_implementation} & \texttt{code\_versioning\_and\_release\_management} \\
\texttt{code\_modularity\_and\_reusability\_improvement} & \texttt{} \\
\texttt{unit\_testing\_implementation} & \texttt{cross\_platform\_development} \\
\texttt{scalability\_planning\_and\_implementation} & \texttt{education\_and\_academia} \\
\texttt{critical\_thinking} & \texttt{research\_methodology} \\
\texttt{academic\_writing} & \texttt{data\_analysis} \\
\texttt{literature\_review} & \texttt{classroom\_management} \\
\bottomrule
\end{longtable}
\end{sidewaystable}
\begin{sidewaystable}
\caption{(Part 23 of 52) 3266 skills extracted from topics in Tables~\ref{tab:list-of-skills-SDA-claude-topics-pt1}, \ref{tab:list-of-skills-SDA-claude-topics-pt2}, and \ref{tab:list-of-skills-SDA-claude-topics-pt3}.}
\centering
\begin{longtable}{p{8.5cm}p{8.5cm}p{8.5cm}}
\toprule
\textbf{Skills} &  \\
\toprule
\texttt{data\_analysis\_for\_content\_performance} & \texttt{user\_experience\_design} \\
\texttt{content\_curation\_and\_aggregation} & \texttt{digital\_storytelling} \\
\texttt{interactive\_content\_development} & \texttt{mobile\_content\_optimization} \\
\texttt{content\_localization\_and\_translation} & \texttt{digital\_asset\_management} \\
\texttt{content\_monetization\_strategies} & \texttt{audience\_segmentation\_and\_targeting} \\
\texttt{multimedia\_content\_production} & \texttt{live\_streaming\_and\_webinar\_hosting} \\
\texttt{augmented\_reality\_content\_creation} & \texttt{virtual\_reality\_experience\_design} \\
\bottomrule
\end{longtable}
\end{sidewaystable}
\begin{sidewaystable}
\caption{(Part 24 of 52) 3266 skills extracted from topics in Tables~\ref{tab:list-of-skills-SDA-claude-topics-pt1}, \ref{tab:list-of-skills-SDA-claude-topics-pt2}, and \ref{tab:list-of-skills-SDA-claude-topics-pt3}.}
\centering
\begin{longtable}{p{8.5cm}p{8.5cm}p{8.5cm}}
\toprule
\textbf{Skills} &  \\
\toprule
\texttt{content\_creation} & \texttt{audience\_engagement} \\
\texttt{social\_listening} & \texttt{analytics\_interpretation} \\
\texttt{trend\_identification} & \texttt{crisis\_management} \\
\texttt{community\_building} & \texttt{hashtag\_strategy} \\
\texttt{influencer\_collaboration} & \texttt{paid\_advertising\_management} \\
\texttt{brand\_voice\_development} & \texttt{visual\_content\_design} \\
\texttt{scheduling\_and\_automation} & \texttt{cross\_platform\_integration} \\
\bottomrule
\end{longtable}
\end{sidewaystable}
\begin{sidewaystable}
\caption{(Part 25 of 52) 3266 skills extracted from topics in Tables~\ref{tab:list-of-skills-SDA-claude-topics-pt1}, \ref{tab:list-of-skills-SDA-claude-topics-pt2}, and \ref{tab:list-of-skills-SDA-claude-topics-pt3}.}
\centering
\begin{longtable}{p{8.5cm}p{8.5cm}p{8.5cm}}
\toprule
\textbf{Skills} &  \\
\toprule
\texttt{crisis\_management} & \texttt{intergovernmental\_coordination} \\
\texttt{civic\_engagement\_facilitation} & \texttt{human\_rights\_protection} \\
\texttt{legislative\_process\_navigation} & \texttt{judicial\_decision\_making} \\
\texttt{law\_enforcement\_administration} & \texttt{international\_treaty\_negotiation} \\
\texttt{journalism\_content\_production} & \texttt{research\_and\_fact\_checking} \\
\texttt{interview\_techniques} & \texttt{news\_writing} \\
\texttt{storytelling\_and\_narrative\_structure} & \texttt{copy\_editing} \\
\bottomrule
\end{longtable}
\end{sidewaystable}
\begin{sidewaystable}
\caption{(Part 26 of 52) 3266 skills extracted from topics in Tables~\ref{tab:list-of-skills-SDA-claude-topics-pt1}, \ref{tab:list-of-skills-SDA-claude-topics-pt2}, and \ref{tab:list-of-skills-SDA-claude-topics-pt3}.}
\centering
\begin{longtable}{p{8.5cm}p{8.5cm}p{8.5cm}}
\toprule
\textbf{Skills} &  \\
\toprule
\texttt{event\_planning\_and\_execution} & \texttt{revenue\_management} \\
\texttt{inventory\_control} & \texttt{staff\_training\_and\_development} \\
\texttt{guest\_satisfaction\_monitoring} & \texttt{crisis\_management} \\
\texttt{cultural\_sensitivity} & \texttt{upselling\_techniques} \\
\texttt{complaint\_handling} & \texttt{time\_management} \\
\texttt{team\_leadership} & \texttt{health\_and\_safety\_compliance} \\
\texttt{financial\_budgeting} & \texttt{marketing\_and\_promotion} \\
\bottomrule
\end{longtable}
\end{sidewaystable}
\begin{sidewaystable}
\caption{(Part 27 of 52) 3266 skills extracted from topics in Tables~\ref{tab:list-of-skills-SDA-claude-topics-pt1}, \ref{tab:list-of-skills-SDA-claude-topics-pt2}, and \ref{tab:list-of-skills-SDA-claude-topics-pt3}.}
\centering
\begin{longtable}{p{8.5cm}p{8.5cm}p{8.5cm}}
\toprule
\textbf{Skills} &  \\
\toprule
\texttt{composting\_and\_soil\_amendment} & \texttt{farm\_data\_management\_and\_analysis} \\
\texttt{crop\_yield\_forecasting} & \texttt{agricultural\_drone\_operation} \\
\texttt{vertical\_farming\_techniques} & \texttt{agroforestry\_implementation} \\
\texttt{post\_harvest\_handling\_and\_storage} & \texttt{food\_safety\_management} \\
\texttt{hazard\_analysis\_critical\_control\_point\_implementation} & \texttt{} \\
\texttt{sanitation\_standard\_operating\_procedures\_development} & \texttt{} \\
\texttt{food\_safety\_risk\_assessment} & \texttt{food\_allergen\_management} \\
\bottomrule
\end{longtable}
\end{sidewaystable}
\begin{sidewaystable}
\caption{(Part 28 of 52) 3266 skills extracted from topics in Tables~\ref{tab:list-of-skills-SDA-claude-topics-pt1}, \ref{tab:list-of-skills-SDA-claude-topics-pt2}, and \ref{tab:list-of-skills-SDA-claude-topics-pt3}.}
\centering
\begin{longtable}{p{8.5cm}p{8.5cm}p{8.5cm}}
\toprule
\textbf{Skills} &  \\
\toprule
\texttt{blockchain\_governance\_system\_development} & \texttt{data\_structure\_optimization} \\
\texttt{peer\_to\_peer\_network\_management} & \texttt{blockchain\_performance\_tuning} \\
\texttt{transaction\_validation\_algorithm\_design} & \texttt{blockchain\_api\_development} \\
\texttt{digital\_signature\_implementation} & \texttt{blockchain\_testing\_and\_debugging} \\
\texttt{blockchain\_integration\_with\_existing\_systems} & \texttt{} \\
\texttt{business\_strategy\_planning} & \texttt{market\_analysis} \\
\texttt{competitive\_intelligence\_gathering} & \texttt{swot\_analysis\_creation} \\
\bottomrule
\end{longtable}
\end{sidewaystable}
\begin{sidewaystable}
\caption{(Part 29 of 52) 3266 skills extracted from topics in Tables~\ref{tab:list-of-skills-SDA-claude-topics-pt1}, \ref{tab:list-of-skills-SDA-claude-topics-pt2}, and \ref{tab:list-of-skills-SDA-claude-topics-pt3}.}
\centering
\begin{longtable}{p{8.5cm}p{8.5cm}p{8.5cm}}
\toprule
\textbf{Skills} &  \\
\toprule
\texttt{energy\_efficiency\_assessment} & \texttt{power\_quality\_monitoring} \\
\texttt{fault\_detection\_and\_diagnosis} & \texttt{peak\_load\_management} \\
\texttt{energy\_market\_analysis} & \texttt{transmission\_loss\_reduction} \\
\texttt{voltage\_regulation} & \texttt{distributed\_generation\_management} \\
\texttt{energy\_policy\_implementation} & \texttt{carbon\_footprint\_calculation} \\
\texttt{energy\_auditing} & \texttt{power\_system\_stability\_analysis} \\
\texttt{energy\_cost\_optimization} & \texttt{learning\_strategy\_development} \\
\bottomrule
\end{longtable}
\end{sidewaystable}
\begin{sidewaystable}
\caption{(Part 30 of 52) 3266 skills extracted from topics in Tables~\ref{tab:list-of-skills-SDA-claude-topics-pt1}, \ref{tab:list-of-skills-SDA-claude-topics-pt2}, and \ref{tab:list-of-skills-SDA-claude-topics-pt3}.}
\centering
\begin{longtable}{p{8.5cm}p{8.5cm}p{8.5cm}}
\toprule
\textbf{Skills} &  \\
\toprule
\texttt{sports\_equipment\_analysis} & \texttt{statistical\_data\_analysis} \\
\texttt{athlete\_monitoring\_techniques} & \texttt{exercise\_physiology\_application} \\
\texttt{sports\_psychology\_implementation} & \texttt{motion\_capture\_data\_interpretation} \\
\texttt{fatigue\_management\_assessment} & \texttt{sports\_specific\_testing\_protocols} \\
\texttt{periodization\_planning} & \texttt{strength\_and\_conditioning\_programming} \\
\texttt{sports\_technology\_integration} & \texttt{performance\_enhancement\_strategies} \\
\texttt{research\_methodology\_application} & \texttt{textile\_production\_methods} \\
\bottomrule
\end{longtable}
\end{sidewaystable}
\begin{sidewaystable}
\caption{(Part 31 of 52) 3266 skills extracted from topics in Tables~\ref{tab:list-of-skills-SDA-claude-topics-pt1}, \ref{tab:list-of-skills-SDA-claude-topics-pt2}, and \ref{tab:list-of-skills-SDA-claude-topics-pt3}.}
\centering
\begin{longtable}{p{8.5cm}p{8.5cm}p{8.5cm}}
\toprule
\textbf{Skills} &  \\
\toprule
\texttt{thesis\_statement\_formulation} & \texttt{argument\_construction} \\
\texttt{literature\_review\_synthesis} & \texttt{citation\_management} \\
\texttt{academic\_tone\_maintenance} & \texttt{paragraph\_structuring} \\
\texttt{evidence\_evaluation} & \texttt{logical\_flow\_creation} \\
\texttt{abstract\_writing} & \texttt{proofreading\_and\_editing} \\
\texttt{plagiarism\_avoidance} & \texttt{academic\_vocabulary\_usage} \\
\texttt{data\_analysis\_interpretation} & \texttt{scholarly\_source\_identification} \\
\bottomrule
\end{longtable}
\end{sidewaystable}
\begin{sidewaystable}
\caption{(Part 32 of 52) 3266 skills extracted from topics in Tables~\ref{tab:list-of-skills-SDA-claude-topics-pt1}, \ref{tab:list-of-skills-SDA-claude-topics-pt2}, and \ref{tab:list-of-skills-SDA-claude-topics-pt3}.}
\centering
\begin{longtable}{p{8.5cm}p{8.5cm}p{8.5cm}}
\toprule
\textbf{Skills} &  \\
\toprule
\texttt{legal\_system\_analysis} & \texttt{legal\_research\_and\_analysis} \\
\texttt{case\_law\_interpretation} & \texttt{statutory\_interpretation} \\
\texttt{legal\_writing\_and\_drafting} & \texttt{comparative\_law\_analysis} \\
\texttt{legal\_precedent\_application} & \texttt{legal\_argument\_construction} \\
\texttt{legal\_system\_evaluation} & \texttt{judicial\_decision\_analysis} \\
\texttt{legal\_policy\_assessment} & \texttt{legal\_ethics\_application} \\
\texttt{legal\_reform\_proposal\_development} & \texttt{legal\_data\_visualization} \\
\bottomrule
\end{longtable}
\end{sidewaystable}
\begin{sidewaystable}
\caption{(Part 33 of 52) 3266 skills extracted from topics in Tables~\ref{tab:list-of-skills-SDA-claude-topics-pt1}, \ref{tab:list-of-skills-SDA-claude-topics-pt2}, and \ref{tab:list-of-skills-SDA-claude-topics-pt3}.}
\centering
\begin{longtable}{p{8.5cm}p{8.5cm}p{8.5cm}}
\toprule
\textbf{Skills} &  \\
\toprule
\texttt{multiplayer\_networking} & \texttt{game\_balancing} \\
\texttt{project\_management} & \texttt{quality\_assurance\_testing} \\
\texttt{health\_and\_medicine} & \texttt{diagnose\_medical\_conditions} \\
\texttt{perform\_physical\_examinations} & \texttt{interpret\_lab\_results} \\
\texttt{administer\_medications} & \texttt{conduct\_surgical\_procedures} \\
\texttt{develop\_treatment\_plans} & \texttt{monitor\_patient\_vital\_signs} \\
\texttt{provide\_emergency\_medical\_care} & \texttt{analyze\_medical\_imaging} \\
\bottomrule
\end{longtable}
\end{sidewaystable}
\begin{sidewaystable}
\caption{(Part 34 of 52) 3266 skills extracted from topics in Tables~\ref{tab:list-of-skills-SDA-claude-topics-pt1}, \ref{tab:list-of-skills-SDA-claude-topics-pt2}, and \ref{tab:list-of-skills-SDA-claude-topics-pt3}.}
\centering
\begin{longtable}{p{8.5cm}p{8.5cm}p{8.5cm}}
\toprule
\textbf{Skills} &  \\
\toprule
\texttt{battlefield\_network\_administration} & \texttt{weapons\_system\_integration} \\
\texttt{threat\_detection\_analysis} & \texttt{command\_and\_control\_optimization} \\
\texttt{sensor\_fusion\_implementation} & \texttt{ballistic\_missile\_defense\_operation} \\
\texttt{electromagnetic\_spectrum\_management} & \texttt{combat\_simulation\_modeling} \\
\texttt{data\_link\_system\_configuration} & \texttt{stealth\_technology\_application} \\
\texttt{directed\_energy\_weapon\_operation} & \texttt{artificial\_intelligence\_integration} \\
\texttt{quantum\_computing\_utilization} & \texttt{mathematics\_and\_statistics} \\
\bottomrule
\end{longtable}
\end{sidewaystable}
\begin{sidewaystable}
\caption{(Part 35 of 52) 3266 skills extracted from topics in Tables~\ref{tab:list-of-skills-SDA-claude-topics-pt1}, \ref{tab:list-of-skills-SDA-claude-topics-pt2}, and \ref{tab:list-of-skills-SDA-claude-topics-pt3}.}
\centering
\begin{longtable}{p{8.5cm}p{8.5cm}p{8.5cm}}
\toprule
\textbf{Skills} &  \\
\toprule
\texttt{research\_project\_management} & \texttt{construction\_safety\_standards} \\
\texttt{risk\_assessment\_and\_hazard\_identification} & \texttt{} \\
\texttt{personal\_protective\_equipment\_selection} & \texttt{safety\_plan\_development} \\
\texttt{osha\_regulation\_compliance} & \texttt{fall\_protection\_system\_implementation} \\
\texttt{scaffolding\_safety\_inspection} & \texttt{heavy\_machinery\_operation\_safety} \\
\texttt{confined\_space\_entry\_procedures} & \texttt{emergency\_response\_planning} \\
\texttt{safety\_training\_program\_design} & \texttt{accident\_investigation\_and\_reporting} \\
\bottomrule
\end{longtable}
\end{sidewaystable}
\begin{sidewaystable}
\caption{(Part 36 of 52) 3266 skills extracted from topics in Tables~\ref{tab:list-of-skills-SDA-claude-topics-pt1}, \ref{tab:list-of-skills-SDA-claude-topics-pt2}, and \ref{tab:list-of-skills-SDA-claude-topics-pt3}.}
\centering
\begin{longtable}{p{8.5cm}p{8.5cm}p{8.5cm}}
\toprule
\textbf{Skills} &  \\
\toprule
\texttt{launch\_vehicle\_integration} & \texttt{data\_analysis\_methods} \\
\texttt{data\_cleaning} & \texttt{statistical\_analysis} \\
\texttt{data\_visualization} & \texttt{hypothesis\_testing} \\
\texttt{regression\_analysis} & \texttt{time\_series\_analysis} \\
\texttt{cluster\_analysis} & \texttt{principal\_component\_analysis} \\
\texttt{exploratory\_data\_analysis} & \texttt{predictive\_modeling} \\
\texttt{data\_interpretation} & \texttt{feature\_selection} \\
\bottomrule
\end{longtable}
\end{sidewaystable}
\begin{sidewaystable}
\caption{(Part 37 of 52) 3266 skills extracted from topics in Tables~\ref{tab:list-of-skills-SDA-claude-topics-pt1}, \ref{tab:list-of-skills-SDA-claude-topics-pt2}, and \ref{tab:list-of-skills-SDA-claude-topics-pt3}.}
\centering
\begin{longtable}{p{8.5cm}p{8.5cm}p{8.5cm}}
\toprule
\textbf{Skills} &  \\
\toprule
\texttt{social\_media\_optimization} & \texttt{digital\_asset\_management} \\
\texttt{video\_compression\_techniques} & \texttt{interactive\_media\_development} \\
\texttt{user\_experience\_design} & \texttt{project\_management} \\
\texttt{copyright\_law\_understanding} & \texttt{audience\_analysis} \\
\texttt{content\_strategy\_development} & \texttt{cross\_platform\_adaptation} \\
\texttt{negotiation\_strategy\_development} & \texttt{active\_listening} \\
\texttt{conflict\_resolution} & \texttt{emotional\_intelligence} \\
\bottomrule
\end{longtable}%
\end{sidewaystable}

\clearpage

\begin{sidewaystable}
\caption{(Part 1 of 2) 13 Query/Task Types extracted from interactions with Claude-3.5-Sonnet}
\label{tab:list-of-skills-SDA-query-types-claude}
\centering
%
\end{sidewaystable}
\begin{sidewaystable}
\caption{(Part 2 of 2) 13 Query/Task Types extracted from interactions with Claude-3.5-Sonnet}
\centering
%
\end{sidewaystable}

\clearpage
\end{small}

\clearpage

\clearpage

\section{Skill Extraction Prompts}
\appendixtoc{Skill Extraction Prompts}

\subsection{Prompt for Skill Extraction (\instructSMSDD)}\label{appendix:skill-extraction-SDD}
\begin{question}
Consider the following question. Label this question with a skill that would be required to solve the question. Basically, you should be able to use the skill as a dictionary key in python. The skill name should be lower case letters only. The skill name should be very descriptive and you may use multiple words to describe the skills required in the question. If you do use multiple words per question, then join them by an underscore.\\\{text\}\\Your answer should be as follows:\\<name of the skill>, reason: <reason for the skill>
\end{question}

\subsection{Prompt for Semantic Clustering (\instructSMSDD)}\label{appendix:semantic-clustering-SDD}

\begin{question}
Given the list of skills required to solve various questions, your task is to categorize these skills into descriptive and specific groups. Each category should not only capture the essence of the skills it includes but also reflect clear, distinct areas of expertise or application. Use terminology that is precise and specific to the tasks those skills accomplish. Categories should be narrow enough to provide meaningful insight into the specialization they represent. Format category names in lowercase, joining multiple words with underscores.\\For each category you create, provide a comprehensive rationale explaining:\\- Why these particular skills are grouped together.\\- How the category name specifically reflects the specialization and distinctiveness of the skills.\\\\\#\#\# Skills List:\\\{skills\_joined\_str\}\\\\\#\#\# Examples:\\- Category name: `web\_development`\\- Included skills: html\_css\_design, javascript\_interactivity, web\_performance\_optimization\\- Rationale: This category focuses specifically on the skills required to build and optimize web interfaces, distinguishing it from broader software development.\\\\- Category name: `machine\_learning\_techniques`\\- Included skills: machine\_learning\_model\_creation, understanding\_algorithmic\_fairness, machine\_learning\_application\_in\_finance\\- Rationale: These skills are grouped under 'machine\_learning\_techniques' because they directly relate to the development and application of machine learning models, emphasizing specific use cases and ethical considerations, which are distinct from general programming skills.\\\\Please ensure your category names are informative, specific, and reflective of the skills grouped under them. This will aid in their practical utility and clarity.
\end{question}
\clearpage

\subsection{Prompt for Creating Synthetic Data (\instructSMSDD)}\label{appendix:synthetic-data-creation-SDD}

\textbf{Prompt 1}
\begin{question}
I am creating a dataset of engaging human-AI interactions that demonstrate the capabilities and value of AI models across diverse skills. Please help me generate high-quality examples for this dataset.\\\\I will first give you a set of \{num\_skills\} skills that are valuable to humans.\\\\Next, you would create a query that a human might plausibly ask whose answer involves using those \{num\_skills\} skills. \\\\For this query you will generate a good length answer using your best helpfulness and wisdom. Please try to avoid generic sound queries and answers. People often have concrete problems and would give important details. The answer should also give some concrete details and examples, but not get verbose.\\\\Before I give you the set of \{num\_skills\} skills, I will provide some examples involving 2 skills.\\Example 1: \\\#\#\# Skills:  self\_reflection\_skills, linguistics\\\#\#\# Instruction: I often find myself in conflicts at work due to miscommunication and differing perspectives. How can I improve my conflict resolution skills by incorporating self-reflection and effective communication techniques into my daily interactions with colleagues?\\\#\#\# Response: To enhance your conflict resolution skills in the workplace, integrate these self-reflection and linguistic strategies into your daily routine:\\Self-Reflection Techniques:\\End-of-Day Review: Spend 10 minutes reflecting on interactions, noting successes and areas for improvement. For example, \"I successfully de-escalated a tense situation with a client by actively listening to their concerns and finding a mutually beneficial solution.\"\\Conflict Log: Record details of conflicts, your responses, and outcomes in a dedicated journal. Analyze patterns monthly to identify growth opportunities. For instance, \"I noticed I tend to become defensive when criticized. I will practice pause and respond techniques to improve my reactions.\"\\Linguistic Techniques:\\Empathy Statements: Use phrases that demonstrate understanding and validate others' perspectives, such as \"I can see why you feel that way\" or \"That must be challenging for you.\" In a recent meeting, saying \"I understand your frustration with the project timeline\" helped a coworker feel heard and open to finding a solution together.\\Reflective Listening: Paraphrase the other person's key points before responding to ensure clarity and show you're listening. For example, \"So, what I'm hearing is that you're concerned about the budget allocation. Let's discuss some options to address that.\"\\Integration into Routine:\\Weekly Practice Sessions: Dedicate an hour each week to role-play conflict scenarios with a trusted colleague, focusing on applying self-reflection insights and linguistic techniques. Provide each other with constructive feedback for continuous improvement.\\Real-Time Application: When a conflict arises, take a moment to mentally review your phrase bank and select an appropriate response. After the interaction, briefly reflect on what worked well and what could be improved next time.\\By consistently practicing these strategies, you'll build stronger habits and enhance your overall conflict resolution skills in the workplace.\\Example 2: \\\#\#\# Skills: psychological\_understanding, data\_handling\_and\_analysis\\\#\#\# Instruction: I recently started working as a data analyst for a meal kit delivery service. We've noticed a significant increase (20\%) in customer churn over the past quarter. How can I leverage data analysis and psychological principles to identify the root causes of churn and develop targeted strategies to improve customer retention?\\\#\#\# Response: To address the 20\% increase in customer churn for your meal kit service, employ a combination of data analysis and psychological insights:\\Data Analysis:\\Customer Segmentation: Analyze churn rates by customer segments such as age, location, and dietary preferences. For example, if you find that vegetarian customers have a higher churn rate, you can focus on improving vegetarian meal options.\\Churn Timing Analysis: Identify common timeframes when churn occurs, such as after the first month or after specific promotions end. If churn spikes after the first month, consider implementing a more robust onboarding process or extending promotional offers.\\Psychological Insights:\\Exit Surveys: Conduct exit surveys to gather direct feedback from churning customers. Include open-ended questions to uncover deeper insights. For instance, \"What was the primary reason for canceling your subscription?\" and \"What could we have done better to keep you as a customer?\"\\Customer Journey Mapping: Map out the customer journey to identify pain points and opportunities for improvement. Apply psychological principles like the Peak-End Rule to prioritize the most impactful moments. Focus on creating a positive onboarding experience and ensuring a satisfying last interaction before cancellation.\\Actionable Strategies:\\Personalized Retention Campaigns: Develop targeted email campaigns for at-risk segments, addressing their specific concerns. For example, send vegetarian customers an email highlighting new plant-based recipes and offering a discount on their next box.
\end{question}
\begin{question}
Flexible Subscription Options: Introduce more flexible subscription plans, such as the ability to skip weeks or customize delivery frequency. This addresses the psychological need for autonomy and control, reducing the likelihood of churn.\\By leveraging data-driven insights and psychological principles, you can create a targeted approach to reduce churn and improve customer satisfaction in your meal kit service.\\
Example 3:\\\#\#\# Skills: food\_related\_knowledge\_and\_skills, internet\_technologies\\\#\#\# Instruction: I'm a food truck owner looking to expand my online presence and attract more customers. I want to create a website that allows customers to customize their meals, track my truck's real-time location, and place orders for pickup. What web technologies should I consider to build a user-friendly and feature-rich website for my food truck business?\\\#\#\# Response: To create a robust and interactive food truck website with meal customization and real-time location tracking, consider the following web technologies:\\Front-End:\\React: Use React, a popular JavaScript library, for building a dynamic and responsive user interface. React's component-based architecture allows for easy management of real-time updates, such as displaying the current location of the food truck.\\HTML5 Geolocation API: Leverage the HTML5 Geolocation API to capture the user's location (with their permission) and display nearby food truck locations on an interactive map.\\Leaflet or Google Maps: Integrate Leaflet, an open-source mapping library, or Google Maps to display food truck locations and enable users to get directions. Customize the map markers to show additional information like menu items or special offers.\\Back-End:\\Node.js and Express: Use Node.js, a server-side JavaScript runtime, and the Express web application framework to handle server-side logic and API routing. Node.js's non-blocking I/O model is well-suited for real-time applications like location tracking.\\WebSocket: Implement WebSocket communication to enable real-time updates between the server and the client. When a food truck's location changes, the server can push the updated coordinates to connected clients, ensuring users always see the most current location.\\Firebase Realtime Database: Store and sync food truck location data in real-time using Firebase Realtime Database. This NoSQL cloud database allows for efficient storage and retrieval of location data, enabling seamless real-time updates across multiple devices.\\User Experience:\\Progressive Web App (PWA): Develop the website as a PWA to provide a native app-like experience. PWAs offer features like offline accessibility, push notifications, and home screen installation, enhancing user engagement and convenience.\\Responsive Design: Ensure the website is fully responsive and optimized for various screen sizes, particularly mobile devices. Use CSS media queries and flexible layouts to create a seamless experience across desktops, tablets, and smartphones.\\Accessibility: Follow web accessibility guidelines (WCAG) to ensure the website is usable by people with disabilities. Implement proper headings, alt text for images, and keyboard navigation to create an inclusive user experience.\\By combining these web technologies, you can create a feature-rich and user-friendly food truck website that offers meal customization and real-time location tracking, providing a seamless experience for your customers.\\Example 4: \\\#\#\# Skills: algorithmic\_and\_programming\_skills, digital\_marketing\\\#\#\# Instruction: As a software engineer, I've developed a sophisticated recommendation engine that I believe could greatly benefit e-commerce businesses. I want to showcase my work through a technical blog post and live demo to attract potential clients and demonstrate the value of my solution. How can I structure my blog post and demo to effectively highlight the technical aspects and business benefits of my recommendation engine?\\\#\#\# Response: To effectively showcase your recommendation engine and attract potential e-commerce clients, structure your technical blog post and live demo as follows:\\Blog Post Structure:\\Attention-Grabbing Introduction: Start with a compelling hook that highlights the value of personalized recommendations in e-commerce. Use statistics or real-world examples to demonstrate the impact on sales and customer satisfaction.\\Problem Statement: Clearly define the challenges e-commerce businesses face in providing relevant product recommendations at scale. Discuss common pain points like data sparsity, cold-start problems, and real-time updates.\\Technical Deep Dive: Explain the core components of your recommendation engine, such as collaborative filtering, content-based filtering, or hybrid approaches. Use diagrams and code snippets to illustrate your architecture and key algorithms. Highlight any innovative techniques you've employed, such as deep learning or reinforcement learning.\\Performance Metrics: Present quantitative results that showcase the effectiveness of your recommendation engine. Include metrics like precision, recall, F1 score, and mean average precision. Compare your results to industry benchmarks or popular open-source recommendation libraries to demonstrate your engine's superiority.\\Scalability and Efficiency: Discuss how your recommendation engine handles large-scale data and real-time updates. Explain your strategies for efficient data processing, such as parallel computing or incremental updates. Provide performance benchmarks to highlight the speed and scalability of your solution. \\Live Demo: E-commerce Store Integration: Create a mock e-commerce store that seamlessly integrates your recommendation engine. Showcase personalized product recommendations based on user interactions, such as viewed items, purchases, or ratings.
\end{question}
\begin{question}
Real-Time Recommendations: Demonstrate how your engine adapts in real-time as users navigate the store. For example, show how the recommendations update dynamically based on the user's browsing history or cart contents.\\Explanations and Transparency: Provide clear explanations for each recommendation, such as \"Customers who bought this item also bought...\" or \"Recommended based on your recent searches.\" This transparency builds trust and helps users understand the relevance of the recommendations.\\
A/B Testing: Implement an A/B testing feature that allows potential clients to compare the performance of your recommendation engine against a default or random recommendation system. Visualize the results through metrics like click-through rates or conversion rates to highlight the tangible benefits of your solution.\\Promotion and Engagement:\\Targeted Social Media Ads: Create compelling social media ads that highlight the key features and benefits of your recommendation engine. Target decision-makers in the e-commerce industry, such as marketers, product managers, and CTOs. Use eye-catching visuals and clear calls-to-action to drive traffic to your blog post and live demo.\\LinkedIn Thought Leadership: Share your blog post on LinkedIn and engage in discussions related to e-commerce personalization. Participate in relevant groups and forums to establish yourself as a thought leader in the field. Offer valuable insights and answer questions to build trust and credibility with potential clients.\\Webinars and Workshops: Host webinars or workshops that provide a hands-on experience with your recommendation engine. Walk participants through the setup process and demonstrate how to integrate your solution into their existing e-commerce platforms. Offer Q\&A sessions to address specific concerns and showcase your expertise.\\By structuring your technical blog post and live demo in this manner, you can effectively showcase the capabilities and benefits of your recommendation engine. Combine this with targeted digital marketing techniques to reach and engage potential e-commerce clients, ultimately driving interest and adoption of your solution.
Example 5: \\\#\#\# Skills: geographical\_and\_historical\_knowledge, analytical\_thinking\\\#\#\# Instruction: I'm a history enthusiast planning a two-week trip to Eastern Europe with my family. We're particularly interested in exploring medieval architecture and learning about the region's experiences during the Cold War era. How can I create an itinerary that combines visits to historical sites, museums, and cultural experiences to gain a deeper understanding of Eastern Europe's past and present?\\\#\#\# Response: Here's a curated itinerary for a historically rich two-week trip through Eastern Europe, focusing on medieval and Cold War sites:\\Week 1: Poland and Czech Republic\\Krakow (3 days):\\Wawel Castle: Explore this 14th-century castle complex, which served as the residence of Polish kings. Marvel at the Gothic and Renaissance architecture, and visit the Dragon's Den, a legendary cave beneath the castle.\\Oskar Schindler's Factory Museum: Learn about the Holocaust and Oskar Schindler's efforts to save Jewish workers during World War II. The museum offers a immersive experience, recreating the wartime atmosphere of Krakow.\\Nowa Huta: Take a guided tour of this planned socialist city, built during the Cold War era. Visit the iconic Lord's Ark Church, which became a symbol of resistance against the communist regime.\\Warsaw (2 days):\\Old Town: Stroll through the meticulously reconstructed Old Town, which was destroyed during World War II. Visit the Market Square, the Warsaw Barbican, and St. John's Cathedral to admire the Gothic and Renaissance architecture.\\Warsaw Uprising Museum: Discover the story of the 1944 Warsaw Uprising, a major resistance operation against Nazi occupation. The museum features interactive exhibits, original artifacts, and a 3D movie that brings history to life.\\Palace of Culture and Science: Explore this imposing Stalinist-era skyscraper, which remains the tallest building in Poland. Take an elevator to the observation deck for panoramic views of Warsaw.\\Prague (2 days):\\Prague Castle: Visit the world's largest ancient castle complex, dating back to the 9th century. Explore the Gothic St. Vitus Cathedral, the Romanesque St. George's Basilica, and the Golden Lane, a picturesque row of colorful houses.\\Charles Bridge: Walk across this iconic 14th-century stone bridge, lined with baroque statues. Enjoy street musicians and artists, and take in the stunning views of the Vltava River and the Old Town.\\Nuclear Bunker Tour: Experience the chilling reality of the Cold War by visiting a former Soviet nuclear bunker. Learn about the strategies and technology employed during this tense period in history.\\Week 2: Hungary and Romania\\Budapest (3 days):\\Buda Castle: Discover this 13th-century castle, which has been rebuilt and expanded over the centuries. Visit the Hungarian National Gallery, the Budapest History Museum, and the Matthias Church, known for its colorful tiled roof.\\House of Terror Museum: Explore this powerful museum dedicated to the victims of the fascist and communist regimes in Hungary. The exhibits are housed in the former headquarters of the Arrow Cross Party and the communist secret police.\\Memento Park: Visit this open-air museum showcasing monumental statues from the Soviet era. Learn about the propaganda and ideology behind these imposing sculptures.\\Bucharest (2 days):\\Palace of Parliament: Tour the world's largest civilian building, constructed during the communist era under the rule of Nicolae Ceaușescu.
\end{question}
\begin{question}
 Marvel at the opulent interiors and learn about the controversial history of this massive structure.\\Old Town: Explore the charming streets of Bucharest's Old Town, lined with historical buildings, churches, and cafes. Visit the ruins of the Old Princely Court, which served as the residence of Wallachian princes.\\
Revolution Square: Pay tribute to the heroes of the 1989 Romanian Revolution at this significant square. See the Memorial of Rebirth, which honors those who lost their lives fighting against the communist regime.\\Travel Tips:\\Book guided tours with knowledgeable local guides to gain deeper insights into the historical context and personal stories behind each site.\\Stay in centrally located accommodation to minimize travel time and maximize your exploration of each city.\\Use public transportation or ride-sharing services to navigate within cities, as parking can be challenging and expensive in historical areas.\\By following this itinerary, you'll embark on a fascinating journey through Eastern Europe's medieval and Cold War past, gaining a profound appreciation for the region's rich history and resilience.\\\\Now, it is your turn! Given the guidelines and examples above, please create a query that a human might plausibly ask whose answer involves using all skills below:\\\#\#\# Skills: \{skills\_str\}
\end{question}

\textbf{Prompt 2}
\begin{question}
Thanks! It looks like the response may have been cut off. Please rewrite the entire response, making sure it fits within the length constraint. No need to include verbose or extraneous information. Thanks!
\end{question}

\textbf{Prompt 3}
\begin{question}
Pretend you are the human asking the question, and make a list of weaknesses and strengths of the response. Currently the response looks a bit generic. Perhaps it could benefit from some concrete examples and details.
\end{question}

\textbf{Prompt 4}
\begin{question}
Now, please refine your instruction and response (keep the strong parts and work on the weak parts).
\end{question}

\textbf{Prompt 5}
\begin{question}
It looks like the improved response may have been cut off. Please rewrite the entire improved response, making sure it fits within the length constraint. No need to include verbose or extraneous information. Thanks again!
\end{question}

\subsection{Prompt for Extrapolating Skills (\instructSMSDA)}\label{appendix:skill-extraction-SDA}
\begin{question}
First, consider the following definitions:\\
- Knowledge: A set of information, facts, descriptions, and theories acquired through education or experience.\\
- Topic: A subset of Knowledge that focuses on a particular area or subject.\\
- Skill: A function that transforms Knowledge into actions or performances that achieve specific outcomes. \\ \\
Now, for the topic of {topic}, please provide me with multiple skills (in snake case) that could be relevant to the topic.
\end{question}

\subsection{Prompt for Creating Synthetic Data (\instructSMSDA)}\label{appendix:synthetic-data-creation-SDA}
\textbf{Prompt 1}
\begin{question}
I am creating a dataset of engaging human-AI interactions that demonstrate the capabilities and value of AI models across diverse skills. Please help me generate high-quality examples for this dataset.\\\\I will first give you a set of \{num\_skills\} skills that are valuable to humans.\\\\Next, you would create a query that a human might plausibly ask whose answer involves using those \{num\_skills\} skills. \\\\For this query you will generate a good length answer using your best helpfulness and wisdom. Please try to avoid generic sound queries and answers. The answer should also give some concrete details and examples, but not get verbose.\\\\Before I give you the set of \{num\_skills\} skills, I will provide some examples involving 2 skills.\\Example 1: \\\#\#\# Task Type: Information-Seeking\\\#\#\# Skills: psychological\_understanding, data\_handling\_and\_analysis\\\#\#\# Instruction: What are the primary psychological theories that explain the impact of social media on adolescent self-esteem, and what evidence supports these theories? Additionally, what statistical techniques have been used in the literature to dissect these relationships, and how do these techniques help in understanding the nuances of psychological data?\\\#\#\# Response: The impact of social media on adolescent self-esteem is a widely studied topic within psychology. Various theories and statistical techniques are employed to understand and measure these effects. Below, I'll outline some of the primary theories, the supporting evidence, and the statistical techniques used in this research.\\\#\#\# Response: The impact of social media on adolescent self-esteem is a widely studied topic within psychology. Various theories and statistical techniques are employed to understand and measure these effects. Below, I'll outline some of the primary theories, the supporting evidence, and the statistical techniques used in this research.\\\\ **Psychological Theories and Evidence**\\\\Social Comparison Theory: Adolescents often assess their worth by comparing themselves with peers on social media. A study by Vogel et al. (2014) found that Facebook usage was associated with negative self-perceptions when users compared themselves to others they perceived as better off. This suggests a direct correlation between the frequency of social media use and increased feelings of inferiority.\\\\Self-Presentation Theory: Adolescents feel pressure to present themselves in an idealized way on platforms like Instagram and Snapchat. Research by Chou and Edge (2012) highlights that the need for positive feedback and fear of negative evaluations can lead to cycles of anxiety and lowered self-esteem when these expectations are not met.\\\\Media Richness Theory: Different types of social media platforms impact self-esteem differently. A study by Sherman et al. (2016) compared text-based versus image-based platforms, finding that image-based platforms (like Instagram) can exacerbate body image concerns and anxiety more than text-based platforms due to the intense focus on physical appearance.\\\\Statistical Techniques and Applications\\\\Regression Analysis: Used extensively to quantify the relationship between social media usage and self-esteem. For instance, studies might control for variables like age, gender, and socioeconomic status to isolate the effect of social media usage.\\\\Structural Equation Modeling (SEM): SEM enables researchers to construct a comprehensive model of how different factors like peer influence, media consumption, and personal traits interact to affect self-esteem. An example is Kross et al.’s (2013) study, which modeled the pathways through which social media affects mood and self-esteem over time.\\\\Longitudinal Studies and Growth Curve Modeling: These techniques provide insights into how self-esteem trajectories change over time with social media use. For instance, longitudinal research might track self-esteem changes from early to late adolescence, revealing critical periods where impacts are most pronounced.\\\\Mediation and Moderation Analyses: These are crucial for exploring underlying mechanisms and boundary conditions. For example, studies might explore how the negative impact of social media on self-esteem can be mediated by social support or moderated by individual resilience.\\\\ **Contradictory Evidence**\\\\While many studies suggest a negative impact of social media on self-esteem, some research offers a different perspective. For instance, Valkenburg et al. (2006) found that online interactions can enhance well-being by providing social support and a platform for self-expression. These findings suggest that the impact of social media on self-esteem is not universally negative and can vary based on usage patterns and individual differences.\\\\ **Practical Implications**\\\\Understanding these dynamics is crucial for developing targeted interventions. For instance, educational programs that teach critical media literacy can help adolescents navigate social media healthily. Moreover, knowing which aspects of social media use trigger negative self-comparisons can guide the development of features that promote positive interaction and reduce the emphasis on appearance.\\\\Example 2: \\\#\#\# Task Type: Help-Seeking\\\#\#\# Skills:  self\_reflection\_skills, linguistics\\\#\#\# Instruction: I often find myself in conflicts at work due to miscommunication and differing perspectives. How can I improve my conflict resolution skills by incorporating self-reflection and effective communication techniques into my daily interactions with colleagues?\\\#\#\# Response: To enhance your conflict resolution skills in the workplace, integrate these self-reflection and linguistic strategies into your daily routine:\\Self-Reflection Techniques:\\End-of-Day Review: Spend 10 minutes reflecting on interactions, noting successes and areas for improvement. For example, \"I successfully de-escalated a tense situation with a client by actively listening to their concerns and finding a mutually beneficial solution.\"\\Conflict Log: Record details of conflicts, your responses, and outcomes in a dedicated journal. Analyze patterns monthly to identify growth opportunities. For instance, \"I noticed I tend to become defensive when criticized. I will practice pause and respond techniques to improve my reactions.\"\\Linguistic Techniques:\\Empathy Statements: Use phrases that demonstrate understanding and validate others' perspectives, such as \"I can see why you feel that way\" or \"That must be challenging for you.\" In a recent meeting, saying \"I understand your frustration with the project timeline\" helped a coworker feel heard and open to finding a solution together.\\Reflective Listening: Paraphrase the other person's key points before responding to ensure clarity and show you're listening. For example, \"So, what I'm hearing is that you're concerned about the budget allocation. Let's discuss some options to address that.\"\\Integration into Routine:\\Weekly Practice Sessions: Dedicate an hour each week to role-play conflict scenarios with a trusted colleague, focusing on applying self-reflection insights and linguistic techniques. Provide each other with constructive feedback for continuous improvement.\\Real-Time Application: When a conflict arises, take a moment to mentally review your phrase bank and select an appropriate response. After the interaction, briefly reflect on what worked well and what could be improved next time.\\By consistently practicing these strategies, you'll build stronger habits and enhance your overall conflict resolution skills in the workplace.\\\\Example 3: \\\#\#\# Task Type: Information-Seeking\\\#\#\# Skills: geographical\_and\_historical\_knowledge, analytical\_thinking\\\#\#\# Instruction: What were the major geopolitical changes in Europe following World War II, and how have these changes influenced current European political boundaries?\\\#\#\# Response: **Major Geopolitical Changes in Europe After World War II**\\\\1. Division of Germany: The division of Germany into East and West Germany in 1949 was a critical symbol of the Cold War's onset. This division, marked physically by the Berlin Wall from 1961, lasted until the reunification in 1990 and significantly influenced European political dynamics.\\\\2. Creation of the Iron Curtain: Winston Churchill's term, the \"Iron Curtain,\" described the ideological and physical boundary separating the Soviet-controlled Eastern Bloc and the Western democracies. This division dictated political alignments and conflicts during the Cold War, influencing policies and international relations.\\\\3. Decolonization: Following the war, European powers like Britain and France began a significant process of decolonization. This period saw the independence of nations such as India in 1947 and the rapid decolonization of Africa in the 1960s, marking a major shift in global dynamics and European foreign policies.\\\\4. Formation of the European Union (EU): The EU's origins lie in the European Coal and Steel Community in 1951, evolving into the European Economic Community by 1957. These alliances, expanding to include more countries over the decades, aimed to foster economic cooperation and prevent further wars in Europe, influencing both economic and political policies within the continent.\\\\5. NATO and the Warsaw Pact: The establishment of NATO in 1949 by Western countries was a strategic move for collective security against the Soviet threat. The Soviet response, the Warsaw Pact in 1955, defined the military alliances in Europe, solidifying the East-West divide.\\\\ **Influence on Current European Political Boundaries**\\\\1. German Reunification: The fall of the Berlin Wall in 1989 and the subsequent reunification of East and West Germany in 1990 reshaped Germany's role in Europe, altering both its internal and external political boundaries.\\\\2. EU Expansion: The EU's expansion has included many former Eastern Bloc countries, fundamentally changing the political landscape of Europe. The Schengen Agreement, implemented in 1995, minimized the importance of national boundaries within the EU, promoting free movement and economic integration.\\\\3. Breakup of Yugoslavia and the Soviet Union: The disintegration of Yugoslavia into seven successor states throughout the 1990s and the Soviet Union into 15 independent countries in 1991 dramatically redrew political boundaries. These events, rooted in ethnic tensions and political upheavals, continue to influence regional stability and alignments.\\\\ **Case Studies: Key Treaties and Shifts in Alliances**\\\\1. Treaty of Paris (1951): This treaty established the European Coal and Steel Community, a foundational step towards European integration. It set precedents for future economic policies and cooperative frameworks within Europe, promoting peace and economic stability across former wartime adversaries.\\\\2. Impact of NATO and the Warsaw Pact on Poland: Poland's transition from a Warsaw Pact member to a NATO member in 1999 exemplifies the dramatic shift in military and political alliances. This change significantly influenced Poland's foreign and security policies, reflecting broader geopolitical shifts in Eastern Europe post-Cold War.\\\\\\\\Now, it is your turn! Given the guidelines and examples above, please create a query that aligns with the following task type:\\\{query\_type\}\\\\ Include specific text snippets when they are essential to analyze and validate your query effectively. Make sure that the query is one that a human might plausibly ask whose answer involves using all skills below:\\\#\#\# Task Type: \{query\_str\}\\\#\#\# Skills: \{skills\_str\}
\end{question}

\textbf{Prompt 2}
\begin{question}
Thanks! It looks like the response may have been cut off. Please rewrite the entire response, making sure it fits within the length constraint. No need to include verbose or extraneous information. Thanks!
\end{question}

\textbf{Prompt 3}
\begin{question}
Pretend you are the human asking the question, and make a list of weaknesses and strengths of the response. Currently the response looks a bit generic. Perhaps it could benefit from some concrete examples and details.
\end{question}

\textbf{Prompt 4}
\begin{question}
Now, please refine your instruction and response (keep the strong parts and work on the weak parts).
\end{question}

\textbf{Prompt 5}
\begin{question}
It looks like the improved response may have been cut off. Please rewrite the entire improved response, making sure it fits within the length constraint. No need to include verbose or extraneous information. Thanks again!
\end{question}
\clearpage

\section{Comparison of Responses}
\appendixtoc{Comparison of Responses}

In this section, we show side-by-side comparisons of responses from various models to the following instruction from AlpacaEval. 

\begin{question}
I'm trying to teach myself to have nicer handwriting. Can you help?
\end{question}

Observe that responses produced by models finetuned on \instructSM{} tend to include concrete examples (whereas responses produced by the other models do not always).

\textbf{Response by \llamaiii{8} SFT on \instructSMSDA{(k=2)}-4K}
\begin{question}
Absolutely! Improving your handwriting involves both technique and practice. Here’s a concise guide to help you achieve nicer handwriting:

**1. Posture and Grip:** \\
   - **Posture:** Sit upright with your paper at a 45-degree angle to reduce arm and hand movement, which helps prevent smudging. \\
   - **Grip:** Use a relaxed, comfortable grip with your pen or pencil resting lightly in your fingers. This promotes smoother movement and reduces pressure on the paper.

**2. Pen Lifts:** \\
   - **When to Lift:** Between words and sentences to create clear breaks, especially in cursive. \\
   - **Purpose:** Pen lifts help maintain the integrity of each letter, preventing them from blending together and enhancing readability.

**3. Slant and Direction:** \\
   - **Natural Slant:** Write with a natural slant that's comfortable for your hand, typically a slight upward slope helps legibility. \\
   - **Direction:** Write continuously without lifting your pen too much, maintaining a fluid motion can enhance the flow and elegance of your handwriting.

**4. Pressure and Speed:** \\
   - **Pressure:** Apply consistent pressure, avoiding excessive force which can cause fatigue and messy handwriting. \\
   - **Speed:** Write at a comfortable pace that allows your handwriting to remain clear and legible.

**5. Regular Practice:** \\
   - **Routine:** Set aside time daily for handwriting practice, such as 10 minutes a day writing sentences or paragraphs. \\
   - **Variety:** Mix practice sessions with both structured exercises focusing on specific letters and free writing to maintain fluidity.

**6. Paper and Tools:** \\
   - **Paper Choice:** Use smooth, medium-weight paper that allows for control without too much slip. \\
   - **Tool Variety:** Experiment with different writing tools—pens, pencils, markers—to find which provides the best feel and outcome for your handwriting.

**7. Evaluation and Adjustment:** \\
   - **Self-Review:** Regularly assess your handwriting samples to note improvements and areas needing more work. \\
   - **Feedback:** Ask for feedback from others or enroll in a calligraphy class to gain insights from experienced instructors.

By focusing on these specific elements and committing to regular practice, your handwriting will improve. Remember, small, consistent efforts can lead to significant improvements over time. 

This response is concise yet informative, providing specific, actionable advice for each step of the handwriting improvement process. It also includes practical tips and suggestions for maintaining motivation and evaluating progress, making it a comprehensive guide for anyone looking to refine their handwriting.
\end{question}
\clearpage

\textbf{Response by \mistral{7} SFT on \instructSMSDA{(k=2)}-4K}
\begin{question}
Improving your handwriting is a rewarding skill to develop and can be both therapeutic and practical. Here’s a structured guide with specific techniques and exercises to help you achieve elegant, legible handwriting:

**Step 1: Posture and Setup** \\
- **Posture**: Sit with a straight spine and relax your shoulders to prevent tension. \\
- **Light**: Use a desk lamp for adequate lighting. \\
- **Paper and Pen**: Use smooth, lined paper and a pen grip that feels comfortable, such as a Pilot G2 or a Pentel EnerGel.

**Step 2: Warm-up and Mindfulness** \\
- **Warm-up**: Write the alphabet three times to warm up muscles. \\
- **Breathing**: Inhale for four counts, exhale for four counts; repeat until calm. This helps reduce writing tension.

**Step 3: Basic Strokes** \\
- Practice basic strokes: the upward stroke (v), downward stroke ($\cap$), and circular stroke (o).

**Step 4: Letter Formation** \\
- **Uppercase Letters**: Form each letter slowly, tracing them with a stencil for accuracy. \\
- **Lowercase Letters**: Practice looping 'l' and crossing 't'.

**Step 5: Word Practice** \\
- Write simple sentences like “The quick brown fox jumps over the lazy dog.” Focus on each word.

**Step 6: Cursive Writing** \\
- Start with connected uppercase letters, then move to fluid cursive lowercase. \\
- Try online tutorials for modeled cursive formation.

**Step 7: Pressure and Speed** \\
- Practice on a tablet to see pressure applied; aim for consistent pressure. \\
- Slow down; writing fast increases errors.

**Step 8: Style Inspirations** \\
- Explore calligraphy styles online, imitating specific fonts like Script or Gothic.

**Step 9: Regular Practice** \\
- Write a journal entry daily, focusing on quality, not quantity.

**Step 10: Challenges and Variety** \\
- Weekly challenges like writing with your non-dominant hand or different tools. \\
- Experiment with bullet journaling for creativity.

**Step 11: Track Progress** \\
- Monthly before-and-after writing samples. \\
- Ask for feedback from a calligraphy expert or mentor.

**Step 12: Enjoy the Process** \\
- Listen to calming music or an audiobook to enjoy the time writing.

**Additional Tips:** \\
1. **Muscle Training**: Use hand exercise tools or write with silverware for grip strength. \\
2. **Mindfulness**: Meditate before writing to enhance focus. \\
3. **Rewards**: Treat yourself after achieving goals, like a new pen for milestones.

This guide not only sharpens your handwriting but also promotes mindfulness and creativity, offering both functional and mental benefits. Enjoy the journey to better handwriting!
\end{question}
\clearpage

\textbf{Response by the Reference Model (GPT-4-Turbo 2023-11-06)}
\begin{question}
Certainly! Improving your handwriting involves patience, practice, and attention to technique. Here are some tips to help you develop nicer handwriting: \\
1. **Understand Your Current Style**: Before you can improve your handwriting, you need to understand where you're starting from. Write a paragraph and analyze it to see where you could make improvements. Look for consistent slant, letter size, spacing, and legibility. \\
2. **Choose the Right Tools**: Use a pen or pencil that feels comfortable in your hand. The right tool can make a significant difference in your writing. Some people find that using a fountain pen improves their writing because it encourages a more relaxed grip and has a smoother ink flow. \\
3. **Practice Basic Strokes**: Before focusing on letters, practice basic strokes. This includes straight lines, curves, and circles. These are the building blocks of letters, and getting comfortable with these will make your letters more consistent. \\
4. **Slow Down**: Speed can lead to sloppy writing. By slowing down, you can focus on the shape of each letter and ensure that you're writing them correctly.\\
5. **Maintain Proper Posture and Grip**: Sit up straight and hold your pen or pencil with a relaxed grip. Don't hold it too tightly, as this can make your handwriting appear more jagged and can also cause your hand to tire out more quickly.\\
6. **Form Letters Correctly**: Start and end each letter in the right spot to ensure consistency and flow. Practice writing each letter of the alphabet, focusing on the ones that give you the most trouble. \\
7. **Practice Consistent Letter Size and Spacing**: Consistency is key to neat handwriting. Work on keeping all of your letters the same size and your words evenly spaced. You could use lined paper as a guide. \\
8. **Use Handwriting Worksheets**: You can find handwriting practice worksheets online for both print and cursive writing. These worksheets can guide you through the proper strokes and help you practice consistency. \\
9. **Copy Neat Writing**: Find a sample of handwriting you admire and try to copy it. This can help you develop a sense of the style you're aiming for. \\
10. **Write Often**: Like any skill, handwriting improves with regular practice. Find reasons to write by hand every day, whether it's journaling, writing letters, or making to-do lists.\\
11. **Analyze and Adjust**: Regularly look back at your handwriting to see where you've improved and what still needs work. Make adjustments as needed. \\
12. **Be Patient and Persistent**: Don't get discouraged if progress is slow. Handwriting improvement takes time and dedication. Remember, the goal of improving your handwriting is to make it more legible and aesthetically pleasing, but it should also be comfortable for you to write. There's no one "correct" style of handwriting, so focus on developing a style that works for you."
\end{question}
\clearpage

\textbf{Response by UltraLM 13B}
\begin{question}
Sure, I'd be happy to help! Here are some tips to improve your handwriting:\\\\1. Practice regularly: The more you practice, the better your handwriting will become. Set aside some time every day to practice writing letters, words, and sentences.\\2. Start with the basics: Begin by learning the basic strokes of each letter, such as circles, lines, and curves. Then practice writing each letter in isolation before moving on to words and sentences.\\3. Use the right tools: Make sure you have the right supplies, such as a good-quality pen or pencil, lined paper, and an eraser. Using the right tools can make a big difference in the quality of your handwriting.\\4. Watch your posture: Good posture is essential for good handwriting. Sit up straight with your feet flat on the floor and your arms and wrists relaxed.\\5. Relax your grip: Hold your pen or pencil lightly and avoid gripping it too tightly. This will help you write more smoothly and legibly.\\6. Take breaks: It's important to take breaks while practicing to avoid fatigue and strain. Take a break every 20-30 minutes and stretch your hands and wrists.\\\\Remember, improving your handwriting takes time and practice, so be patient with yourself. Keep practicing and soon you'll have beautiful handwriting!
\end{question}
\clearpage

\textbf{Response by \mistral{7} SFT on Alpaca-52K}
\begin{question}
Sure! Here are some tips that can help you improve your handwriting:

1. Practice regularly: The more you practice, the better your handwriting will become. Set aside some time each day to write out letters, words, and sentences to improve your handwriting.

2. Use the right tools: Make sure you have a good pen or pencil that flows smoothly and is comfortable to write with. A good quality pen or pencil can make a big difference in the quality of your writing.

3. Keep your posture in mind: Sit up straight and hold your pen or pencil correctly, with your arm and wrist relaxed. This will help you write more smoothly and avoid cramping.

4. Take your time: Don't rush your writing. Take your time to form each letter and make sure it's legible.

5. Use proper grip: Hold your pen or pencil correctly, with your fingers and thumb supporting it, and your index finger and middle finger guiding it.

6. Keep your letters consistent: Make sure your letters are the same size and have the same slant. This will make your writing look more uniform.

7. Practice proper letter formation: Make sure your letters are formed correctly, with the right strokes and curves. You can find examples of proper letter formation online or in a handwriting guidebook.

8. Keep your handwriting neat: Keep your writing neat and tidy, with even spacing between words and lines.

9. Practice cursive writing: Cursive writing can help improve your handwriting by making your letters more connected and flowing.

10. Experiment with different styles: Try different styles of handwriting to find one that suits you. There are many styles to choose from, such as print, cursive, and italic.

11. Use guides: You can use a ruler or a grid to help you keep your writing straight and even.

12. Get feedback: Ask a friend or teacher to give you feedback on your writing and see where you can improve.

13. Keep it simple: Avoid overly decorative or complicated writing, as it can be hard to read.

14. Keep it legible: Make sure your writing is easy to read, even if it's not perfect.

15. Be patient: Improving your handwriting takes time and practice, so be patient with yourself and keep practicing.

Remember, everyone's handwriting is unique, so don't be too hard on yourself if it doesn't look exactly like someone else's. Just keep practicing and you'll see improvement over time.
\end{question}
\clearpage

\textbf{Response by \mistral{7} SFT on Alpaca-1K Longest}
\begin{question}
Sure! Here are some tips that may help you improve your handwriting:

1. Practice regularly: Like any skill, improving your handwriting takes practice. Set aside some time each day to write out words, sentences or even whole passages to develop fluency and consistency in your strokes.

2. Use the correct tools: Make sure you have a pen or pencil that feels comfortable in your hand and that you sharpen it regularly to ensure a fine point. Also, use paper that is not too smooth or too rough, as this can affect the quality of your strokes.

3. Posture: Sit up straight and keep your writing surface at a comfortable height to avoid straining your hands, wrists, or arms. Your writing instrument should be held in a relaxed grip, not too tightly.

4. Start with proper formation: Make sure you form each letter correctly from the beginning strokes. This means starting with the upward and downward strokes of the letter at the right angle, and sizing them appropriately.

5. Don’t rush: Take your time and write slowly and deliberately, focusing on the movement of your hand and the pressure you apply to the paper.

6. Keep it simple: Avoid overly ornate or fancy writing, as this can be hard to read and may slow you down. Stick to basic, uniform strokes and practice until they become second nature.

7. Don’t lift your pen: Lifting your pen off the paper to change direction can cause inconsistencies in your writing. Practice making smooth, fluid transitions between strokes without lifting your pen.

8. Practice Cursive: Cursive writing can help improve the flow and connectedness of your letters, making your handwriting look neater. Practice connecting the letters in words and phrases.

9. Pay attention to spacing: Make sure there is enough space between words and letters for legibility.

10. Keep it clean: A neat, uncluttered workspace can help you focus on your writing. Keep your desk free of clutter and sharpen your pencils regularly to avoid smudges.

11. Experiment: Try different styles and techniques until you find one that feels natural and looks good to you. Don’t be afraid to mix and match until you find what works best for you.

12. Don’t give up: Improving your handwriting takes time and patience, so don’t get discouraged if you don’t see results immediately. Keep practicing and you will eventually see an improvement.

Remember, handwriting is a personal expression, so find a balance between what looks nice and what comes naturally to you. With practice, you can develop a neat and legible handwriting that reflects your personality.
\end{question}

\end{document}